\documentclass[phd]{osudissert96}

\def\BibTeX{{\rm B\kern-.05em{\sc i\kern-.025em b}\kern-.08em
    T\kern-.1667em\lower.7ex\hbox{E}\kern-.125emX}}

\usepackage[T1]{fontenc} %straight quotes
\usepackage[utf8]{inputenc}  % only if using old TeX engine
\usepackage{lmodern}
\usepackage{setspace}
\usepackage[parfill]{parskip}

\usepackage{booktabs, multirow, array}
\usepackage{tabularx, longtable, tabulary}
\usepackage{ltablex}
\usepackage{makecell}

\usepackage[normalem]{ulem}     % for underlining (\uline)
\usepackage{hyperref}
\hypersetup{
    colorlinks=true,
    linkcolor=blue,
    citecolor=blue,
    urlcolor=blue
}

\usepackage{amsmath, amssymb, amsthm, amsfonts, mathrsfs}

\usepackage[ruled,vlined]{algorithm2e}
\usepackage{algorithmic}

\usepackage{graphicx}
\usepackage{caption}
\usepackage{pdflscape}
\usepackage[table]{xcolor}
\usepackage{bigstrut}
\usepackage{fvextra} % wrapping verbatim
\keepXColumns

\usepackage[comma,sort&compress,numbers]{natbib}

\usepackage{rotating}
\usepackage{tensor}
\usepackage{authblk}  % for author affiliations
\usepackage{enumitem} % for better lists
\usepackage{appendix}
\usepackage{multicol}

\usepackage{chngcntr} %%%continuous ordering of figures
\usepackage{tcolorbox}  % Box + caption
\usepackage{float} %prevents figure floating

\renewcommand\typesetChapterTitle[1]{#1}

\begin{document}

%
% First, declare the parts of your title page 
%

\author{Yue Chu}
\title{Leveraging Language Models and Machine Learning in Verbal Autopsy Analysis}

\authordegrees{M.B.B.S, MSPH, M.A.}  % Degrees thus far, not including this one.
\unit{Department of Sociology}

\advisorname{Samuel Clark}
\member{Tyler McCormick}
\member{Zhenke Wu} 
\member{David Melamed, and Grzegorz Rempala} %too many, no room for more lines

% The following creates the title page
\maketitle

% Copyright page.
\disscopyright

%
% Abstract goes here.
%

\begin{abstract}
  %  Word limit: 500

This thesis advances the growing body of knowledge at the intersection of advanced natural language processing (NLP), epidemiology, and global health. It is one of the first to comprehensively investigate the application of domain-adapted, pre-trained language models (PLMs) for cause of death (COD) classification in the verbal autopsy (VA) context, as well as the exploration of multimodal fusion strategies that explicitly leverage the inherently multimodal nature of VA data - supported by empirical evidence. It is also among the first to quantitatively characterize the landscape of information sufficiency in VA.

Timely and accurate COD estimates are critical to inform policy and program priorities, as well as tracking progress toward various targets defined in the Sustainable Development Goals framework, particularly in resource-constrained settings. In countries without comprehensive civil registration and vital statistics systems, VA is a critical tool for estimating COD and quantifying the burden of disease. In VA, trained interviewers ask proximal informants for details on the signs, symptoms, and circumstances preceding a death. The resulting data are multimodal with both unstructured narratives and structured questions. Physicians primarily use narratives to identify a COD. In contrast, existing automated VA cause classification algorithms only use the questions - a situation that ignores the additional information available in the narratives. Recently, automated algorithms have become increasingly important in routine (non-research) mortality surveillance applications because they are cheap, quick, and generate reproducible results.
 
In this thesis, we investigate how the VA narrative can be used for automated COD classification using PLMs and machine learning (ML) techniques. Using empirical data from South Africa, we demonstrate that with the narrative alone, transformer-based PLMs with task-specific fine-tuning outperform leading question-only algorithms (such as InSilicoVA) at both the individual and population levels. The narrative-only approach performs particularly well in identifying non-communicable diseases compared to the existing question-only approach.
 
Building on the unimodal findings, we explore various multimodal fusion strategies combining narratives and questions in unified frameworks. Multimodal approaches further improve performance in COD classification, confirming that each modality has unique contributions and may capture valuable information that is not present in the other modality.
 
Using empirical evidence, we characterize physician-perceived information sufficiency in VA data. We describe variations in sufficiency levels by age and COD and demonstrate that classification accuracy is affected by sufficiency for both physicians and automated methods. Finally, we investigate the potential for ML methods to predict and explain physician-perceived sufficiency. These findings reveal where VA data, coming from the VA instrument and interview, need to be improved to further refine cause classification.
 
Overall, this thesis demonstrates the value of the VA narrative in enhancing COD classification. Our findings underscore the need for more high-quality data from more diverse settings to use in training and fine-tuning PLM/ML methods, and additionally, they offer valuable insights to guide the rethinking and redesign of the VA instrument and interview. This study provides a clear example of how PLM and ML can greatly improve existing approaches in epidemiology, population health, and social science.

% \begin{quote}
% \end{quote}

\end{abstract}

%  Dedication
\dedication{To my family, whose love and guidance have shaped who I am.}

\begin{acknowledgements}

I am deeply grateful to Pam Groenewald and the South African National Cause of Death Validation Project team for collecting, managing, and sharing the data used in this dissertation. I thank Jason Thomas and Richard Li for their valuable input on data processing and methodology. I also acknowledge the Translational Data Analytics Institute (TDAI) for access to computing resources at the Ohio Supercomputer Center. My sincere thanks go to Clarissa Surek-Clark for the many enjoyable field trips together and for generously sharing her profound knowledge of VA interviews and practical field experience.

I am especially thankful to my advisor, Sam Clark, for being a deeply supportive and encouraging mentor throughout this journey. I am particularly grateful for his generosity in sharing knowledge and experience without reservation, which has allowed me to grow both intellectually and personally. I thank my outstanding committee members—Zhenke Wu, Tyler McCormick, Dave Melamed, and Greg Rempala—for their support, time, and insightful feedback throughout this process.

I am indebted to my mentors at Fudan University — Xu Qian, Hong Jiang, and Guiying Wu—who guided me through my first research projects and fellowship applications as an undergraduate. I am also grateful to Li Liu and Bob Black, and to my time at Johns Hopkins Bloomberg School of Public Health, for opening the door to the field of global health and epidemiology, and for inspiring me through their dedication, integrity, and passion for impactful research.

I thank all my friends who helped keep me grounded over the years. Special thanks to Yi, Shuo, Ximin, Lang, Romy, and DTY for always being there, no matter how far apart we are. I would also like to thank Hanli—this journey truly began with an email she sent during our first winter break in college. That message planted the seed for my curiosity about JHSPH and global health, and inspired me to reflect on the path I wanted to pursue. I am grateful to my two furry companions, Vodka and Hazzie, for their comforting presence, especially during the hardships of the pandemic. And to Nintendo, Roger Federer, Tchaikovsky, Bach, Coldplay, Queen, Green Day, Beyond, and POI — thank you for helping me through stressful times.

Finally, I thank my mom and dad. They have always listened to me, even when they did not fully understand my work. They gave me the freedom to explore the world without imposing boundaries, yet were always there behind me. So I was never afraid to take on challenges, make mistakes, or venture into completely new areas. They are the best parents I could ever ask for, and I would not have come this far without their unconditional and unwavering love and support. 

% 谨将此文献给我最亲爱的妈妈爸爸，感谢他们无条件的爱与支持。

\end{acknowledgements}

\begin{vita}

\dateitem{2012}{M.B.B.S, Fudan University}

\dateitem{2014}{M.S.P.H, Johns Hopkins Bloomberg School of Public Health}

\dateitem{2020}{M.A., Sociology, The Ohio State University}

\dateitem{2018-present}{Sociology, The Ohio State University}

%\begin{publist}
%
%\researchpubs
%
%\pubitem{Y.~Chu, Marston M, Dube A, Festo C, Geubbels E, Gregson S, Herbst K, Kabudula C, Kahn K, Lutalo T, Moorhouse L.
%\newblock ``Temporal changes in cause of death among adolescents and adults in six countries in eastern and southern Africa in 1995–2019: a multi-country surveillance study of verbal autopsy data''.
%\newblock {\em The Lancet Global Health}. 2024 Aug 1;12(8):e1278-87.}
%
%\end{publist}

\begin{fieldsstudy}

\majorfield{Sociology} 

%\vspace{\baselineskip}  % adds one empty line space (same as between paragraphs)

Specialization: Graduate Interdisciplinary Specialization in Demography \\

Minor Field: Statistics

%\vspace{\baselineskip}  % adds paragraph-like spacing

\end{fieldsstudy}

\end{vita}

% Table of Contents 

\tableofcontents
\listoftables
\listoffigures

% List of abbreviations
\begin{listofabbr}

%\begin{multicols}{1}
\begin{description} %[leftmargin=1.5cm,labelsep=0.5cm]
  \item[ANN] Artificial neural network
%  \item[AUC] Area under curve
  \item[BERT] Bidirectional Encode Representations from Transformers
  \item[BoW] Bag-of-words
  \item[COD] Causes of death
  \item[COPD] Chronic obstructive pulmonary disease
  \item[CSMF] Cause-specific mortality fractions
  \item[GBDT] Gradient-boosted decision trees
  \item[EHR] Electronic Health Records
  \item[FN] False negative
  \item[FNN] Feedforward neural network
  \item[FP] False positive
  \item[GPT] Generative pre-trained transformer
  \item[HDSS] Health and Demographic Surveillance Sites
  \item[HIV/AIDS] HIV and Acquired Immunodeficiency Syndrome
  \item[ICD-10] International Classification of Disease, 10th version
  \item[KNN] K-nearest neighborhood
  \item[ML] Machine learning
  \item[NCD] Non-communicable disease
  \item[NLP] Natural language processing
  \item[PHMRC] Population Health Metrics Research Consortium
  \item[PLM] Pretrained language model
  \item[SANCOD] South Africa National Cause of Death Validation Project
  \item[SHAP] SHapley Additive exPlanations
  \item[SMOTE] Synthetic minority over-sampling techniques
  \item[SOTA] State-of-the-art 
  \item[SVM] Support Vector Machines
  \item[TB] Tuberculosis
  \item[TFiDF] Term frequency – inversed document frequency
  \item[TN] True negative
  \item[TP] True positive
  \item[VA] Verbal Autopsy
  \item[WHO] World Health Organization
\end{description}
%\end{multicols}

\end{listofabbr}

\newpage

\pagenumbering{arabic}
\setlength{\parskip}{0pt}

% Chapters

\chapter{Introduction}
\label{intro.ch}

\section{Overview}

Timely and accurate cause of death (COD) estimates are essential for understanding disease burden, guiding public health interventions, and tracking population health progress. Yet globally, most deaths—particularly in low- and middle-income countries with the highest mortality burdens—are not medically certified.\cite{Nkengasong2020} In these settings, COD estimation is hindered by the absence of comprehensive civil registration and vital statistics systems. Verbal autopsy (VA) is a critical tool for ascertaining the underlying COD in such contexts. When a death is identified, trained fieldworkers conduct in-person household interviews with a close caregiver or next-of-kin of the deceased to collect information on the signs, symptoms, and circumstances preceding death, along with demographic characteristics. This information is recorded in a multimodal format, combining both unstructured free-text narratives and structured questionnaire responses.

Conventionally, trained physicians review information from both narratives and questions from VA to fill in a death certificate-like record listing underlying, immediate, and contributing causes. While effective, this approach is time-consuming, expensive, and heavily dependent on the availability of manual labor of experts, thus extremely challenging to scale up in low-resource settings. Automated approaches streamlining COD ascertainment have gained increasing interest for their efficiency, reproducibility, and cost-effectiveness.~\cite{Clark2021, McCormick2016} Several computer-based algorithms, such as InterVA,~\cite{Byass2019} InSilicoVA,~\cite{McCormick2016, Clark2015} Tariff~\cite{Serina2015} etc., have been developed to automate COD ascertainment from VA data. But mainstream algorithms used in routine non-research mortality surveillance rely exclusively on structured questions. The narratives in VA, though recognized as a critical component for both household interviews~\cite{Loh2021} and physician decision-making,~\cite{King2016} are traditionally used only through manual coding and qualitative interpretation, and haven’t yet been utilized in automated COD classification using VA data. 

With the burst of natural language processing (NLP) and machine learning (ML), the value of narrative in VA has gained increasing interest in epidemiology and bioinformatics research. A few studies explored the potentials of using VA narratives in automated COD classification, however the model performance largely varied across studies and settings,~\cite{Danso2014, Jeblee2019, Blanco2021, Manaka2022, Manaka2022a, cejudo2023cause} and the narratives mainly were processed alone rather than in a synergistic multimodal with the structured questions.~\cite{Manaka2022a, cejudo2023cause} The emergence of large pretrained language models (PLMs) since late 2010s marked a paradigm shift in NLP. Transformer-based models like Bidirectional Encoder Representations from Transformers (BERT)~\cite{devlin2019bert} and domain-adapted variants (such as BioClinicalBERT~\cite{alsentzer2019publicly}, BlueBERT~\cite{peng2019transfer}) have shown superior ability to capture contextual nuances, and have achieved state-of-the-art (SOTA) performances in multiple downstream NLP tasks. Yet no studies have thoroughly explored the potential of these models in the context of VA. Studies are still needed to understand the value of narrative in VA further, and to investigate how narratives could be incorporated into the automated COD classification pipeline for enhanced performance. 

\section{Research Questions}

The overarching aim of this thesis is to investigate promising approaches for improving COD estimations from VA, leveraging the recent advancements in PLM and ML techniques. 

In particular, this thesis consists of three closely connected parts that set out to answer the following research questions:

\begin{itemize}
\item \textbf{Research question 1:} How much can we learn about the underlying COD from the VA narratives? How do current SOTA PLMs perform in the context of VA? 
	\begin{itemize}
	\item To address this question, we evaluate the performance of COD classification using VA narratives alone, applying popular PLMs adapted to the biomedical and clinical domain. (Chapter 3)
	\end{itemize}
\item \textbf{Research question 2:} How can information from both modalities - the narratives and the structured questions - be combined for automated COD classification using VA? Would multimodal learning approaches improve classification performance compared to the current unimodal algorithm?
	\begin{itemize}
	\item To address this question, we propose and test a few multimodal fusion strategies, using information from both the narratives and the questions for COD classification. (Chapter 4)
	\end{itemize}
\item \textbf{Research question 3:} What is the level of information sufficiency in the empirical VA? Could current models identify cases where VA is insufficient for determining underlying COD? 
	\begin{itemize}
	\item We descriptively report the level of information sufficiency of VA as perceived by trained physicians, and explore the ability of ML models to predict information sufficiency of VA data. (Chapter 5) 
	\end{itemize}
\end{itemize}

\section{Structure of thesis}

The rest of this thesis is organized as follows. 

\begin{itemize}

\item \textbf{Chapter~\ref{review.ch}} reviews the background of this research. It introduces the pretrain-finetuning paradigm with PLMs, common methods for hyperparameter optimization, various multimodal fusion strategies, and commonly used evaluation metrics when applying PLM/ML methods.

\item \textbf{Chapter~\ref{plmva.ch}} investigates the narrative-only, unimodal approach for automated COD classification. Using empirical VA data from South Africa, it evaluates the performance of task-specific fine-tuned PLMs in COD classification. The chapter also explores the robustness of the top-performing model via sensitivity analyses on training data size, evaluation metrics, and COD grouping categories.

\item \textbf{Chapter~\ref{multimodal.ch}} investigates the multimodal approach for automated COD classification. It first describes the detailed implementation of proposed multimodal fusion strategies in the context of VA. Then it applied and compared different fusion strategies: narrative-only classification with PLMs from the previous chapter, question-only classification with major ML models, multimodal learning with proposed strategies, comparing against the current question-only statistical model using InSilicoVA. 

\item \textbf{Chapter~\ref{sufficiency.ch}} explores the level of information sufficiency for assigning COD using VA, as perceived by physicians. It first illustrates the level and variation of information sufficiency and demonstrates the association between information sufficiency and classification accuracy. Then it explores the potential to identify and understand information sufficiency with NLP/ML techniques. 

\item \textbf{Chapter~\ref{conclusion.ch}} concludes the thesis with discussion and directions for future research. 

\item And finally, \textbf{Appendix~\ref{app:method}} is an appendix with additional methodological details that is not included in the main text, and \textbf{Appendix~\ref{app:figs}} is an appendix containing additional graphs and figures of the results. 

\end{itemize}

  %intro
\chapter{Background}
\label{review.ch}

This chapter reviews basic methods closely related to the modeling approaches in the rest of the thesis: (1) the pretrain-finetune paradigm and domain-adapted PLMs; (2) hyperparameter optimization; (3) multimodal fusion strategies; (4) evaluation metrics. 

\section{Pretrained language models}

The introduction of transformer architecture by Vaswani et al. in 2017~\cite{vaswani2017attention} marked a revolutionary advance in NLP. By introducing an innovative multi-headed self-attention mechanism, the transformer effectively captures long-range dependencies and contextual meanings in the text as a whole, while learning various feature spaces for nuanced patterns and context in the text in parallel with efficiency. Transformers have since formed the backbone of most major large language models that have been dominating the NLP application, such as BERT, GPT, etc.

\textbf{BERT}~\cite{devlin2019bert} and its variants have been widely recognized for their strong performance in language understanding since their introduction in 2018. Compared to other transformer-based models, BERT learns word representations using a deep bidirectional approach, considering both left and right contexts simultaneously. The model begins by tokenizing the input text and mapping each token to a high-dimensional vector, integrating three components: token embeddings, segment embeddings, and position embeddings of the text. Special tokens such as [CLS] and [SEP] are added to the input sequence, where [CLS] serves as a summary representation attending to all tokens in the sequence and [SEP] denotes sentence separation. The architecture of BERT-style models stacks multiple bidirectional Transformer encoder layers (12 in BERT-base and 24 in BERT-large), each containing multi-head self-attention and feedforward sublayers that iteratively refine contextual representations of the tokens. For downstream classification tasks, the final hidden state corresponding to the [CLS] token is fed into a fully connected layer with softmax activation, producing class probabilities (e.g., for predicting underlying COD). The predicted label is typically taken as the class with the highest probability in a multiclass classification setting.

\subsection{The pretrain-finetune paradigm with PLMs}

The key strength of PLMs like BERT is its two-stage training paradigm:
\begin{enumerate}
\item \textbf{Pre-training :} The foundation models are first trained from scratch on large-scale corpora, either general or domain-specific, for self-supervised objectives. The model learns contextual representations of words in the domain by iteratively updating the weights (learned parameters) across various layers of the model architecture. 
\item \textbf{Fine-tuning:} The pre-trained model is fine-tuned on smaller, task-specific labeled datasets using supervised learning to adapt the model to the task with minimal modifications to the pre-trained architecture. In this stage, the model weights (parameters) are initialized from pre-trained weights and further updated through fine-tuning to learn task-specific patterns. 
\end{enumerate}

The domain-specific pre-training \& task-specific fine-tuning pipeline can be particularly beneficial in the field of medicine and health, where high-quality labeled data are scarce and the terminologies are highly specialized and domain-specific. Task-specific fine-tuning a BERT-style model pre-trained in the biomedical and clinical domains would require much less training data while taking advantage of the learnt domain knowledge, which makes it promising in contexts like VA. 

The base BERT model is pre-trained with English Wikipedia and BookCorpus for self-supervised tasks of masked language modeling and next sentence prediction to learn word representations in the general English language. For biomedical/clinical domain adaptation, popular corpora for further pre-training include scientific publications and electronic health records (EHR), for example:

\begin{itemize}
\item \textbf{PubMed:} PubMed is an open-access, actively-updated archive containing over 30 million peer-reviewed research articles in the field of biomedicine, medicine, and health. PubMed abstracts are a widely used data source by many PLMs for biomedical/clinical adaptation. Full-text articles from PubMed Central (PMC) are also used in addition to abstracts in some models. 
\item \textbf{MIMIC:} Medical Information Mart for Intensive Care III (MIMIC-III),~\cite{Johnson2016} consists of over 2 million de-identified notes from the EHR of 58,976 unique hospital admissions from 38,597 ICU patients between 2001 and 2012 at Beth Israel Hospital in Boston, USA. 
\item \textbf{i2b2:} Informatics for Integrating Biology \& the Bedside (i2b2, now n2c2)~\cite{uzuner2011i2b2} is an open-source database of a large-scale de-identified EHR with annotated clinical corpora like clinical narratives from US hospitals. This dataset is often used as a benchmark dataset for model development and evaluation for clinical domain NLP.
\end{itemize}

\subsection{Popular PLMs in biomedical and clinical domain}

A summary of popular transformer-based BERT-style PLMs contextualized in biomedical and/or clinical domains, and the corpora used for domain-specific pre-training, is in Table~\ref{info:plms}. 

\begin{landscape}

\begin{table}[ht]
\centering
\captionsetup{format=plain,font=small,labelfont=bf, justification=justified, width=\textwidth}
\begin{tabulary}{\linewidth}{llLl}
 \hline
Model                                   & Base model        & Pre-training corpora                    & Domain            \\
  \hline
BioBERT~\cite{Lee2020}                     & BERT-base         & PubMed abstracts, PMC full texts                      & Biomed            \\
BioClinicalBERT~\cite{alsentzer2019publicly}          & BioBERT           & PubMed abstracts, ClinicalTrials.gov, MIMIC-III notes   & Biomed + Clinical \\
BlueBERT~\cite{peng2019transfer}                & BERT-base         & PubMed abstracts, MIMIC-III notes                     & Biomed + Clinical \\
BiomedBERT\cite{gu2021domain}            & BERT-base         & PubMed abstracts, PMC full texts, Clinical notes (MIMIC-IV and i2b2)    & Biomed + Clinical \\
ClinicalBERT~\cite{wang2023optimized}            & BERT         & EHR$^*$       & Clinical          \\
RoBERTa-PM$^\dag$~\cite{Lewis2020} & RoBERTa-distilled & PubMed abstracts, PMC full texts, MIMIC-III, Medical subword vocabularies learnt from PubMed & Biomed + Clinical \\
BioElectra~\cite{kanakarajan2021bioelectra}       & ELECTRA-base      & PubMed abstracts and PMC full texts                                                          & Biomed        \\   
  \hline
\multicolumn{4}{l}{\footnotesize{$^*$ A multi-center dataset with a large corpus of 1.2 billion words of diverse diseases from EHR of hospitalized patients with Type-2 Diabetes}} \\
\multicolumn{4}{l}{\footnotesize{from January 2013 to April 2021 in Zhongshan Hospital and Qingpu Hospital in Shanghai, China.}} \\
\multicolumn{4}{l}{\footnotesize{$^\dag$ for RoBERTa-base-PM-M3-Voc-distill-align-hf checkpoint}} \\
\end{tabulary}
\caption{Popular PLMs in the biomedical and clinical domain}
\label{info:plms}
\end{table}

\end{landscape}

\section{Hyperparameter optimization}
\label{info:hpo}

\subsection{Common hyperparameters}

PLMs, similar to ML models, learn from data through \textbf{gradient descent}. In the process, model parameters are iteratively adjusted and optimized to minimize the loss, a function measuring the difference between predicted and true values. The standard loss function used for multiclass classification is categorical \textit{cross entropy loss}, which measures the difference between the predicted and the true probability distributions. 

The model training is governed by a set of hyperparameters, which set the gradient descent update strategy and tune the speed of convergence. Key hyperparameters for configuring a BERT-style PLM include:

\begin{itemize}
\item \textbf{Learning rate:} The step size in gradient descent updates. A smaller learning rate means more conservative and robust updates, but slower to train.
\item \textbf{Weight decay:} L2 regularization during gradient updates. It penalizes large model weights to discourage overly complex models and helps prevent overfitting.
\item \textbf{Warm-up ratio:} The proportion of total training steps used as a warm-up phase. During warm-up, the learning rate gradually increases from a small initial value to the target learning rate to stabilize training and prevent divergence early on.
\item \textbf{Batch size:} The number of training samples processed in one forward and backward pass before a weight update. Smaller batch sizes mean more frequent updates and may better escape local minima, but are more computationally expensive.
\item \textbf{Max epochs:} The maximum number of complete passes through the training dataset.
\item \textbf{Gradient accumulation steps:} The number of forward-backward passes over mini-batches whose gradients are accumulated before updating the model weights. 
\item \textbf{Freeze layer:} The number of bottom layers kept fixed during fine-tuning. Freezing layers preserves foundational knowledge while reducing training time and overfitting.
\end{itemize}

\subsection{Hyperparameter optimization}

\textbf{Hyperparameter optimization (HPO)} essentially is trialing different combinations of hyperparameters within the predefined search space(s), to identify the configuration that yields the best performance. For each trial, a new combination of hyperparameters is sampled to configure the model, which is then trained and evaluated based on the specified evaluation metric. Performances from all trials are compared, and the configuration that produces the best validation performance is selected as the optimized set of hyperparameters, which will then be used for final model training with the full training dataset.

One commonly used method for exploring the search space is grid search, which exhaustively evaluates all possible combinations of hyperparameters in the search space. While straightforward, grid search can be computationally expensive—particularly for large models like PLMs. Alternatively, Bayesian optimization substantially improves the sufficiency by modeling the past searches to inform the more promising regions to sample next based on previous trials. This approach is widely used in automated HPO tools such as Optuna and Ray Tune, especially for tuning large models.

In this thesis, we used \textbf{Optuna} for HPO for PLM and ML models. Optuna is a lightweight HPO framework that uses Bayesian optimization (e.g., Tree-structured Parzen Estimators) to explore complex hyperparameter spaces efficiently. Common configuration for customized Optuna setup includes:

\begin{itemize}
\item \textbf{Number of trials:} A fixed number of search iterations.
\item \textbf{Pruning:} Early stopping of unpromising trials if the interim performance is not outperforming past trials, helping to reduce computational cost. The pruning process can be further configured with early stopping patience (degree of tolerance before terminating the trial), startup trials (where early trails won't be pruned), and warm-up steps (minimum number of training steps to run before pruning). 
\item \textbf{Objective:} Evaluation metrics used to select the best model from all trials. In multiclass classification task, it can be minimizing cross-entropy loss, or maximizing metrics like weighted F1.
\end{itemize}

\section{Multimodal fusion strategies}
\label{info:mm}

In reality, data in medical and health domain usually take on multiple forms, or modalities, from structured tabular data from survey questions or lab results, to unstructured data such as open narratives, images and audios. VA with unstructured narratives and structured question responses, for example, is inherently a text-tabular multimodal data. Different modalities usually carry unique information, and models can benefit greatly from integrating data from different modalities, especially for classification tasks.

What is fused and when the fusion is performed are two critical components in designing any multimodal framework. Fusion strategies are typically categorized in two ways:
\begin{itemize}
\item \textbf{Timing-based}
\begin{itemize}
\item \textit{Early fusion}: Raw inputs or low-level features from multiple modalities are concatenated and jointly fed into a unified model for downstream tasks.
\item \textit{Late fusion}: Each modality is processed or modeled separately, and model outputs or predictions are combined for final modeling or prediction.
\end{itemize}

\item \textbf{Content-based}
\begin{itemize}
\item \textit{Data-level fusion}: Data from different modalities are combined close to their raw form with minimal processing prior to any modeling, and concatenated data are used as input to the model for the downstream task.
\item \textit{Feature-level fusion}: Features are first extracted independently from each modality, then concatenated and used as input to the modeling pipeline for the downstream task.
\item \textit{Prediction-level fusion}: Modality-specific models are trained independently as base models, predictions from the base models are combined using ensemble learning or through a meta-learner for the final prediction.
\end{itemize}
\end{itemize}

Timing-based terminology can be inconsistently presented in the literature and can lead to confusion. For example, feature-level fusion is sometimes referred to as early fusion and other times as late fusion, depending on the degree of feature processing. We therefore adopt the content-based approach to define and present our multimodal fusion strategies used in this thesis. 

\subsection{Data-level fusion}

Data-level fusion approach is mostly used for data with only text and tabular features. Tabular data are converted to text format, and concatenated text containing information from both modalities is directly used as inputs to NLP / PLM models for downstream tasks. Conversion from tabular to text can be done using simple key-value concatenation, or transformed into more natural-sounding sentences or paragraphs, which leverages the power of contextual embedding from the transformer-based PLMs more. 

For modeling with concatenated text, all text analysis approaches, NLP or PLMs, should work in theory. However, we should note that language models using vanilla transformer architecture, like BERT-based models, have a constraint of a maximum of 512 input tokens. Input token constraint is less of a problem with narrative-only analysis with VA, the length of which mostly falls well below the limitation. The fused text might exceed such a limit, thus calling for either PLMs that are specialized for handling long sequences or short-sequence PLMs with text truncation or a more complex model architecture.

Popular PLMs for processing long sequences include:
\begin{itemize} 
\item \textbf{Longformer~\cite{beltagy2020longformer}:} Longformer uses sliding window attention and a sparse local attention mechanism to reduce computational complexity, which constrains the length of the input sequence in the BERT-based models. It greatly extends the maximum input sequence length from 512 in BERT-based models to 4096 input tokens. 
\item \textbf{BigBird~\cite{zaheer2020big}:} BigBird, developed by Google, also uses combined sparse attention mechanisms - global, random, and sliding window - and can also handle up to 4096 input tokens. 
\item \textbf{Clinical Longformer and Clinical BigBird}: Li et al. \cite{li2022clinical} further adapted the Longformer and BigBird model in the clinical domain, by pre-training with clinical notes from MIMIC-III. 
\end{itemize} 

\subsection{Feature-level fusion}
\label{info:automm}

The feature-level fusion approach independently extracts features from each modality and uses concatenated features for classification models. It usually involves the following steps: 

\begin{enumerate}
\item \textbf{Feature extraction:} extract features from each modality using the most appropriate method given the nature of the data. This approach is not limited to text-tabular data but can handle all input formats - text, tabular, images, audio, etc.. 
	\begin{itemize} 
	\item \textbf{Unstructured data:} For text data, the features can be either linguistic features extracted using traditional NLP (e.g. bag-of-words (BoW) or Term Frequency–Inverse Document Frequency (TF-iDF) \cite{erickson2020autogluon}), word embeddings (e.g. Word2Vec, GloVe), or contextual embeddings from PLMs like BERT.~\cite{Tang2024} For other modalities, neural networks and transformer encoders are commonly used to extract feature embeddings from images or audios.(e.g., ImageNet for image data in AutoMM~\cite{Tang2024})
	\item \textbf{Structured data:} structured / tabular data can be used without further processing,\cite{erickson2020autogluon} or more often encoded using models (e.g., ft-transformer in AutoMM~\cite{Tang2024}) for easier and unified fusion with features from other modalities.
	\end{itemize} 
\item \textbf{Feature fusion:} features from different modalities can be fused via concatenation.
%, or using a more complex design such as an attention mechanism. 
	\begin{itemize} 
	\item \textbf{Dimension reduction:} based on what features are extracted from each modality, some might need to reduce dimensions to a reasonable size before feature fusion. Especially when fusing high-dimensional features (e.g., text-based features) directly with low-dimensional sparse features (e.g., raw VA questions), it is highly recommended to reduce dimension using methods like singular value decomposition (SVD) to prevent having large embeddings dominating the learning signal downstream. 
	\end{itemize} 
\item \textbf{Model training and evaluation:} fused features are used as input for training the models, usually machine learning or deep learning models. In AutoMM~\cite{Tang2024}, Multilayer Perceptron (MLP), a small feedforward network, is used with a softmax layer for the final multiclass classification.
\end{enumerate}

\subsection{Decision-level fusion}

Different modalities, each with distinct data structures, require different model architectures to handle them appropriately. Decision-level fusion is designed to take advantage of the strengths of modality-specific model designs by combining their predictions after being configured and trained separately. 

\begin{enumerate}
\item \textbf{Base models:} Separate models are fit with data for each modality, with each model optimized independently. One or more models can be trained for each modality. And the choice of models is not necessarily limited to ML/PLMs, but can include traditional statistical models as well. 
\item \textbf{Concatenate predictions from base models:} predictions from base models are concatenated, and used as input for ensemble learning.
\item \textbf{Aggregate predictions:} the final prediction can be derived from simple aggregation of base model predictions (voting), or can be predicted by training a meta learner to combine base-model-predictions.
\end{enumerate}

\section{Evaluation metrics}
\label{info:eval}

We used a range of metrics to evaluate model performance in this thesis, accounting for both overall accuracy and class imbalance. 

\textbf{Individual level evaluation}

\begin{itemize}
\item \textbf{Accuracy:}  The proportion of correctly classified cases, where true positive (TP), false positive (FP), true negative (TN), and false negative (FN) are outcomes from the confusion matrix. 
\[
\text{Accuracy} = \frac{TP + TN}{TP + FP + TN + FN}
\]
\item \textbf{Precision:} The proportion of correct positive predictions. 
\[
\text{Precision} = \frac{TP}{TP + FP}
\]
\item \textbf{Recall:}  The proportion of actual positives correctly predicted.
\[
\text{Recall} = \frac{TP}{TP + FN}
\]
\item \textbf{F1 Score:}  The harmonic mean of precision and recall. 
\[
\text{F1 score} = \frac{2 \cdot Precision \cdot Recall}{Precision + Recall}
\]
\end{itemize}

In the multiclass classification setting, we can calculate an overall measure aggregating the per-class measures for precision, recall, and F1 score. There are various aggregation methods:
\begin{itemize}
\item \textit{weighted:} the weighted average of all per-class measures weighted by class support. e.g. where $C$ is the total number of classes, $n$ is class size, $N$ is total sample size: 
\[
\text{weighted F1} =  \sum_{i=1}^{C} \frac{n_i}{N}F1_{i}
\]
\item \textit{macro:} the unweighted mean for all per-class measures regardless of class size. e.g.
\[
\text{Macro F1} = \frac{1}{C} \sum_{i=1}^{C} F1_{i}
\]
\item \textit{micro:} the measure aggregating contributions of all classes, e.g.

\[
\text{Micro F1} = \frac{2 \cdot \sum_{i=1}^{C}TP_{i}}{2 \cdot \sum_{i=1}^{C}TP_{i} + \sum_{i=1}^{C}FP_{i} + \sum_{i=1}^{C}FN_{i}}
\]
\end{itemize}

\textbf{Population level evaluation}

\begin{itemize}
\item \textbf{Cause-Specific Mortality Fraction (CSMF) Accuracy:}  A measure for how well predicted CSMFs match the true distribution.~\cite{Murray2014}
\[
\text{CSMF Accuracy} = 1 - \frac{\sum_{i=1}^{C} \left| \text{CSMF}_{i}^{true} - \text{CSMF}_{i}^{pred} \right| }{2 \left( 1 - \text{min}_i(\text{CSMF}_{i}^{true}) \right)}
\]
\item \textbf{Chance-Corrected CSMF Accuracy (CCCSMF Accuracy):}  CSMF accuracy adjusted for agreement by chance. We used 0.632 as the mean of random allocation consistent with the convention in Flaxman et al. 2015.~\cite{flaxman2015measuring}
\[
\text{CCCSMF Accuracy} = \frac{\text{CSMF Accuracy} - \text{mean of random allocation}}{1 - \text{mean of random allocation}}
\]
\end{itemize}
   % review
\chapter{Automated cause of death classification using verbal autopsy narratives}
\label{plmva.ch}

\section{Introduction}

This chapter investigates the narrative-only unimodal approach for automated COD classification using domain-adapted PLMs, based on empirical VA data. 

\subsection{VA narratives}

\begin{figure}
\begin{center}
\includegraphics{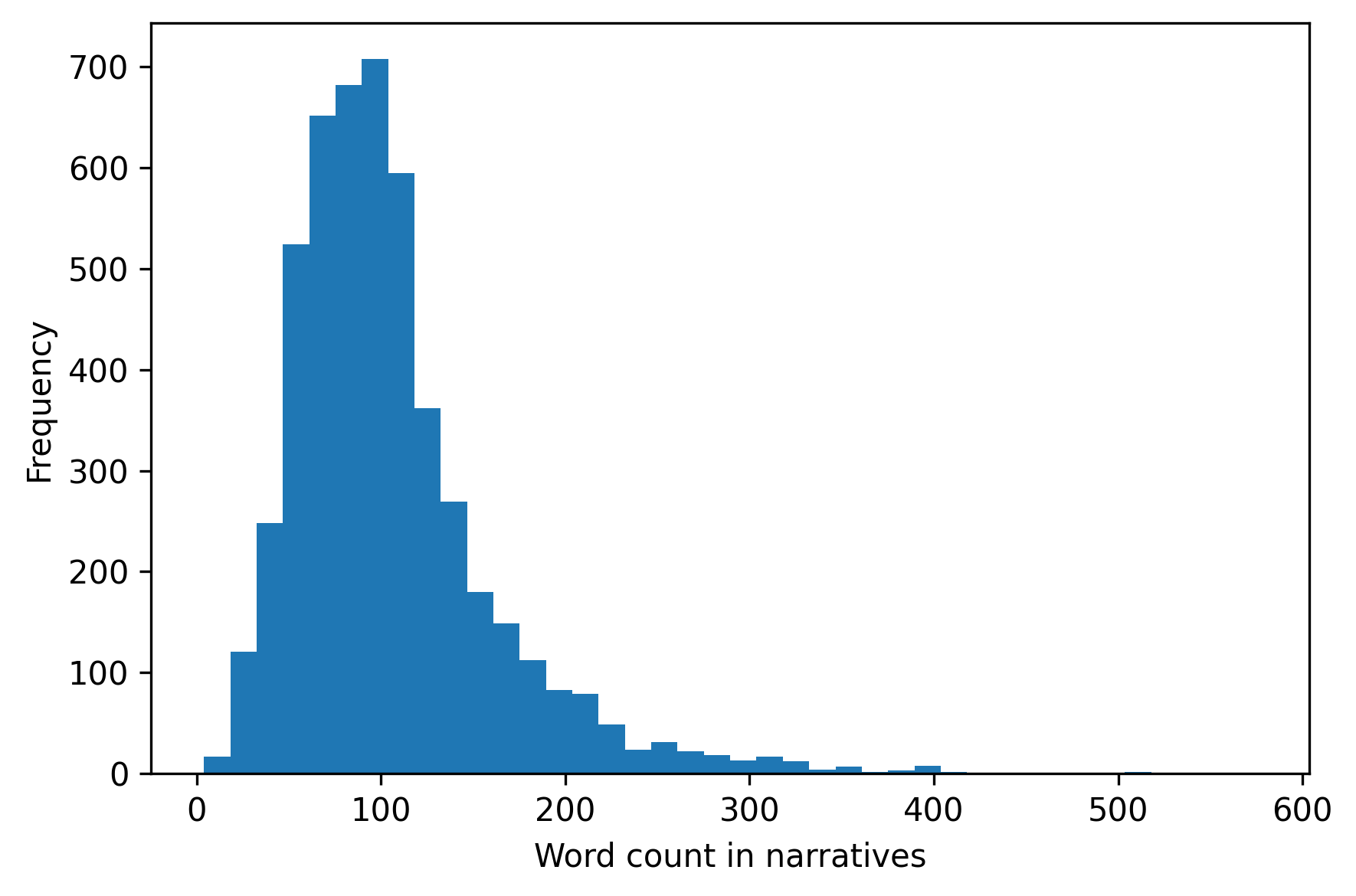}
\caption[Distribution of word count for VA narratives]{Distribution of word count for VA narratives}
\caption*{\footnotesize\textit{Note:} Distribution based on adult VAs used in this study.}
\label{fig:dist_wordcount}
\end{center}
\end{figure}

A typical VA interview usually starts with an open-ended question that allows the respondent to freely describe the events that led to the death, right after clarifying the identity of the deceased with demographic information and before the structured questions. \cite{Nichols2018} The trained interviewers summarize and transcribe the opening stories of respondents, and record them in VA as unstructured narratives.\cite{surek2020verbal} Typically, a narrative would record key information such as the timing and/or sequence of events, indicating medical history or progression of disease, signs and symptoms preceding death, past diagnoses, treatments received, and health services utilization. The information is not consistently documented in standard medical terminology, and may reflect interpretation by the respondent and/or interviewer, introducing potential biases. The narratives are usually brief in length (Figure~\ref{fig:dist_wordcount}), with varying quality and content, and often contain typos, spelling and grammatical errors. A few examples are presented in the textbox below. 

\begin{tcolorbox}[title=A few examples of VA narrative, colback=gray!5, colframe=black, fonttitle=\bfseries]
\begin{Verbatim}[breaklines=true]
"""
Case 1:
My sister in law was a person who was diagnosed with HIV. One day she went to the toilet, fall down and died.
"""
Case 2:
He has been sick for a long time with Hypertension, Diabeti and has Arthritis and was on medication for all those diseases. There was sometimes shortage of medication where he was collecting them at HEALTH CENTRE. He started to complain about body weakness and was taken to HOSPITAL and admitted for 3 weeks. He was unable to walk due to stroke that affected his left side of the body and kidney disease. He passed away at the hospital.
"""
Case 3:
The decedent started being sick in [YEAR MASKED] and was admitted to HOSPITAL. He decedent was vomiting and they found that he is diabetic. He started taking treatment at CLINIC from [YEAR MASKED]. In [DATE MASKED] he was admitted again at HOSPITAL because of diabetes was very low. He was admitted for 2 days and he passed away on the 3 day on the [DATE MASKED].
"""
\end{Verbatim}
\end{tcolorbox}

The narrative has been a critical component not only for the interview process, but also for physicians' decision-making in assigning underlying COD. It allow spontaneous recall of symptoms and events, as well as critical diagnostic information, such as sequence and timing of events, severity onset and progression, past diagnosis, treatment history and health services usage, which might not be well covered by structured questions. Field experiences revealed that the opening-stories make rapport building and probing easier for interviewers \cite{Loh2021} and report additional information to better understand COD and/or health service usage.\cite{King2016} Some physicians reported that they many times even rely more on the narratives rather than the structured questions when coding underlying COD. \cite{surek2020verbal} While qualitative studies also reported that narratives had limited values in improving the quality of questioning\cite{Loh2021} or improving COD assignment,\cite{King2016} and fail to explain discrepancies in diagnosis between physician-coded COD using all available modalities and computer-coded COD using only structured questions.\cite{rankin2012exploring} 

\subsection{Review of methods using narratives for automated COD classification}

Conventionally, the processing and analysis of VA narratives relied heavily on manual coding and qualitative interpretation from trained physicians. With the advent of NLP in the past decade, researchers have began to explore automated methods to classify COD using VA narratives. 

Early attempts used traditional NLP approaches, where linguistic features (e.g. part-of-speech taggings) \cite{danso2013linguistic, Jeblee2019} or frequency-based features (bag-of-words, term-frequency-inverse document frequency) \cite{danso2014comparative} are extracted from the training text, after feature selection, are fed to classic ML models (e.g. Random forest (RF), Support Vector Machine(SVM) etc.) for classification.\cite{danso2013linguistic, danso2014comparative} These methods are straightforward to apply and require minimal computational resource. However, the features from the text capture minimal semantic meaning and have no context-awareness. It's relatively sensitive to the word distribution and representation in the training data. Therefore, they mostly achieved less optimal performances compared to later methods.\cite{danso2013linguistic, Danso2014, Danso2015, Danso2023} These approaches also sometimes involve heavy feature engineering and feature selection,\cite{mujtaba2017automatic} which makes it hard to generalize to other settings. 

The deep learning approach with text representation using word embeddings/character embeddings and neural networks then dominates the next phase of text analysis in the VA context. Embeddings are static dense vector representations of words (in word embeddings methods like word2vec,~\cite{Jeblee2021} GloVe~\cite{Mapundu2023}) or subwords (in character embeddings like character n-grams~\cite{Yan2019}), modeled from training data to capture semantic similarity, analogical relationships between words based on their occurrence and co-occurrence the task-specific context. The embeddings are predicted from neural networks (e.g. word2vec) or count-based models (GloVe), and are usually used with neural networks~\cite{Yan2019, Mapundu2023,Blanco2020, Jeblee2021} or machine learning models~\cite{Blanco2020, Manaka2022a} for downstream classifications. Compared to traditional text representation methods, embeddings-based approach significantly improves the classification accuracy.~\cite{Manaka2022a, Yan2019, Mapundu2023} However, the static representation assigns the same vector to the same word regardless of the context, which could be problematic in the context of VA, if there is overlap between everyday language and medical terms or abbreviations. 

Embeddings from Language Models (ELMo) and transformer-based models like BERT represent the next step forward in NLP with contextual word representation through multi-layer model structures. Both models dynamically vectorize the words depending on the surrounding text, greatly improving the ability for disambiguation of word meanings given context. Jeblee et al.~\cite{Jeblee2021} and Manaka et al.~\cite{Manaka2022} both combined ELMo with Feedforward Neural Network (FNN). Manaka et al.~\cite{Manaka2022} and Cejudo et al.~\cite{cejudo2023cause} briefly tested base BERT, one using BERT-based embeddings with FNN and the other used BERT for classification end-to-end. Although the results were not directly comparable with different cause groupings, the performances still seemed to largely vary between studies, indicating the need for more research. Domain-adapted PLMs, which have shown SOTA performance in understanding unstructured clinical notes for various downstream NLP tasks,~\cite{alsentzer2019publicly, gu2021domain, peng2019transfer} have not been adequately investigated in the VA context. 

\subsection{Transferring domain-adapted PLMs to the VA context}

VA narratives share many similarities with clinical notes or discharge summaries found in electronic health records. Both primarily describe a patient's medical history and frequently use specialized terminology. As such, PLMs adapted to the biomedical or clinical domain hold great potential to transfer to the VA context. In addition, the typical length of VA narratives fits well within the sequence limit of the transformer architecture, which makes VA a perfect use case for BERT-based models. 

Nevertheless, the effectiveness of PLMs must still be empirically validated in the VA context, given the substantial differences between clinical and VA settings. Notably, narratives from clinical settings are authored by medically trained professionals using standardized medical terminology and abbreviations, using unified formatting, and usually contain rich, longitudinal, and precise objective examination and test results matching diagnostic gold standards. In contrast, VA narratives are primarily from community deaths retrospectively recorded during household interviews, with non-medical respondents and interviewers who might also lack systematic medical trainings. The narratives are often much shorter, contain a mixture of medical terms and non-medical expressions, are ambiguous with greater variations in details, reflecting cultural idioms and subjective perceptions of the events, and can be subject to errors and biases. Such differences introduce additional challenges in the transferability of existing PLMs, and require task-specific fine-tuning in addition to domain-specific pre-training to boost downstream performances. 

Therefore, this study aimed to explore the application of PLMs in the context of VA, using empirical data from South Africa. We evaluated the performances of transformer-based PLMs, pre-trained in biomedical/clinical domain. We fine-tuned the selected models with task-specific knowledge on COD classification using only the narratives from VA. We compared the performances of the narrative-based approaches to the baseline with the mainstream statistical model used in practice based on only structured questions (namely InSilicoVA). Classification accuracies at both the individual and population levels were evaluated using physician-coded COD as the reference. We also conducted sensitivity analysis to experiment with the impact of COD grouping, training data size, the choice of evaluation metrics, on model performance to better understand the robustness of this approach. 

\section{Methods}

\subsection{Data}

The study used the narratives from verbal autopsies of South African National Cause-of-death Validation Project (SANCOD).\cite{maqungo2024can} SANCOD conducted VA interviews for all deaths registered during September 2017 to April 2018 from 27 sub-districts, which were randomly selected out of 258 sub-districts pseudo stratified socio-economic status across South Africa. 

Interviewers were trained following the WHO 2016 VA field interviewer manual \cite{who2016VAMANual} with regard to identifying the most appropriate respondent for the VA interview. For collection of narratives, the interviewers asked the respondent to describe, in their own words, the events leading up to the death. Interviewers then document symptoms, timing and duration of symptoms, actions taken in response to the symptoms, whether any treatment was given and whether health provider had indicated a diagnosis or cause of death, and to probe where information on above was not given. The narratives were digitized and transcribed into English. 

VA for adolescents and adults aged 12 years and above were used for analyses given overall data availability and quality. VAs without narratives (N=36) or with invalid narratives (N=7) were excluded from the analysis. A total of 4999 VA with narratives were included in the study. The narratives were then corrected for typos, spelling and grammatical errors with GPT-3.5-turbo. More details about exclusion criteria and the prompt used for GPT Application Programming Interface (API) is in Appendix~\ref{app1:narrative_clean}.

We used \textbf{physician-coded cause of death (PCVA)} grouped into COD categories as the "true labels" for training and evaluation. In SANCOD, 75 physicians were recruited and trained to interpret VA narratives and interviews collected using WHO 2016 VA instruments.\cite{awotiwon2022anaconda} Each VA were independently reviewed by two physicians, and completed a death certificate reporting the  underlying COD along with the causal sequence, coded with International Classification of Diseases 10th version (ICD-10). In cases of disagreement, the VA would be reviewed by the quality assurance reviewers panel for adjudication. Physicians were presented with all information from both narratives and structured questions for coding the COD. We then grouped physician-coded ICD-10 labels for underlying COD into \textbf{three levels}: 
\begin{itemize}
\item Level 1: 34 COD categories based on WHO COD categories with minimal adaptation grouping rare causes with "other and unspecified"; 
\item Level 2: 18 COD categories, keep major cause in each diagnostic category, considering both public health interest and the sample-size per class for decent model training (no less than 15 cases per class); 
\item Level 3: 6 COD categories, namely HIV and pulmonary TB, non-HIV/TB infections, non-communicable causes, injuries, maternal conditions, and indeterminate causes.
\end{itemize}
Detailed ICD-10 mapping and COD groupings are in appendix~\ref{tab:icd10}). We used Level 2 COD by default in this thesis for model training, evaluation, and result visualization, unless stated otherwise. 

\subsection{Candidate models}

We randomly split the data into 80\% training (N=3999) and 20\% (N=1000) testing, stratified by COD. PLMs tested in this study are listed in Table~\ref{ch1:plms}. We initialized the model with weights from the model checkpoints publicly released on Huggingface or GitHub. 

\captionsetup{format=plain,font=small,labelfont=bf, justification=justified, width=\textwidth}

\begin{table}[ht]
\centering
\begin{tabularx}{\linewidth}{lX}
\toprule
\textbf{Model} & \textbf{Model checkpoint} \\
\midrule
ClinicalBERT        & \url{https://huggingface.co/medicalai/ClinicalBERT} \\
BioClinicalBERT     & \url{https://huggingface.co/emilyalsentzer/Bio_ClinicalBERT} \\
BlueBERT            & \url{https://huggingface.co/bionlp/bluebert_pubmed_mimic_uncased_L-12_H-768_A-12} \\
BiomedBERT          & \url{https://huggingface.co/microsoft/BiomedNLP-BiomedBERT-base-uncased-abstract-fulltext} \\
RoBERTa-PM          & \url{https://dl.fbaipublicfiles.com/biolm/RoBERTa-base-PM-M3-Voc-distill-hf.tar.gz} \\
BERT			& \url{https://huggingface.co/bert-base-uncased}				\\
\bottomrule
\end{tabularx}
\caption{Selected models and their checkpoints}
\label{ch1:plms}
\end{table}

\subsection[Hyperparameter optimization]{HPO}

We used Optuna to find the best configuration for each model. The Optuna framework, common parameters for configuring PLMs and Optuna optimization process, as well as metrics for model selection and evaluation, have been introduced in the previous chapter (section~\ref{info:hpo} and section~\ref{info:eval}). 

Only training data was used for HPO. For each model, we ran 30 trials with early pruning and a conservative stopping strategy. Within each trial, we applied 5-fold stratified cross-validation for better robustness. The best trial was selected based on the mean validation weighted F1 score across all folds, for a balanced trade-off between precision and recall while considering true cause distribution. The search spaces for hyperparameters were consistent across all models.(Table~\ref{tab:plmhpo_space}) Best performing hyperparameters for each model used for final model training were presented in Table~\ref{tab:plmhpo_best}. 

%For each model, we ran 30 trials with early pruning, with a conservative stopping strategy of startup_trials = 10, n_warmup_steps=1, early_stopping_patience=5 

\begin{table}[ht]
\centering
\captionsetup{format=plain,font=small,labelfont=bf, justification=justified, width=\textwidth}
\begin{tabulary}{\linewidth}{LCC}
 \hline
Hyperparameter & Search strategy & Search space \\ 
  \hline
Learning rate               & Log-uniform & {[}5e-6, 4e-5{]}  \\
Batch size                  & Categorical & {[}8, 16, 32{]}   \\
Epochs   			& Categorical & {[}3, 5, 7{]}     \\
Weight decay                & Log-uniform & {[}1e-6, 0.1{]}   \\
Warmup ratio                & Log-uniform     & {[}0.0, 0.2{]}    \\
Gradient accumulation steps & Categorical                              & {[}1, 2{]}        \\
Freeze layers               & Categorical                              & {[}0, 4, 8, 12{]} 		\\
  \hline
\end{tabulary}
\caption{Hyperparameters optimization for PLMs}
\label{tab:plmhpo_space}
\end{table}

\begin{table}[ht]
\centering
\captionsetup{format=plain,font=small,labelfont=bf, justification=justified, width=\textwidth}
\begin{tabulary}{\linewidth}{Lcccccc}
 \hline
Hyperparameter              & \thead{\makecell{BioClinical \\ -BERT}} & \thead{\makecell{BioMed \\ -BERT}} & \thead{\makecell{Blue \\ -BERT}} & \thead{\makecell{Clinical \\ -BERT}}  & \thead{\makecell{RoBERTa\\ -PM}}  & BERT     \\
  \hline
Learning rate               & 3.05E-05        & 3.76E-05   & 3.76E-05 & 2.26E-05     & 3.46E-05  & 3.86E-05 \\
Batch size                  & 16              & 8          & 8        & 8            & 16        & 16       \\
Number of training epochs   & 7               & 7          & 7        & 7            & 8         & 5        \\
Weight decay                & 6.22E-06        & 5.80E-02   & 1.40E-04 & 8.45E-04     & 3.90E-02  & 2.70E-04 \\
Warmup ratio                & 0.12            & 0.06       & 0.06     & 0.03         & 0.03      & 0.06     \\
Gradient accumulation steps & 1               & 1          & 1        & 1            & 1         & 2        \\
Freeze layers               & 0               & 0          & 0        & -            & 4         & 0       \\
  \hline
\end{tabulary}
\caption{Best performing hyperparameters for PLMs}
\label{tab:plmhpo_best}
\end{table}

\subsection{Model training and evaluation}

We then trained the models using the full training dataset with model-specific configuration from HPO, and evaluated the model performance on the hold-out testing dataset. Model performances were evaluated at both the individual and population levels. Evaluation metrics used in this study have been introduced in chapter~\ref{review.ch} (section~\ref{info:eval}). 

\begin{itemize}
\item Individual level metrics (based on top cause): accuracy, weighted F1 score, weighted precision, weighted recall;
\item Population level metrics (based on full probability distributions): CSMF accuracy, CCCSMF accuracy
\end{itemize}
%balanced accuracy removed - table too wide

We also compared the performance of PLMs using narrative-alone to current statistical algorithm using only questions - InSilicoVA. \cite{McCormick2016} InSilicoVA in this thesis was applied with customized configuration, where the model was trained with custom cause list and symptom-cause probabilities were informed by the empirical data, to ensure that the results were directly comparable. 

\subsection{Sensitivity analysis}

To further understand the robustness of this fine-tuning approach, we run some sensitivity analysis to test the impact of the following factors on the model performance, using selected model (BioClinicalBERT) as example: 

\begin{itemize}
\item COD grouping: we ran the models end-to-end with different levels of COD grouping (level 1, 2, and 3), with model-specific hyperparameter optimization. We then evaluated how model performance changes with changing granularity of cause grouping.
\item training data size: we rerun the models with 10\%, 20\%, 30\%, ... 90\% of the training dataset to fine-tune the PLM, with the subset stratified sampled from the full training dataset. We trained the PLMs on each subset using the same hyperparameters as the main model, and then evaluated on the same hold-out testing data.
\item evaluation metrics: we ran HPO optimized for minimizing cross-entropy loss, and compared to the above results configuring HPO maximizing weighted F1 score, to understand how sensitive PLMs was to model selection. 
\end{itemize}

The analysis was done using Python 3.12 \cite{python312} and Transformer package \cite{wolf2020transformers} with Huggingface backbones.

\section{Results}

\subsection{Distribution of physician coded cause of deaths}

\begin{figure}
\begin{center}
\includegraphics[width=6in]{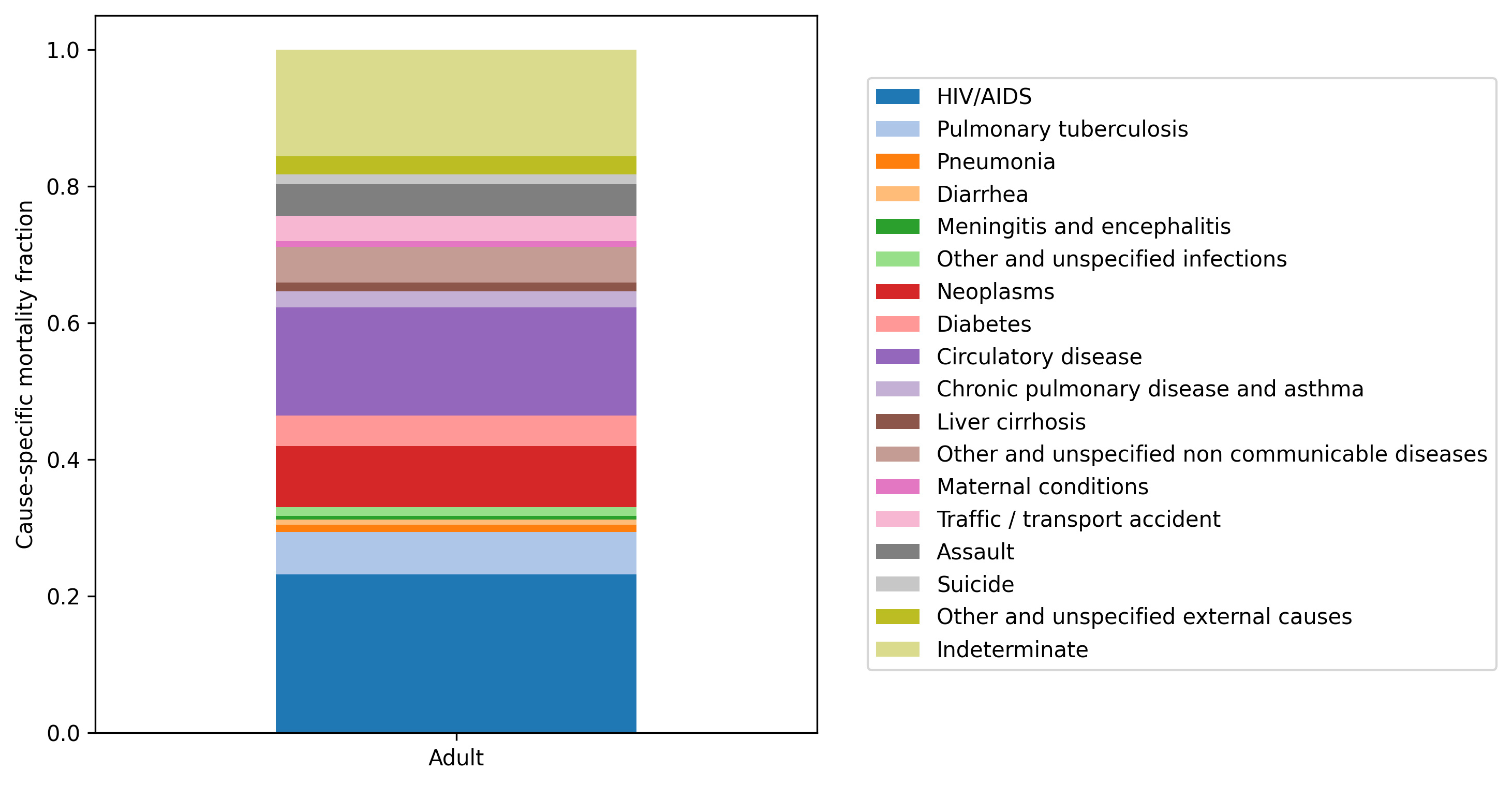}
\caption{Distribution of physician-coded cause of deaths for adults}
\label{fig:pcva_csmf_adult2}
\end{center}
\end{figure}

Distribution of PCVA by level 2 COD grouping is in Figure~\ref{fig:pcva_csmf_adult2}. HIV/TB was the leading COD accounting for 23.2\% of all adult deaths, followed by 15.8\% deaths due to circulatory diseases. 15.6\% of causes were labeled as indeterminate by physicians. 

\subsection{Performance of PLMs on COD classification using narratives alone}

\begin{table}[ht]
\centering
\captionsetup{format=plain,font=small,labelfont=bf, justification=justified, width=\textwidth}
\begin{tabulary}{\linewidth}{lCCCCCCC}
 \hline
Model           & Accuracy & Weighted F1 & Weighted Precision & Weighted Recall & CSMF Accuracy & CCCSMF Accuracy \\
 \hline 
BioClinicalBERT & \textbf{0.646}    & \textbf{0.638}       & \textbf{0.646}              & \textbf{0.646}           & 0.937         & 0.828           \\
BlueBERT        & 0.642    & 0.635       & \textbf{0.646}        & 0.642           & 0.942         & 0.842           \\
RoBERTa-PM         & 0.641    & 0.636       & 0.643              & 0.641           & 0.955         & 0.877           \\
BERT            & 0.631    & 0.623       & 0.618              & 0.631           & \textbf{0.958}         & \textbf{0.885}           \\
BioMedBERT      & 0.624    & 0.617       & 0.621              & 0.624           & 0.955         & 0.877           \\
ClinicalBERT    & 0.612    & 0.601       & 0.604              & 0.612           & 0.946         & 0.853           \\
InSilicoVA (base)     & 0.470    & 0.465       & 0.497              & 0.470           & 0.819         & 0.508           \\
 \hline
\end{tabulary}
\caption{Leaderboard: COD classification with narratives alone}
\vspace{-0.5em}
\caption*{\footnotesize \textit{Note:} The results is based on level 2 COD (18 categories). CSMF Accuracy: Cause-specific mortality fraction accuracy. CCCSMF Accuracy: Chance-corrected CSMF Accuracy.}
\label{tab:plm_leader}
\end{table}

Table \ref{tab:plm_leader} shows the performance leaderboard on classifying COD using VA narratives only. All PLMs achieved accuracy over 60\% at individual level, and over 90\% CSMF accuracy, which outperformed current question-only statistical model approach with InSilicoVA. Models pre-trained with both biomedical and clinical corpus (BioClinicalBERT, BlueBERT and RoBERTa-PM) showed highest performance at individual level, while base BERT showed slightly higher population level accuracy. 

\begin{figure}
\begin{center}
\includegraphics[width=6in]{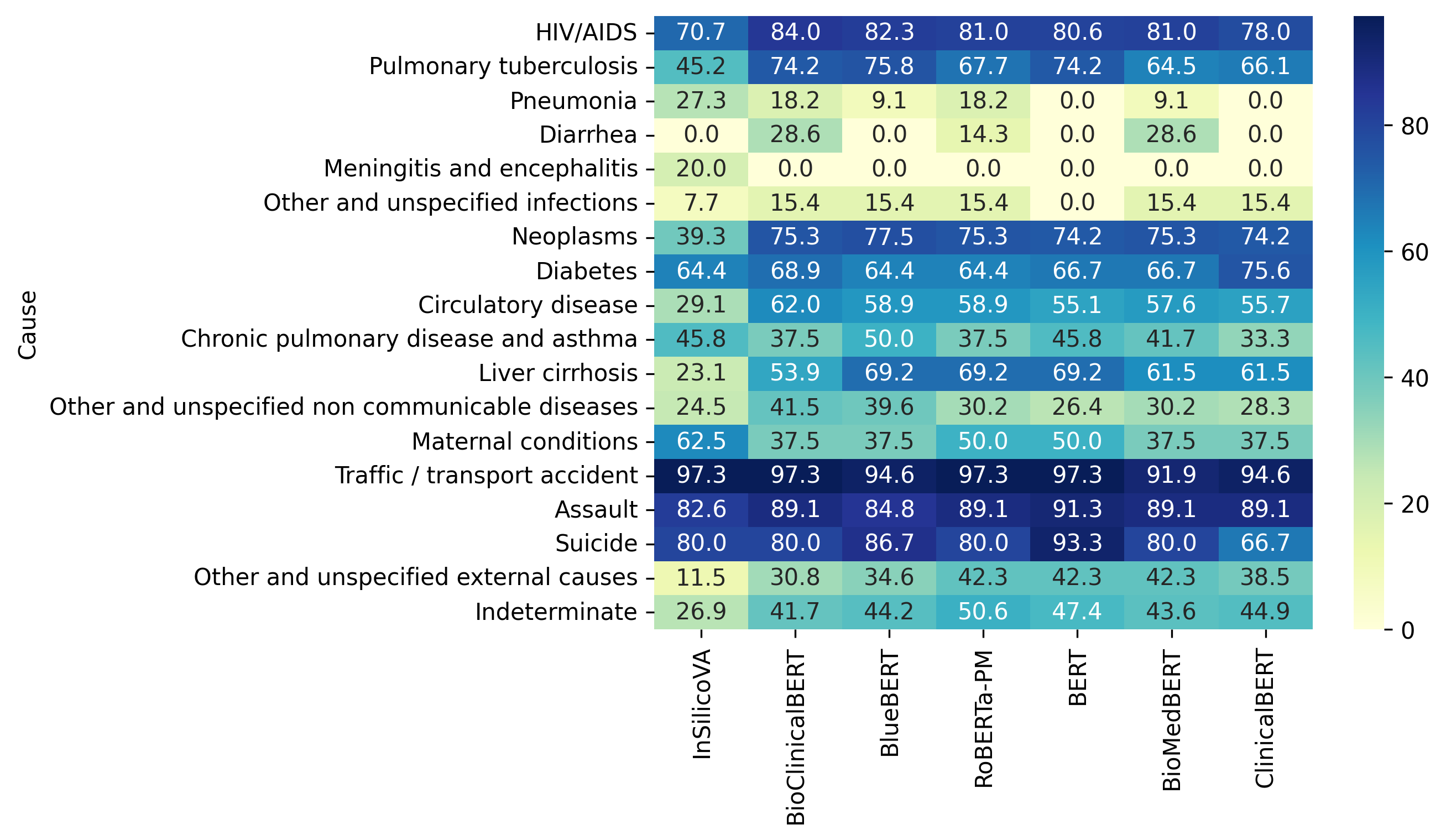}
\caption{Classification accuracy by cause: PLMs with narrative-alone}
\label{fig:plm_eval_heatmap}
\end{center}
\end{figure}

\begin{figure}
\begin{center}
\includegraphics[width=6in]{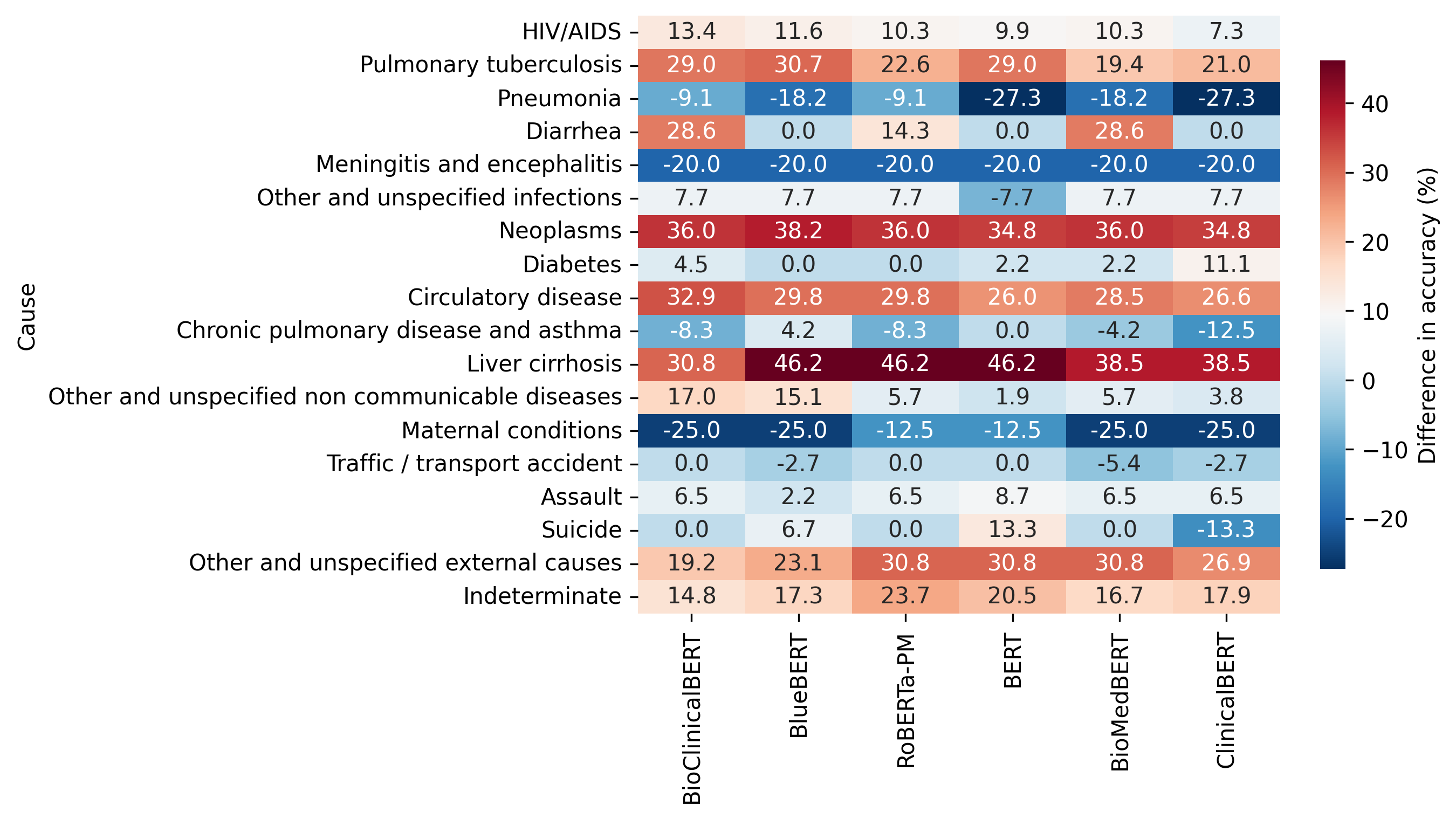}
\caption{Changes in accuracy relative to InSilicoVA as baseline}
\label{fig:plm_eval_heatmap_diff}
\end{center}
\end{figure}

Figure~\ref{fig:plm_eval_heatmap} illustrates the classification accuracy by cause across different models. While the overall performances for PLMs were fairly close based on aggregated evaluation metrics, the cause-level results revealed nuanced variations across models. For major causes (e.g., HIV/AIDS, neoplasms, traffic accidents, assault, and suicide), the accuracies were moderate to high across the board. Greater variations were observed, particularly for small communicable and non-communicable diseases, highlighting differences in model sensitivity to minority classes. Additional results on the confusion matrix comparing PLMs-based predictions to PCVA are in appendix~\ref{app:figs}. (section~\ref{fig:plm_confusion})

Figure~\ref{fig:plm_eval_heatmap_diff} illustrates the absolute change in accuracy relative to InSilicoVA as the baseline. Compared to current question-only statistical model, narrative-based models had significant improvement in identifying major non-communicable diseases like circulatory disease, neoplasms, and liver cirrhosis, and major communicable diseases like HIV and pulmonary tuberculosis (TB).

\subsection{Sensitivity analysis}

\subsubsection{Performance with different COD grouping}

We fine-tuned BioClinicalBERT to classify COD at different levels of categorization. (Table~\ref{tab:plm_sa_cod}) As the classification granularity increased, model accuracy declined, but not by much. With a 34 cause list, PLMs with narrative alone still achieved comparable performance at the individual level with 57.6\% accuracy, and higher CSMF Accuracy compared to previous publications with a 25 cause list (0.90 vs 0.84).~\cite{groenewald2023agreement} A closer look at confusion matrices suggested that most misclassifications at level 3 were due to the inability to distinguish HIV/TB and non-communicable diseases. (Figure~\ref{fig:plm_confusion_cod}) 

\begin{table}[ht]
\centering
\captionsetup{format=plain,font=small,labelfont=bf, justification=justified, width=\textwidth}
\begin{tabulary}{\linewidth}{lCcCCCCC}
 \hline
     \thead{\makecell{COD \\ grouping}}    & Categories & \thead{\makecell{Accu\\racy}}  & Weighted F1 & Weighted Precision & Weighted Recall & CSMF Accuracy & CCCSMF Accuracy \\
 \hline
Level 1 & 34          & 0.576    & 0.562       & 0.562              & 0.576           & 0.903         & 0.736           \\
Level 2 & 18         & 0.646    & 0.638       & 0.646              & 0.646           & 0.937         & 0.828           \\
Level 3 & 6         & 0.694    & 0.691       & 0.693              & 0.694           & 0.947         & 0.855          \\
 \hline
\end{tabulary}
\caption{Performance by COD grouping levels}
\label{tab:plm_sa_cod}
\end{table}

\subsubsection{Impact of training data size}

\begin{figure}
\begin{center}
\includegraphics[width=6in]{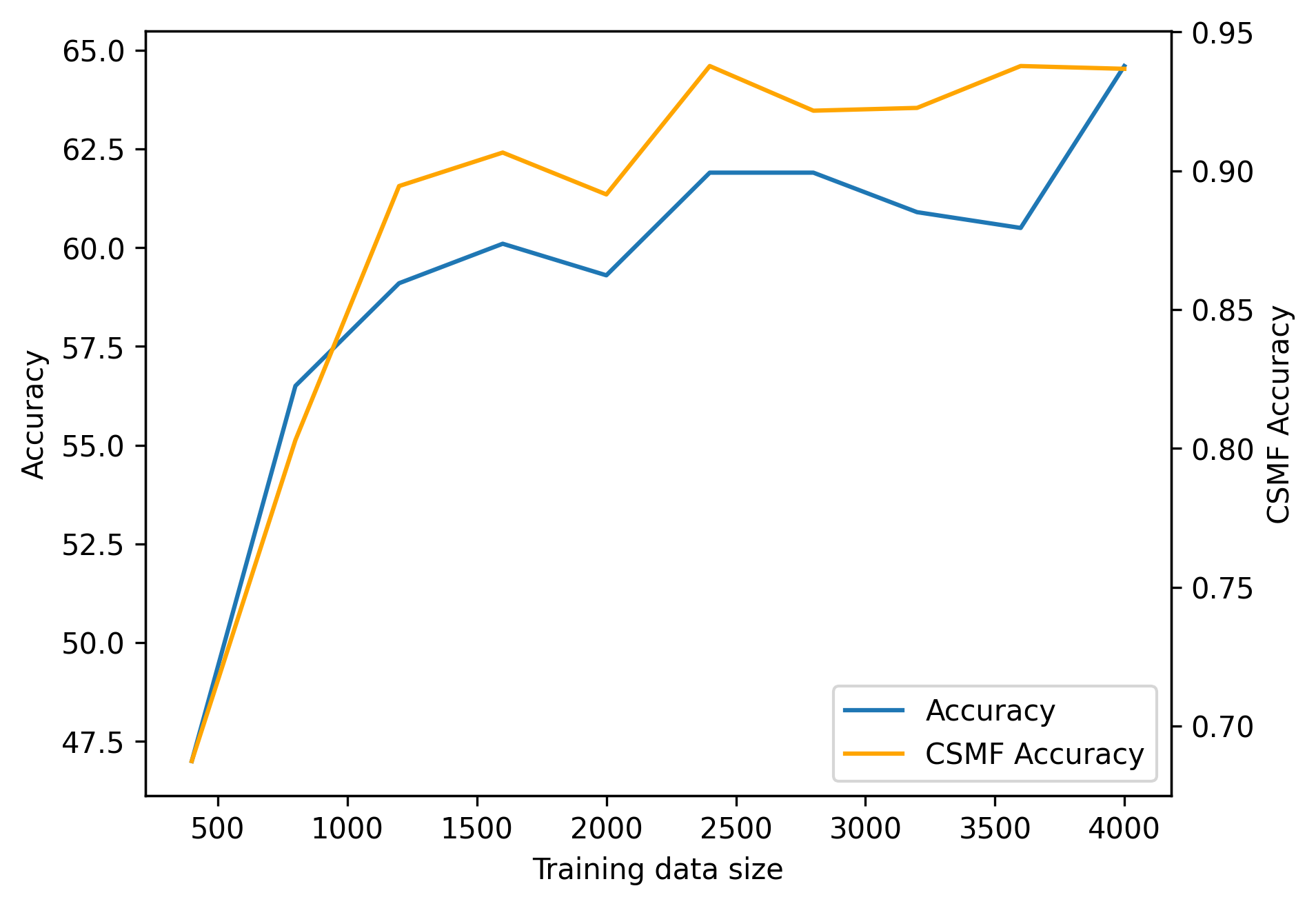}
\caption{Performance by training data size}
\label{fig:plm_sa_samplesize}
\end{center}
\end{figure}

Figure~\ref{fig:plm_sa_samplesize} demonstrates how accuracy and CSMF accuracy changes as the training data size for fine-tuning BioClinicalBERT. As training data size increased from 500 to 4000, the classification accuracy increased from 47.0\% to 64.6\%, the weighted F1 score increased from 39.7\% to 63.8\%, and CSMF Accuracy increased from 0.687 to 0.937. Model performance sharply increased before N=1500, and after having over 2000 deaths for training, increased at a much slower pace. 

Looking at trends in cause-specific accuracy over training data size, as the training size increases, the classification accuracy in the major causes had minimal improvement. The gain in overall accuracy was mainly driven by the ability for picking up minority classes as the sample size increased.(Figure~\ref{fig:plm_sa_samplesize_heatmap})

\subsubsection{Impact of evaluation metrics on hyperparameters}

We tested using cross-entropy loss instead of mean weighted F1 as the metric to select the optimal hyperparameters configuring the final PLMs, The performance changed slightly - with accuracy and weighted F1 score decreasing by about 2\% and CSMF accuracy increase by 1\%. (Table~\ref{tab:plm_sa_loss}) This indicates that depending on the primary interest of the analysis - whether it is to assign the death certificate to individuals, or to track population health status - one needs to tailor the PLMs fine-tuning pipeline to best fit the practical needs. 

\begin{table}[ht]
\centering
\captionsetup{format=plain,font=small,labelfont=bf, justification=justified, width=\textwidth}
\begin{tabulary}{\linewidth}{lCCCCCCC}
 \hline
Metrics     & Accuracy & Weighted F1 & Weighted Precision & Weighted Recall & CSMF Accuracy & CCCSMF Accuracy \\
 \hline 
Weighted F1 & \textbf{0.646}    & \textbf{0.638}       & \textbf{0.646}              & \textbf{0.646}           & 0.937         & 0.828           \\
Cross-entropy loss    & 0.625    & 0.620       & 0.623              & 0.625           & \textbf{0.947}         & \textbf{0.855}           \\
 \hline
\end{tabulary}
\caption{Performance by evaluation metrics used for HPO}
\label{tab:plm_sa_loss}
\end{table}

\section{Discussion and conclusion}

We found that with narratives alone, transformer-based PLMs can classify COD with up to 64.6\% individual-level accuracy, and the population-level CSMF accuracy as high as 0.96, based on empirical evidence from South Africa. 

PLMs showed significantly better performance with the majority classes such as HIV/AIDS and circulatory diseases, while minority classes, especially pneumonia, diarrhea, and meningitis, were most likely to be misclassified. This issue was particularly pronounced for models with less domain context (like base BERT or clinical BERT), indicating that rich domain context might benefit by making the models more sensitive to subtle contextual signals in narratives. Moreover, differential diagnosis between infectious diseases usually requires very specific and nuanced details about onset, pattern, and progression of events, which might not be sufficiently reported by respondents in narratives. Last but not least, as also revealed in the sensitivity analysis, the performances of PLMs on minority classes were affected by the representation of minority classes in the sample. Future research could look at text balancing techniques, such as synthetic text generation, to augment minority class representation, and to further enhance classification performance. 

Compared to other studies on COD classification with VA narratives, transformer-based PLMs achieved among the highest population-level accuracy, but still had room for improvement for individual accuracy. A previous study \cite{Jeblee2021} reported that the ELMo + neural network can be further augmented by adding explicitly extracted timing and event sequence features. Transformer-based deep contextual embeddings can partially capture information like event sequences with position embeddings and self-attention mechanism. Still, those embeddings can not explicitly reason about temporal structure or event chronology. Future studies could explore whether explicitly extracting and incorporating known key features such as timing and sequence of events, could further boost classification performance in the transformer-based PLMs pipeline. 

PLMs with additional domain-specific pre-training performed slightly better than base BERT, although the differences were not substantial. Base BERT showed comparable overall performances and even slightly better performances in identifying assault and suicide. This is not surprising given that diseases were often described in a much less "clinical" way in the context of VA, e.g. type-II diabetes is mostly called "sugar diabetes" in the narratives, while blood glucose or insulin resistance were almost never mentioned. General PLMs might also be able to pick up these contexts through general domain pre-training and/or task-specific fine-tuning, where biomedical/clinical-domain knowledge might not always transfer well to the context of VA. 

On the other hand, it is also worth noting that ClinicalBERT, pre-trained with EHR from diabetes patients, had outstanding performance in identifying diabetes compared to all other models. These findings implies that proper domain-specific pre-training can still be valuable. Future research could investigate the potential of PLMs specifically pre-trained on VA data, if sufficient high-quality VA narratives become available in the future. 

In conclusion, PLMs using narrative alone outperformed current mainstream VA algorithm using structured questions alone, for both individual- and population-level measures. And the accuracies were notably higher for non-communicable diseases such as neoplasms, circulatory disease, liver cirrhosis, compared to InSilicoVA. This not only confirms our hypothesis that narratives alone contain rich information valuable for determining COD, and can be utilized by fine-tuned PLMs, but also implies that narratives might have captured information critical for diagnosis that has not been fully captured in structured questions. This inspires us to explore the multimodal learning approaches to combining knowledge from both narratives and questions in automated COD classification using VA data, which will be introduced in detail in the next chapter.

   % narrative-only
\chapter{Combining knowledge from unstructured narratives and structured questions with multimodal fusion approach}
\label{multimodal.ch}

\section{Introduction}

This chapter investigates COD classification using multimodal learning approaches. We first set the stage by briefly reviewing related works of information fusion in the context of VA. Then, we describe in detail the multimodal fusion strategies proposed and investigated in this study. Lastly, we evaluate multimodal COD classification with empirical VA data from South Africa.

\subsection{Verbal autopsy as multimodal data}

In reality, data in the medical and health domain usually takes on multiple forms, or modalities, from structured tabular data from survey questions or lab results, to unstructured data such as narratives, images, and audio. VA is inherently a multimodal data source, as it combines narrative (text) with structured responses to questions (tabular) about signs, symptoms, and demographic information.

Despite the multimodal nature of VA, traditional algorithms have been developed and validated only using the structured question responses. Few studies to date have explored the use of both modalities in the automated COD classification pipeline using VA, with most of which adopted a feature-fusion approach. Mapundu et al.~\cite{Mapundu2022} used traditional NLP to extract and vectorize text features (n-grams), concatenated to the question responses, and used ML models for downstream classification. Manaka et al. 2022 \cite{Manaka2022a} and Blanco et al. 2021 \cite{Blanco2021} extracted paragraph embeddings or word embeddings from narratives, and then used ML and/or neural networks for classification. Both studies reported improved performance with combined features, with XGBoost (or XGBoost ensemble) as the best performing classifier. Cejudo et al. 2023~\cite{cejudo2023cause} used a stacking ensemble for prediction-level fusion, using logistic regression to combine dual input from the best performing unimodal predictions (XGBoost with the questions and BERT with the narrative). Ensemble learning combining both modalities also showed improved performance. 

\subsection{Gaps in existing research}

Although all studies have reported better classification performance when combining narratives with questions, they only mark the start of the conversation. The validity of some results seems questionable. Two studies reported an individual-level accuracy of 96\% \cite{Mapundu2022} and 99\% \cite{Manaka2022a} for the multimodal approach and 99\% for the unimodal approach, which seemed unrealistically high given the context and the existing body of research. In the other two studies, the multimodal approach at best achieved ~50\% accuracy for adult (51.6\% with \cite{cejudo2023cause} and 50.0\% with \cite{Blanco2021} respectively), indicating the potential room for improvement. Moreover, no study has thus far tested and compared different multimodal approaches in the context of VA, such as data-level fusion or more complex ensemble learning frameworks. And previous studies, mainly from the field of computer science, have primarily focused on advancing model architectures and methodologies. They paid little attention to examining the specific contribution of each modality to the final classification in this particular context - the insight of which could be valuable for improving VA data collection and interpretation. 

Therefore, this study aims to comprehensively investigate various multimodal fusion strategies in the context of VA, leveraging SOTA PLMs and ML techniques. The following section describes in detail the proposed multimodal learning frameworks, fusing knowledge from different modalities at different stages. Then we apply and compare different unimodal and multimodal approaches using empirical data in South Africa.

\section{Multimodal fusion strategies}

\begin{figure}
\begin{center}
\includegraphics[width=6in]{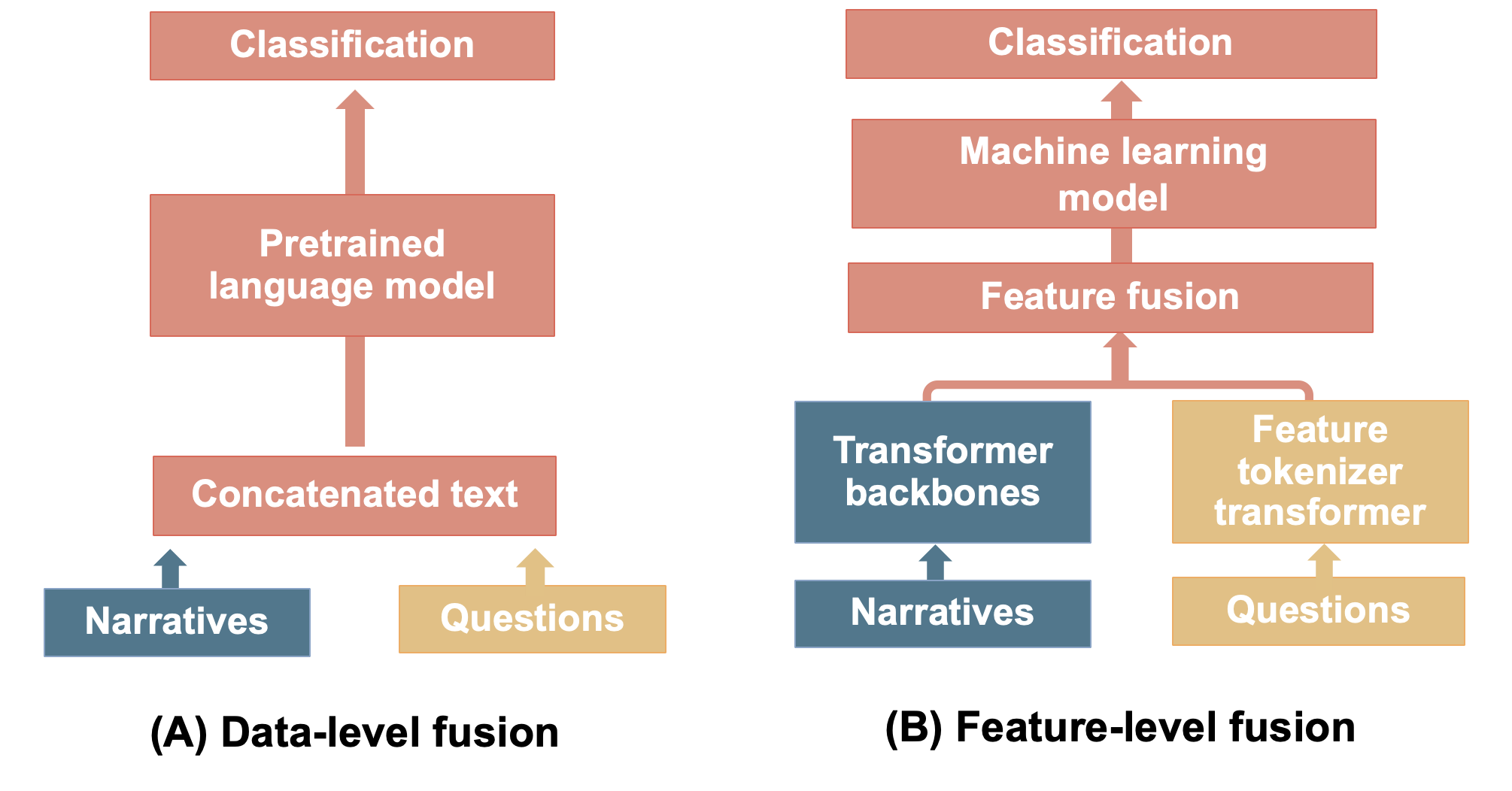}
\includegraphics[width=6in]{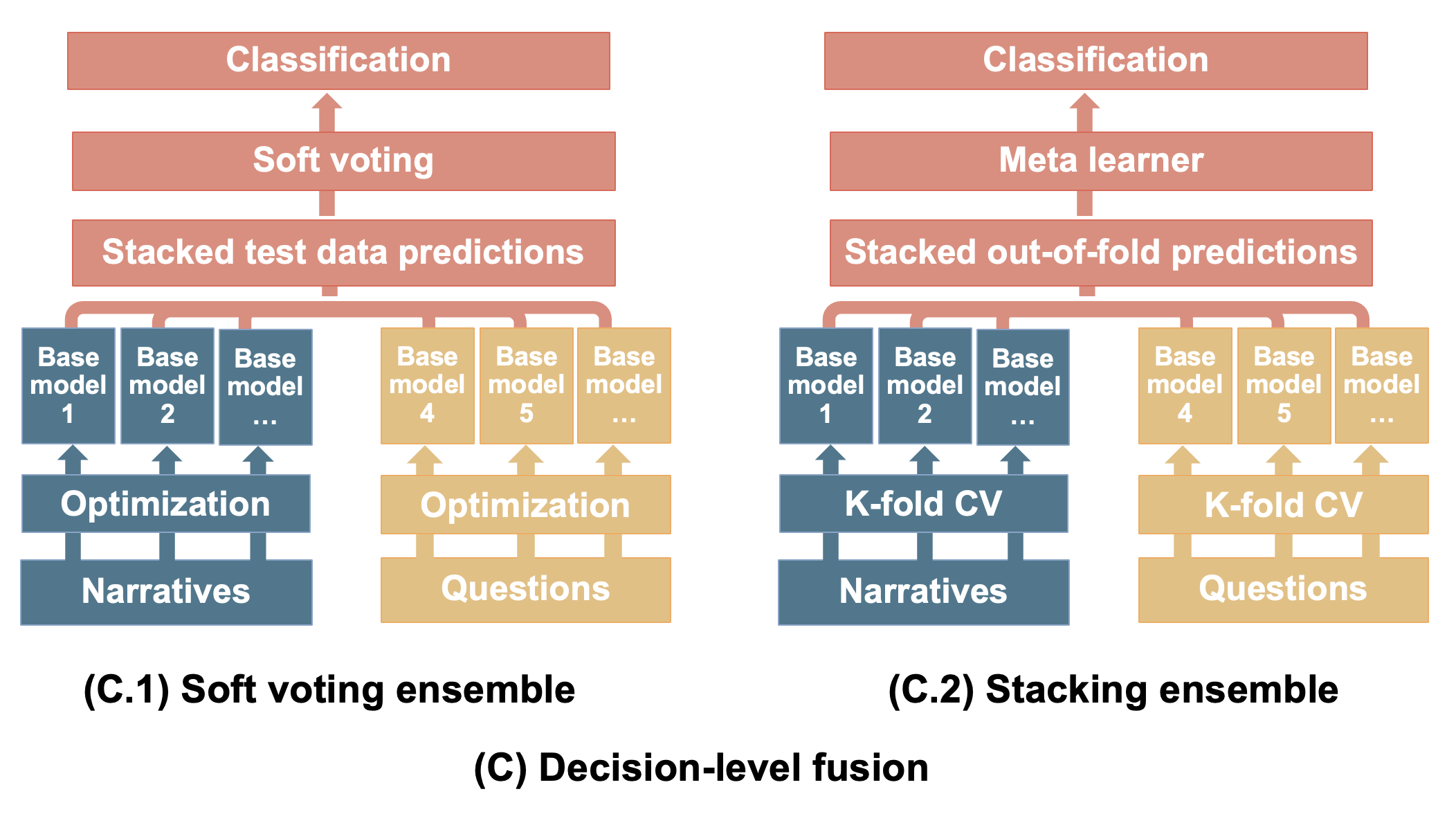}
\caption{Overview of multimodal fusion strategies}
\label{fig:mm_strategies}
\end{center}
\end{figure}

The basic structure of each fusion strategy has been introduced in the previous chapter (Chapter~\ref{review.ch} Section~\ref{info:mm}), thus in this section, we focus on the actual implementation of these strategies as applied in this study using VA data. Figure~\ref{fig:mm_strategies} provides a high-level overview of the proposed multimodal fusion strategies. 

\subsection{Data-level fusion}

The all-text model is the most straightforward and intuitive approach, particularly suitable for text-tabular data like VA. We took a context-rich approach when converting question responses to text. For all indicators reported non-missing values in the structured questions of VA, we converted them into complete sentences incorporating variable descriptions of the signs, symptoms, and demographics. A brief example of the conversion is demonstrated in Table~\ref{tab:qdesc_example}. Detailed conversion strategy is provided in the Appendix~\ref{app1.qdesc}.

\begin{table}[ht]
\captionsetup{format=plain,font=small,labelfont=bf, justification=justified, width=\textwidth}
\centering
\begin{tabularx}{\textwidth}{|p{0.1\textwidth}|p{0.25\textwidth}|p{0.1\textwidth}|X|}
\hline
Variable & Question description & Value & \multicolumn{1}{l|}{Converted text} \\
\hline
i019a       & Was he male?                                                     & No     & \multirow{6}{=}{The deceased was a female. The deceased had fever. The deceased had fever less than one week before death. The deceased did not have chest pain.} \\
i019b       & Was she female?                                                  & Yes    &                                                                                                                                                                   \\
i022c       & Was s(he) aged 15 to 49 years at death?                          & Yes     &                                                                                                                                                                   \\
i147o       & During the illness that led to death, did (s)he have a fever?    & Yes     &                                                                                                                                                                   \\
i148a       & Did the fever last less than a week before death?                & No    &                                                                                                                                                                   \\
i174o       & During the illness that led to death, did (s)he have chest pain? & No     &                  \\
  \hline
\end{tabularx}
\caption{Converting question responses to text: an example}
\label{tab:qdesc_example}
\end{table}

Then we concatenated the raw narratives with the text converted from the question responses, used the combined text as input to fine-tune long-sequence PLMs, such as Clinical Longformer~\cite{li2022clinical} and Longformer~\cite{beltagy2020longformer}, for COD classification. We followed the same methods for stratified train-test split, HPO, model training, and evaluation as described in the previous chapter. We used Optuna with the same search space (Table~\ref{tab:plmhpo_space}), running for 10 trials with pruning to find the best set of hyperparameters for each model. The final hyperparameters used to fit the final model are in Table~\ref{tab:lfhpo_best}.

\begin{table}[ht]
\centering
\captionsetup{format=plain,font=small,labelfont=bf, justification=justified, width=\textwidth}
\begin{tabulary}{\linewidth}{LCC}
 \hline
Hyperparameter              & Clinical Longformer & Longformer \\
  \hline
Learning rate               & 3.76E-05            & 3.76E-05   \\
Batch size                  & 8                   & 8          \\
Number of training epochs   & 7                   & 7          \\
Weight decay                & 1.40E-04            & 1.40E-04   \\
Warmup ratio                & 0.06                & 0.06       \\
Gradient accumulation steps & 1                   & 1          \\
  \hline
\end{tabulary}
\caption{Best set of hyperparameters for selected long-sequence PLMs}
\label{tab:lfhpo_best}
\end{table}

\subsection{Feature-level fusion}

For feature-level fusion, we applied a customized AutoGluon Multimodal model (AutoMM) released by Amazon. The design of the model has been published elsewhere~\cite{Tang2024} and reviewed at high level in the previous chapter (section \ref{info:automm}). In brief, AutoMM uses modality-specific models to extract embeddings from the text (narratives) and the tabular (questions) data independently. Concatenated features were then passed to a multi-layer MLP, and the final underlying COD were predicted via a softmax output layer. We customized PLMs (e.g. BioClinicalBERT, RoBERTa-PM) as the backbone text model, and kept the default feature tokenizer transformer (ft-transformer) for tabular data. As an automated machine learning toolkit by design, the model automatically optimizes the hyperparameters for each model in the process, with preset tuning configurations. 

\subsection{Prediction-level fusion}

%\paragraph{Soft voting ensemble} 
\subsubsection{Soft voting ensemble} 

In addition to the top-performing narrative-only PLMs and InSilicoVA trained in chapter~\ref{plmva.ch}, we additionally fit popular machine learning models as question-only based models. 
\begin{itemize}
\item Narrative-only models: BioClinicalBERT, BlueBERT, BioMedBERT, BERT, and RoBERTa-PM
\item Question-only models: LightGBM,\cite{ke2017lightgbm} CatBoost,\cite{dorogush2018catboost} Gradient Boosting Decision Tree (GBDT),\cite{friedman2001greedy} XGBoost,\cite{chen2016xgboost} and MLP \cite{pedregosa2011scikit}
\end{itemize}

Each model was configured with hyperparameters optimized using Optuna. The search spaces for hyperparameters for each model are in Table~\ref{tab:mlhpo_space}. After obtaining the final predicted probabilities for the hold-out testing data for all models, we performed a soft voting ensemble to get the final predicted label. (Algorithm \ref{alg:voting})

\begin{algorithm}[H]
\caption{Training strategy: soft voting ensemble}
\label{alg:voting}
\begin{algorithmic}[1] 
\REQUIRE Text data $T$, tabular data $X$, labels $Y$, model set $M$, COD set $C$
\STATE \label{alg1:l1} Stratified split of data to training data $(T^{train},X^{train}, Y^{train})$ and testing data $(T^{test},X^{test}, Y^{test})$
\FOR{Each model $m \in M$}
\STATE Train model $m$ on appropriate modality:
\[
m \leftarrow \text{fit}(m, (T^{train}, Y^{train}) \quad \text{or} \quad (X^{train}, Y^{train}))
\]
\STATE Generate prediction probabilities $\hat{P}_{m,C}$ on testing data for all causes $C$:
\[
\hat{P}_{m,C} \leftarrow m(T^{val}) \quad \text{or} \quad m(X^{val})
\]
\ENDFOR
\FOR{Each cause $c \in C$}
\STATE \label{alg1:l5} Calculate soft voted probability: 
\[
\bar{P}_{c}=\frac{1}{M} \sum_{m=1}^{M} P_{c,m}
\]
\ENDFOR
\STATE \label{alg1:l6} Predict final cause with the highest average probability: 
\[
\hat{Y} = \arg\max_{c} \bar{P}_{c}
\]
\end{algorithmic}
\end{algorithm}

\begin{table}[ht]
\centering
\captionsetup{format=plain,font=small,labelfont=bf, justification=justified, width=\textwidth}
\begin{tabulary}{\linewidth}{lll}
 \hline
Hyperparameter & Search strategy & Search space \\ 
  \hline
\textbf{CatBoost}       &                          &                               \\
iterations              & Categorical              & {[}100,150,200{]}             \\
learning\_rate          & Log-uniform              & {[}0.01,0.2{]}                \\
depth                   & Integer                  & {[}3, 9{]}                    \\
boosting\_type          & Categorical              & {[}'Ordered', 'Plain'{]}      \\
l2\_leaf\_reg           & Integer                  & {[}2, 4{]}                    \\
  \hline
\textbf{LightGBM}       &                          &                               \\
learning\_rate          & Log-uniform              & {[}0.01,0.2{]}                \\
n\_estimators           & Categorical              & {[}50,100,200{]}              \\
max\_depth              & Categorical              & {[}-1,3,5,7{]}                \\
num\_leaves             & Categorical              & {[}25,50,100{]}               \\
min\_data\_in\_leaf     & Categorical              & {[}1,5,10{]}                  \\
  \hline
\textbf{GBDT}           &                          &                               \\
learning\_rate          & Log-uniform              & {[}0.01,0.2{]}                \\
n\_estimators           & Categorical              & {[}50,100,150,200{]}          \\
max\_depth              & Categorical              & {[}1,3,5,7{]}                 \\
min\_samples\_split     & Categorical              & {[}2,5,10{]}                  \\
min\_samples\_leaf      & Categorical              & {[}1,2,4{]}                   \\
subsample               & Categorical              & {[}0.8,1.0{]}                 \\
  \hline
\textbf{XGB}            &                          &                               \\
learning\_rate          & Log-uniform              & {[}0.01,0.1{]}                \\
n\_estimators           & Categorical              & {[}50,100,150,200{]}          \\
max\_depth              & Categorical              & {[}1,3,5,7{]}                 \\
booster                 & Categorical              & {[}'gbtree','dart'{]}         \\
gamma                   & Categorical              & {[}0,0.05,0.1,0.2{]}          \\
  \hline
\textbf{MLP}            &                          &                               \\
hidden\_layer\_sizes    & Categorical              & {[}(50,50),(100,),(100,50){]} \\
activation              & Categorical              & {[}'tanh','relu'{]}           \\
solver                  & Categorical              & {[}'adam','sgd'{]}            \\
alpha                   & Categorical              & {[}0.0001,0.001,0.01{]}       \\
learning\_rate          & Categorical              & {[}'constant','adaptive'{]}   \\
max\_iter               & Integer                  & {[}800,3000, step=200{]}      \\
  \hline
\textbf{Random forest}  &                          &                               \\
n\_estimators           & Categorical              & {[}50,100,150,200{]}          \\
max\_depth              & Categorical              & {[}5,10,20,None{]}            \\
min\_samples\_split     & Categorical              & {[}2,5,10{]}                  \\
max\_features           & Categorical              & {[}'auto','sqrt','log2'{]}    \\
min\_samples\_leaf      & Categorical              & {[}1,2,4{]}                  \\
  \hline
\end{tabulary}
\caption{Hyperparameters optimization for ML models}
\label{tab:mlhpo_space}
\end{table}

\subsubsection{Stacking ensemble}

For the stacking ensemble, we essentially adopted a modified Super Learner framework \cite{van2007super}. (Algorithm \ref{alg:stacking})

For each base model, we performed 5-fold stratified cross-validation to generate out-of-fold (OOF) predictions, which were later used to train the meta-learner. Specifically, we split the training data into five folds stratified by COD. In each iteration, the model was trained on four folds to predict the remaining (held-out 5th) fold. Stacking OOF predicted probabilities from all folds, we obtained a complete set of model predictions for that base model. Concatenating those stacked predictions from all base models, we got the input features for the meta-learner. We then trained the meta-learner to combine base model predictions for the final prediction. 

To get predictions for the hold-out testing dataset, we first trained each base model on the full training dataset and obtained the predicted probabilities on the testing data. Then these base-model predictions were stacked and passed to the trained meta-learner to produce the final ensemble predictions for the testing data for model evaluation. 

In the stacking ensemble approach, we used the same set of base models as in the soft voting ensemble approach. For candidate models of meta-learner, standard classifiers were tested, such as logistic regression, random forest, SVM, LightGBM, and K nearest-neighbors (KNN), etc.. 

\begin{algorithm}[H]
\caption{Training strategy: stacking ensemble}
\label{alg:stacking}
\begin{algorithmic}[1] 
\REQUIRE Text data $T$, tabular data $X$, labels $Y$, base model set $M$, meta learner $Meta$, number of folds $k$
\STATE Stratified split of data into training set $(T^{train}, X^{train}, Y^{train})$ and test set $(T^{test}, X^{test}, Y^{test})$
\FOR{each base model $m \in M$}
\STATE Initialize out-of-fold predictions: $\hat{P}_m \leftarrow \emptyset$
\FOR{$i = 1$ to $k$ for $k$-fold stratified cross-validation}
\STATE Split $(T^{train}, X^{train}, Y^{train})$ into training folds $(T^{train}_{-i}, X^{train}_{-i}, Y^{train}_{-i})$ and validation fold $(T^{val}_{i}, X^{val}_{i}, Y^{val}_{i})$
\STATE Train model $m$ on appropriate modality:
\[
m_i \leftarrow \text{fit}(m, (T^{train}_{-i}, Y^{train}_{-i}) \quad \text{or} \quad (X^{train}_{-i}, Y^{train}_{-i}))
\]
\STATE Generate prediction probabilities $\hat{P}_{m,i}$ on validation fold:
\[
\hat{P}_{m,i} \leftarrow m_i(T^{val}_{i}) \quad \text{or} \quad m_i(X^{val}_{i})
\]
\STATE Append $\hat{P}_{m,i}$ to $\hat{P}_m$
\ENDFOR
\ENDFOR
\STATE Concatenate all base model predictions: $X^{meta} \leftarrow \text{hstack}(\hat{P}_{m_1}, \hat{P}_{m_2}, ..., \hat{P}_{m_{|M|}})$
\STATE Train meta learner: $Meta \leftarrow \text{train}(Meta, (X^{meta}, Y^{train}))$
\STATE Generate final predictions on test set:
\[
\hat{P}_{test}^{m} \leftarrow m(T^{test}) \quad \text{or} \quad m(X^{test}) \quad \forall m \in M
\]
\[
X^{meta}_{test} \leftarrow \text{hstack}(\hat{P}_{test}^{m_1}, ..., \hat{P}_{test}^{m_{|M|}})
\]
\[
\hat{Y}_{test} \leftarrow Meta(X^{meta}_{test})
\]
\RETURN Final predictions $\hat{Y}_{test}$
\end{algorithmic}
\end{algorithm}

\section{Methods}

\subsection{Data}

In this study, we used complete VA data including both unstructured narratives and structured questions in SANCOD. For data from structured questions, we used the same VA indicators as the input used to train InSilicoVA.~\cite{li2023openva} 

We used the same stratified training-test splits as in the narrative-only approach to ensure comparability across all models. 

\subsection{Model evaluation}

Same as in the previous chapter, we used the following metrics to evaluate model performance:

\begin{itemize}
\item Individual level metrics (based on top cause): accuracy, weighted F1 score, weighted precision, weighted recall;
\item Population level metrics (based on full probability distributions): CSMF accuracy, CCCSMF accuracy
\end{itemize}

\subsection{Sensitivity analysis}

Similar to the previous chapter with narrative-only models, we also ran sensitivity analysis with the multimodal approaches, to test the impact of the following factors on the model performances: 

\begin{itemize}
%\item COD grouping: we ran the models end-to-end with different levels of COD grouping (level 1, 2, and 3), with model-specific hyperparameter optimization. We then evaluated how model performance changes with changing the granularity of cause grouping.
\item training data size: we reran the models with 10\%, 20\%, 30\%, ... 90\% of the training dataset to fine-tune the Longformer model used in data-level fusion, with the subset stratified sampled from the full training set. We fine-tuned the PLMs using the same hyperparameter configuration as the main model, and then evaluated their performances on the same hold-out testing set.
\item model selection: we compared ensembled results using different subsets of base models in soft voting, to explore the impact of base model selection on final results in the soft voting ensemble approach.
\end{itemize}

The analysis was done in Python 3.12 \cite{python312}, using the Transformer package \cite{wolf2020transformers} for training PLMs and the scikit-learn package \cite{pedregosa2011scikit} for training machine learning models.

\section{Results}

\subsection{Unimodal approach: question-only vs narrative-only}

\begin{figure}
\begin{center}
\includegraphics[width=6in]{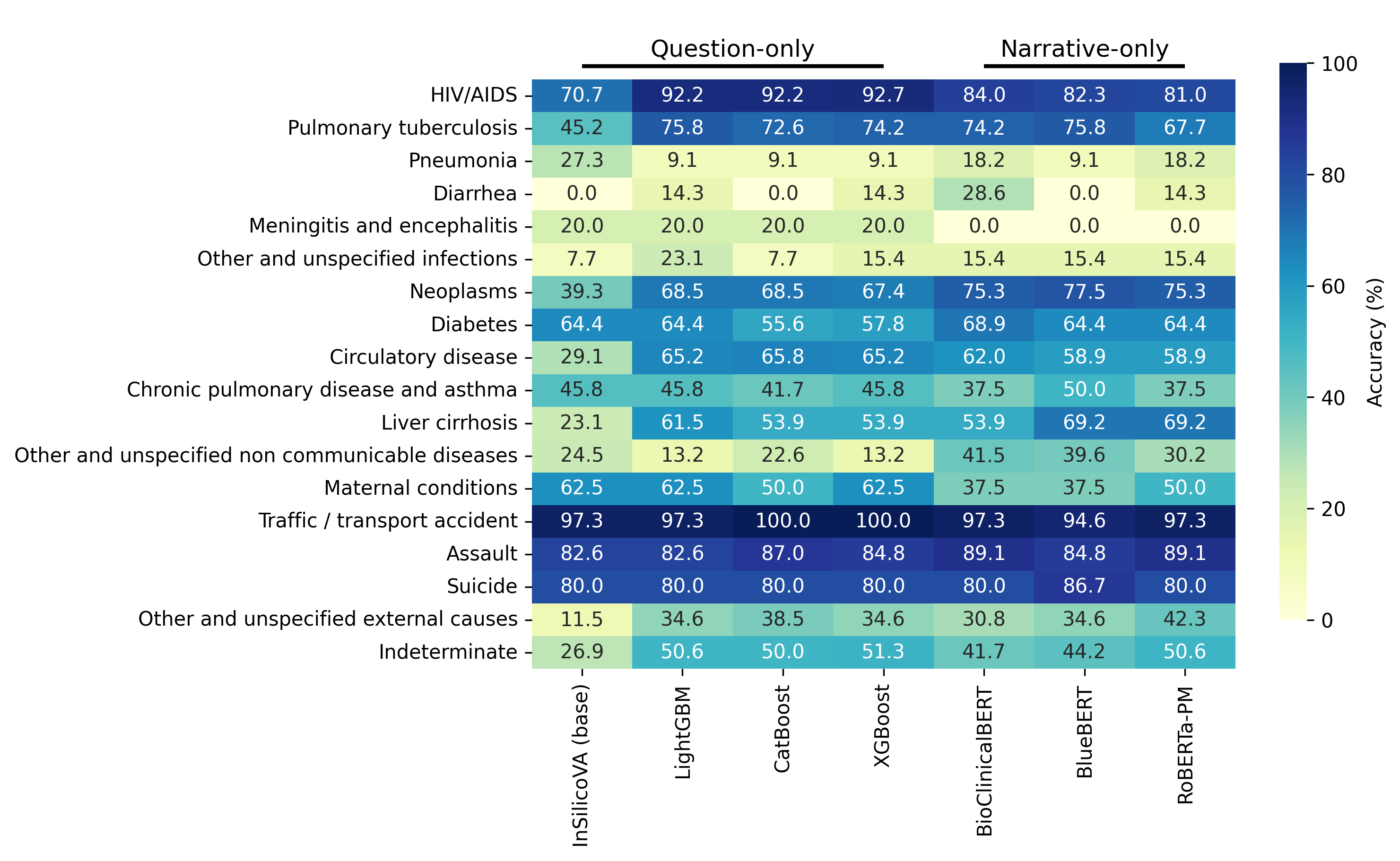}
\caption{Classification accuracy by cause: question-alone vs narrative-alone}
\label{fig:heatmap_unimodal}
\end{center}
\end{figure}

\begin{figure}
\begin{center}
\includegraphics[width=6in]{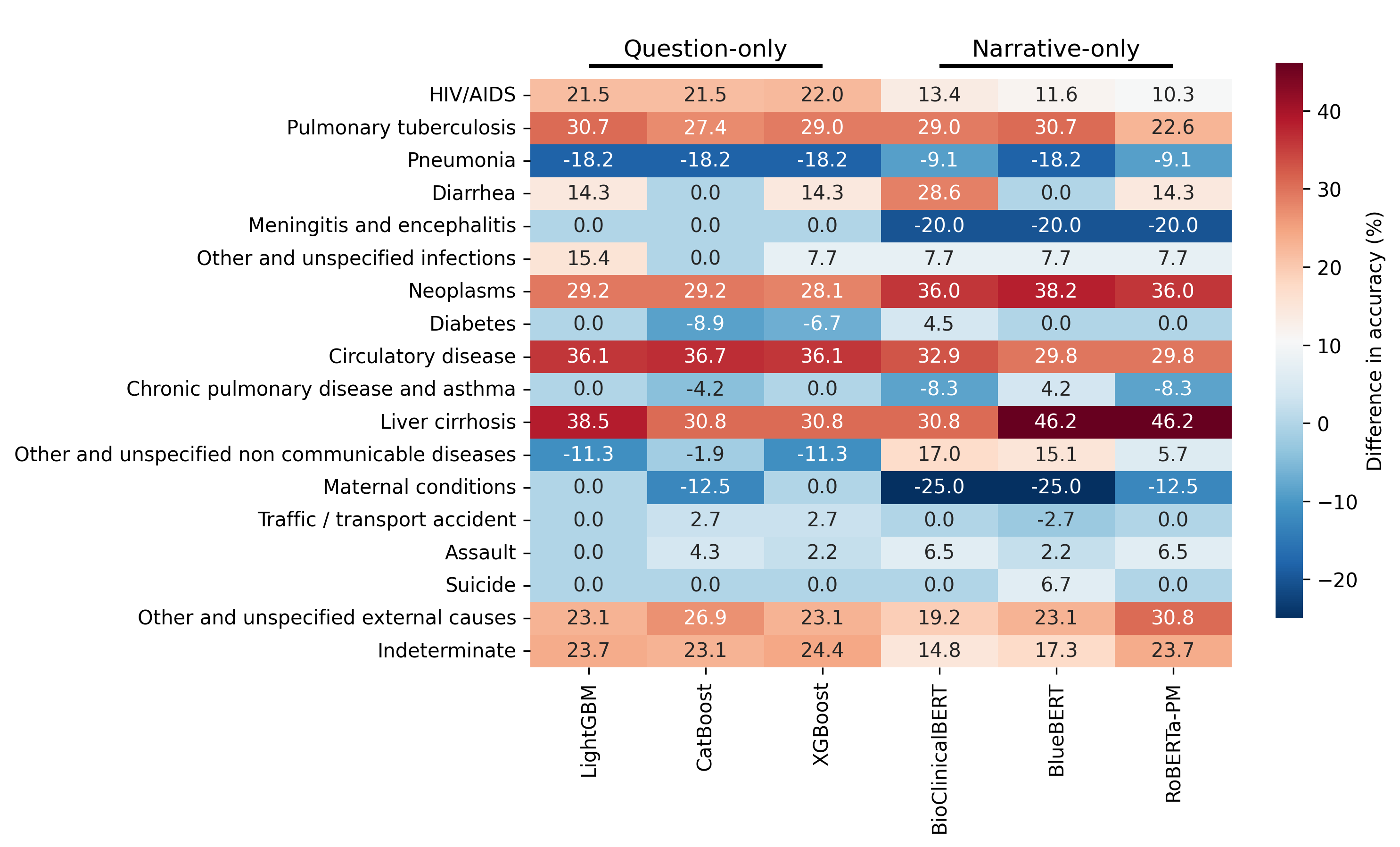}
\caption{Change in accuracy relative to InSilicoVA as baseline: unimodal models}
\label{fig:heatmap_unimodal_diff}
\end{center}
\end{figure}

We first take a closer look at the performance of the unimodal approaches. Figure~\ref{fig:heatmap_unimodal} illustrates the classification accuracy by cause for top-performing models using question-only and narrative-only approaches. Both modalities showed high accuracy with external causes. The question-based approach with ML models showed better performances, especially with HIV/AIDS, and was more stable with minority communicable diseases. The narrative-based approach with PLMs showed higher accuracy with neoplasms, diabetes, and liver cirrhosis. Compared to the current statistical model like InSilicoVA, ML with questions alone showed significant improvement in classifying major communicable and non-communicable diseases, while being less accurate in identifying causes such as pneumonia. (Figure~\ref{fig:heatmap_unimodal_diff})

\subsection{Multimodal approach}

Table \ref{tab:mm_cod} shows the model performance leaderboard on COD classification from all proposed approaches: unimodal with either narratives or questions alone, and multimodal with four proposed strategies. Most multimodal fusion strategies further increased the predicted accuracy compared to the unimodal approaches. Both decision-level fusion and data-level fusion outperformed top question-only models at both the individual and population levels. Performances from more models are in the appendix. (Appendix~\ref{app1:mm_cod_all})

\begin{landscape}

\begin{table}[ht]
\centering
\captionsetup{format=plain,font=small,labelfont=bf, justification=justified, width=\textwidth}
\begin{tabulary}{\linewidth}{llcccccc}
 \hline
Modality   & Model   & Accuracy & \thead{\makecell{Weighted \\ F1}} & \thead{\makecell{Weighted \\ Precision}} & \thead{\makecell{Weighted \\ Recall}} & \thead{\makecell{CSMF \\ Accuracy}} & \thead{\makecell{CCCSMF \\ Accuracy}} \\
 \hline
Question   & InSilicoVA (base)                                      & 0.470    & 0.465       & 0.497              & 0.470           & 0.819         & 0.508          \\
Narrative  & BioClinicalBERT                                        & 0.646    & 0.638       & 0.646              & 0.646           & 0.937         & 0.828           \\
%Narrative  & BlueBERT                                               & 0.642    & 0.635       & 0.646              & 0.642           & 0.942         & 0.842           \\
Question   & LightGBM                                               & 0.665    & 0.654       & 0.665              & 0.665           & 0.899         & 0.727           \\
%Question   & XGBoost                                                & 0.662    & 0.651       & 0.659              & 0.662           & 0.898         & 0.724           \\
Multimodal & A - Data-level fusion$^*$                     & 0.683    & 0.669       & 0.666              & 0.683           & 0.902         & 0.732           \\
Multimodal & B - Feature-level fusion                               & 0.646    & 0.640       & 0.647              & 0.646           & 0.933         & 0.817           \\
Multimodal & C.1 - Soft voting               & \textbf{0.695}    & \textbf{0.681}       & \textbf{0.680}              & \textbf{0.695}           & 0.923         & 0.790           \\
Multimodal & C.2 - Stacking ensemble$^\dag$  & 0.674    & 0.666       & 0.665              & 0.674           & \textbf{0.950}         & \textbf{0.863}           \\
%Multimodal & C - Decision-level fusion (Stacking ensemble with SVM) & 0.650    & 0.646       & 0.646              & 0.650           & 0.954         & 0.874           \\

 \hline
\multicolumn{4}{l}{\footnotesize{$^*$ Data-level fusion with LongFormer}}  \\
\multicolumn{4}{l}{\footnotesize{$^\dag$ Stacking ensemble with KNN as meta learner}} \\
\end{tabulary}
\caption{Comparison of top performing model from unimodal and multimodal approaches}
\label{tab:mm_cod}
\end{table}

\end{landscape}

\begin{figure}
\begin{center}
\includegraphics[width=6in]{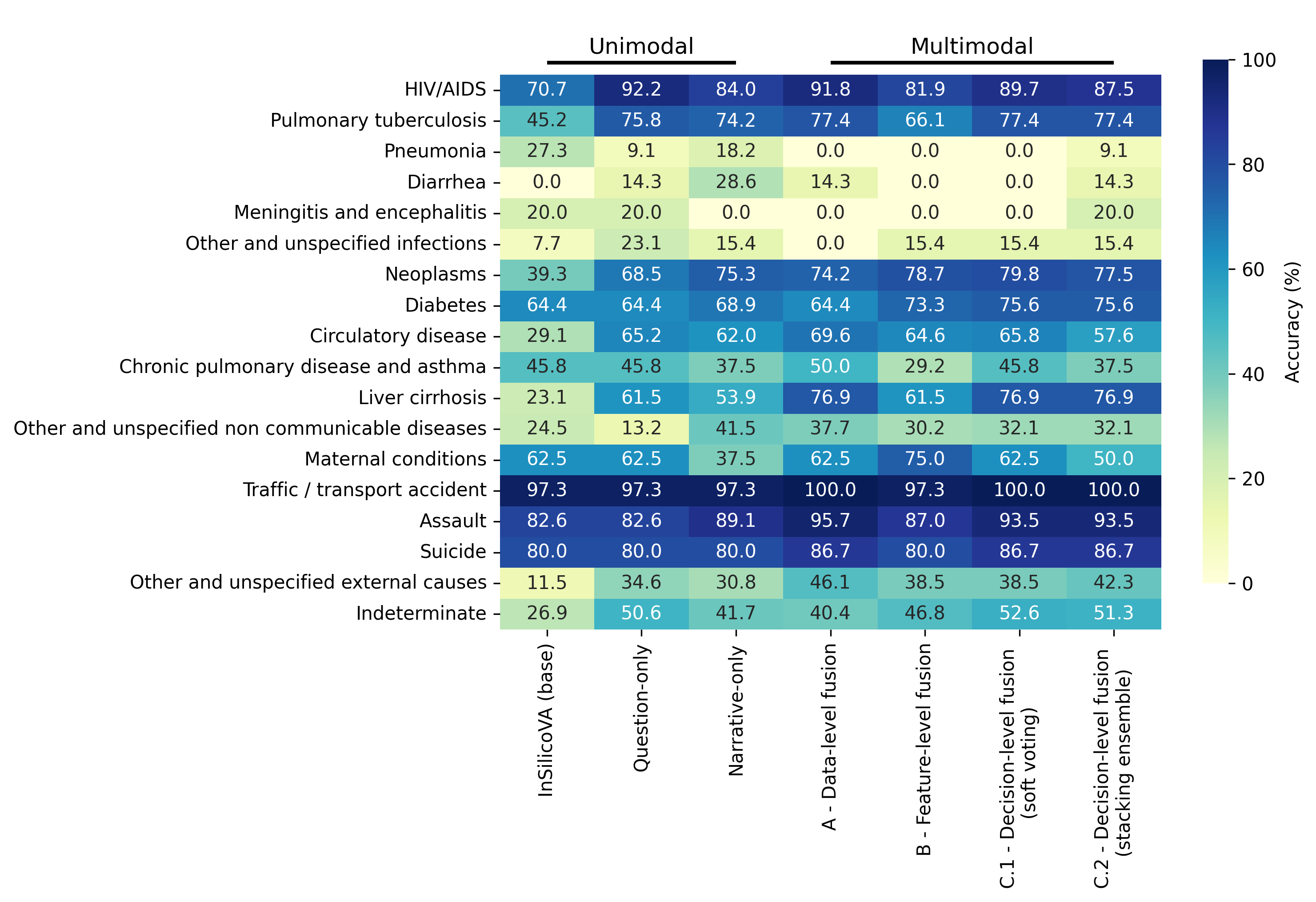}
\caption{Classification accuracy by cause: unimodal vs multimodal}
\label{fig:heatmap_multimodal}
\end{center}
\end{figure}

Figure~\ref{fig:heatmap_multimodal} illustrates the classification accuracy by cause for the model with the highest accuracy for each approach. Soft voting ensemble and data-level fusion with Longformer, though currently achieving the highest overall accuracy, both still tended to favor the majority classes. Compared to question-only approaches, incorporating narrative or narrative-based models further boosts performance for noncommunicable diseases, such as neoplasms, diabetes, and circulatory diseases.  

\begin{figure}
\begin{center}
\includegraphics[width=6in]{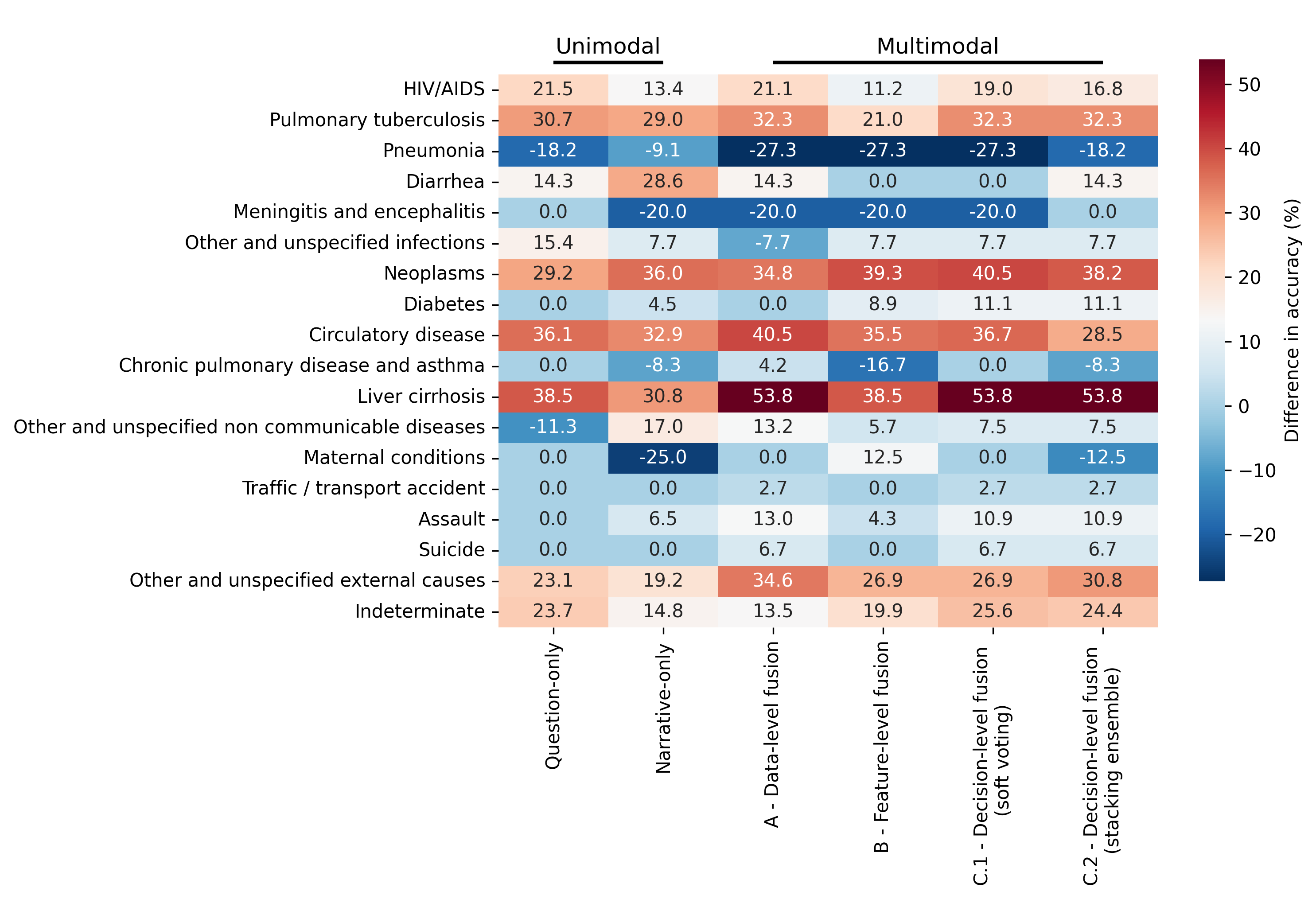}
\caption{Change in accuracy relative to InSilicoVA as baseline: multimodal models}
\caption*{{\footnotesize \textit{Note:} detailed model for each strategy: A. Data-level fusion with Longformer; B. Feature-level fusion with AutoMM, using RoBERTa-PM as text backbone; C.1 Decision-level fusion with soft voting ensemble using all models; C.2 Decision-level fusion with KNN as meta learner.}}
\label{fig:heatmap_multimodal_diff}
\end{center}
\end{figure}

\subsection{Sensitivity analysis}

\subsubsection{Impact of training data size}

\begin{figure}
\begin{center}
\includegraphics[width=6in]{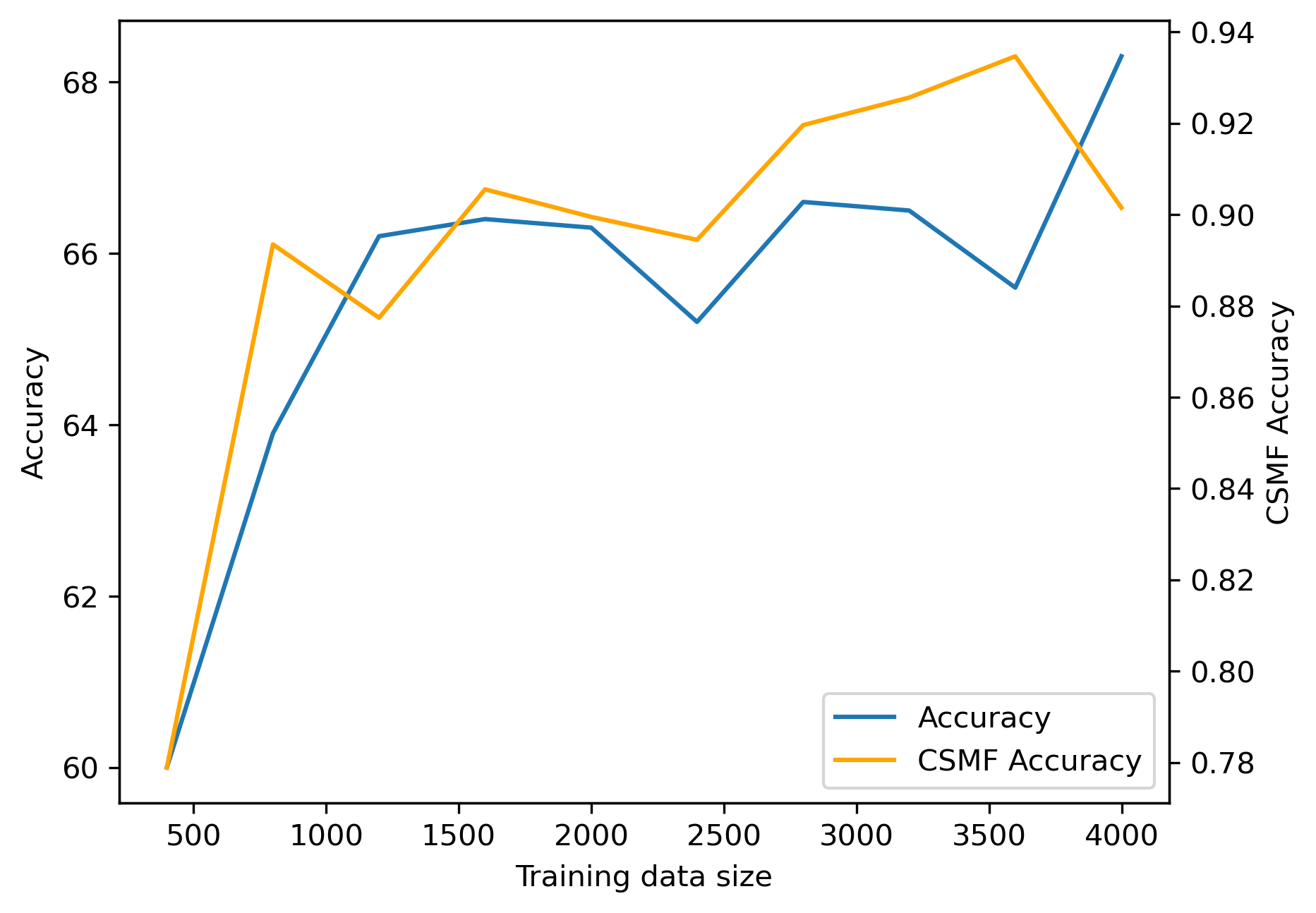}
\caption{Performance by training data size: data-level fusion model}
\label{fig:ensem_sa_samplesize_lf}
\end{center}
\end{figure}

Figure~\ref{fig:ensem_sa_samplesize_lf} demonstrates how accuracy and CSMF accuracy changed as the training data size for fine-tuning Longformer changed. Similar to findings with narrative-only models, the gain stalled with minimal increase after training data size reached about 1500 deaths. 

\subsubsection{Impact of model selection in decision-level fusion}

Table~\ref{tab:ensem_voting_sa} shows the performance of including different subsets of models in the soft voting ensemble. Ensembling multiple models within a single modality improved classification accuracy compared to using any single model. And incorporating models from both modalities further enhanced the individual-level results. There were slight variations in model performance depending on the base models included in the ensemble. We noticed that the inclusion of question-based models contributed more towards improving individual-level results, while the inclusion of narrative-based models contributed more towards improving population-level accuracy.

\begin{landscape}

\begin{table}[ht]
\centering
\captionsetup{format=plain,font=small,labelfont=bf, justification=justified, width=\textwidth}
\begin{tabulary}{\linewidth}{lCCCCCCCC}
 \hline
\multicolumn{1}{c}{\multirow{2}{*}{\textbf{Model selection}}} & \multicolumn{1}{c}{\multirow{2}{*}{\textbf{Accuracy}}} & \multicolumn{1}{c}{\multirow{2}{*}{\textbf{\thead{\makecell{Weighted \\ F1}}}}} & \multicolumn{1}{c}{\multirow{2}{*}{\textbf{\thead{\makecell{Weighted \\ Precision}}}}} & \multicolumn{1}{c}{\multirow{2}{*}{\textbf{\thead{\makecell{Weighted \\ Recall}}}}} & \multicolumn{1}{c}{\multirow{2}{*}{\textbf{ \thead{\makecell{CSMF \\ Accuracy}}}}} & \multicolumn{1}{c}{\multirow{2}{*}{\textbf{\thead{\makecell{CCCSMF \\ Accuracy}}}}} & \multicolumn{2}{c}{\textbf{Change to baseline$^*$}} \\
\cline{8-9}
\multicolumn{1}{c}{}                                          & \multicolumn{1}{c}{}                                   & \multicolumn{1}{c}{}                             & \multicolumn{1}{c}{}                                    & \multicolumn{1}{c}{}                                 & \multicolumn{1}{c}{}                                        & \multicolumn{1}{c}{}                                          & \textbf{Accuracy}    & \textbf{ \thead{\makecell{CSMF \\ Accuracy}}}   \\

 \hline
All models  (base)                                                  & \textbf{0.695}                                                  & \textbf{0.681}                                            & \textbf{0.680}                                                   & \textbf{0.695}                                                & 0.923                                                       & 0.790                                                         & 0.000                & 0.000                    \\
All ML                                                        & 0.669                                                  & 0.660                                            & 0.668                                                   & 0.669                                                & 0.909                                                       & 0.751                                                         & -0.026               & -0.014                   \\
All narrative-only                                            & 0.659                                                  & 0.650                                            & 0.656                                                   & 0.659                                                & \textbf{0.939}                                                       & \textbf{0.833}                                                         & -0.036               & 0.016                    \\
All question-only$^\dag$                                             & 0.673                                                  & 0.663                                            & 0.672                                                   & 0.673                                                & 0.907                                                       & 0.746                                                         & -0.022               & -0.016                   \\

 \hline
 \multicolumn{9}{l}{\textbf{Leaving one model out}} \\
 \hline
All but BioMedBERT      & \textbf{0.708} & \textbf{0.695} & \textbf{0.698} & \textbf{0.708} & 0.923 & 0.790 & 0.013  & 0.000  \\
All but InSilicoVA      & 0.707 & 0.694 & 0.697 & 0.707 & 0.923 & 0.790 & 0.012  & 0.000  \\
All but BlueBERT        & 0.702 & 0.687 & 0.689 & 0.702 & 0.913 & 0.762 & 0.007  & -0.010 \\
All but ClinicalBERT    & 0.701 & 0.687 & 0.691 & 0.701 & 0.917 & 0.773 & 0.006  & -0.006 \\
All but BERT            & 0.699 & 0.684 & 0.690 & 0.699 & 0.916 & 0.771 & 0.004  & -0.007 \\
All but BioClinicalBERT & 0.697 & 0.683 & 0.691 & 0.697 & 0.916 & 0.771 & 0.002  & -0.007 \\
All but RoBERTa-PM      & 0.696 & 0.681 & 0.685 & 0.696 & 0.918 & 0.776 & 0.001  & -0.005 \\
All models              & 0.695 & 0.681 & 0.680 & 0.695 & 0.923 & 0.790 & 0.000  & 0.000  \\
All but CatBoost        & 0.690 & 0.676 & 0.678 & 0.690 & 0.923 & 0.790 & -0.005 & 0.000  \\
All but GBDT            & 0.690 & 0.676 & 0.678 & 0.690 & \textbf{0.924} & \textbf{0.792} & -0.005 & 0.001  \\
All but XGBoost         & 0.689 & 0.675 & 0.673 & 0.689 & \textbf{0.924} & \textbf{0.792} & -0.006 & 0.001  \\
All but LightGBM        & 0.688 & 0.674 & 0.676 & 0.688 & \textbf{0.924} & \textbf{0.792} & -0.007 & 0.001  \\
All but MLP             & 0.687 & 0.672 & 0.674 & 0.687 & \textbf{0.924} & \textbf{0.792} & -0.008 & 0.001 \\
 \hline
\multicolumn{8}{l}{\footnotesize{$^*$ Change as compared to baseline ensembling all models (ML+PLM+InSilicoVA)}} \\
\multicolumn{8}{l}{\footnotesize{$^\dag$ All question-only includes question-based ML and InSilicoVA}} \\
\end{tabulary}
\caption{Performance by COD grouping levels}
\label{tab:ensem_voting_sa}
\end{table}

\end{landscape}

\section{Discussion and conclusion}

In this study, we proposed and evaluated several multimodal fusion strategies that combined the narratives and the structured question responses for automated COD classification using VA. Both decision-level fusion and data-level fusion approaches outperformed the best-performing question-only models, for both individual-level and population-level metrics. These findings implied that each modality in a multimodal data source had unique strengths and contributions. Integrating narratives into the COD classification pipeline can further enhance the accuracy of model-based estimates for both individual and population levels.

The improvement in the classification of non-communicable diseases with the addition of narratives was expected. Although comprehensively reviewing major signs and symptoms from all systems, structured questions could only collect limited information with a pre-defined level of details - mostly the duration and perceived severity, while lacking key signature patterns valuable for differential diagnosis of diseases. Tabular responses only reported the co-occurrence of proximal symptoms, without direct insight into the timing and sequence of events, which can be critical for understanding the chain of events leading to the death. In contrast, unstructured narratives can report medical histories with richer details about the onset and progression of illness, past treatments, health services utilization, diagnostic test results, etc.. By integrating complementary information from the narratives to the structured questions, multimodal approaches can provide a more holistic understanding of the conditions preceding the death, improving the COD ascertainment.

Each multimodal fusion strategy has strengths and weaknesses. Data-level fusion is most straightforward by design, naturally accommodates the text-tabular structure of survey data, and has achieved individual-level performances among the top. However, the application of this approach can hardly be extended to other modalities such as pathological or clinical images, or audio of interviews. It might also be less effective with other tabular data like lab results from Minimally Invasive Tissue Sampling (MITS). MITS data usually contains a large number of variables with limited semantic diversity, which might not benefit as much from deep contextual embeddings of language models. Furthermore, even Longformer-style architectures are designed to handle long sequences efficiently, the constraints on input length for these language models still limit their application to long text data. Lastly, the challenges with under-represented minority classes and the tendency of models to favor major classes still exist. Good balancing techniques for text data are needed to address class imbalances when using data-level fusion. 

By contrast, feature-level fusion and decision-level fusion offer higher flexibility, and can be readily extended to incorporate various modalities beyond text and tabular data. These strategies could be more widely applicable in the medical and health domain, where data often come from diverse sources in various formats such as narratives, structured surveys, laboratory results, and diagnostic images. Feature-level fusion is the most lightweight approach amongst all. It requires significantly lower computational resources to train, while still achieving roughly comparable performance. Off-the-shelf AutoML tools like AutoGluon/AutoMM make this approach relatively accessible and easy to adopt, especially for nontechnical researchers. Although based on our experience in this study, one might still need to customize the model configurations and tune the hyperparameters properly, in order to achieve the optimal performance. Moreover, decision-level fusion is the most flexible strategy, allowing for the integration of various types of base models, including traditional statistical models, machine learning models, and language models. At the same time, it is also the most computationally expensive approach, particularly when using the stacking ensemble design. And it requires careful choice and calibration of each base model for optimal and robust outcomes.

The models used in the current experiment of ensemble learning are relatively homogenous in architecture. PLMs mostly have the BERT-based structures. Question-based models were selected from top-performers of popular ML models, and top models - LightGBM, CatBoost, XGBoost - were all tree-based models built on gradient boosting frameworks showing roughly similar performance. Ensembling more heterogeneous models would benefit from more diverse perspectives, reduce the risk of bias towards correlated error, and improve the overall generality of the model. More work could be done to broaden the candidate pool of base models, to assess whether ensembling more heterogeneous models could improve the performance and the robustness of the models. Another limitation of the study is the lack of uncertainty estimates. Current methods only report point estimates for individual predictions and CSMF, which might suggest a level of precision that is not warranted given sampling and model uncertainty. Properly quantifying uncertainty is important for cross-setting comparisons and interpreting trends over time, as it helps researchers and policy makers to more precisely distinguish statistically significant differences from random noise or fluctuations, therefore reducing the risk of misinformed decisions about policy priorities and resource allocation. 

Future research should investigate the appropriate ways to address class imbalance in different multimodal approaches. For example, techniques like SMOTE (Synthetic Minority Over-sampling Technique) can be incorporated into the feature-level fusion framework, while decision-level fusion would require modality-specific balancing techniques. The impact of data balancing on model building and interpretation remains to be fully understood. In addition, data might not always be available for all modalities across settings or cases. Future research can develop more flexible strategies to better handle missing modalities. 

Lastly, we should also note that "the model is only as good as the data it is fed". The quantity and quality of information required for diagnosis vary greatly across causes. The consistently low and unstable performance in identifying minority causes for all models - particularly non-HIV/TB infections - might also imply insufficiency of information for determining these challenging causes in the VA data, rather than the less optimal modeling strategies. This leads up to the question we would like to address in the next chapter.

    % multimodal
\chapter{Sufficiency of information in verbal autopsy}
\label{sufficiency.ch}

\section{Introduction}

This chapter descriptively explores the sufficiency of information in empirical VA. 

The performance of models for automated COD classification depends not only on the modeling strategy but also largely on the sufficiency of information in the data. In cases where information is insufficient, it is equally challenging for even physicians to determine the underlying COD. Therefore, understanding the sufficiency of information in VA as a tool for COD data collection is a critical first step towards the ultimate goal of producing accurate estimates of mortality burden by cause, in mortality surveillance and vital registration systems in resource-limited settings.

Sufficiency of information in medical or VA settings doesn't necessarily mean the length of narratives or the number of symptoms reported. But rather, it means the information with adequate details and patterns, in other words, the "most useful" information for differential diagnosis, which is the core element in clinical decision making. It can either be the presence of a definitive feature clearly pointing to or excluding a cause, or the concurrence of symptoms and/or progression patterns distinguishing between probable causes. 

For example, in the cases presented in the text box below, Case 1 clearly explained the proximal events with a timeline (e.g., an acute condition with high severity), included signature symptom patterns (e.g., concurrence of severe headache and stiff neck), supported by clinical diagnosis (e.g., "meningitis"), and structured questions. Therefore, despite its brevity and the limited number of symptoms reported, the case was rated by the physicians as having excellent sufficiency to assign the underlying COD. In contrast, Case 2 lacked a logical roadmap connecting the reported medical conditions, nor were the conditions supported by specific symptoms directly leading to death, limiting its diagnostic value. While in Case 3, although particular symptoms and timeline were reported, many symptoms were vague and nonspecific, which could be shared by multiple infectious diseases. This highlights the fundamental challenge of relying solely on VA as the ultimate solution for COD ascertainment, in the absence of necessary etiological or pathological findings for accurate diagnosis. 

\begin{tcolorbox}[title=A few examples of VA narrative, colback=gray!5, colframe=black, fonttitle=\bfseries]
\label{tab:narrative_example}
\begin{Verbatim}[breaklines=true]
"""
Case 1 - from excellent sufficiency case:
Two weeks before death the deceased had severe head ache and stiff neck. He was taken to HOSPITAL where he was admitted for two days and then referred to HOSPITAL. He was admitted for a week and got diagnosed with meningitis. He died on the [DATE MASKED] at hospital. 
"""
Case 2 - from poor sufficiency case:
The decendent has been sick for 7 years at first it was TB he got treatment for it and was healed then later on he had stroke the whole left side he went to the clinic and hospital and he healed after some time, [YEAR MASKED] he was attacked by stroke again his whole right side this time he went to hospital and was tolds that he also have high blood pressure and is HIV positive but then he died in the hospital before began the treatment on the [DATE MASKED].
"""
Case 3 - from poor sufficiency case:
DECEDENT was sick with TB; High blood pressure and asthma for a few years. During [DATE MASKED] he was in and out of hospital and also in  [DATE MASKED] he spent 2 weeks in hospital. In  [DATE MASKED] he do spent a week in hospital then in  [DATE MASKED] he get seriously ill once more on  [DATE MASKED] he started having a swollen stomach and his feet also got swollen and his feet were black. He also had breathlessness, severe cough, chest pain and he was vomiting a lot. He was hospitalized that same day  [DATE MASKED] and spent a week there until he died in [DATE MASKED].
"""
\end{Verbatim}
\end{tcolorbox}

Quantitatively measuring the sufficiency of information in a survey can be extremely challenging to conceptualize and operationalize. Our discussions with physicians revealed that human assessment of sufficiency was highly subjective and largely experience-based, rather than guided by any clearly defined or standardized rubric. However, with a multimodal framework powered by NLP and ML techniques, we were hoping to take an initial look "under the hood", to explore which survey elements might have been contributing to the physician-perceived information sufficiency. We believe such insights could be helpful to inspire future research, support the development of more formal evaluation frameworks, inform the improvement of VA data collection protocols and instruments, and identify the gaps in the overall COD classification pipeline. 

In this study, we first illustrate the level of information sufficiency in empirical data, and explore the association between perceived information sufficiency and the COD classification accuracy for both physicians and models. Then we use a feature-fusion approach to predict the levels of information sufficiency from VA data, and describe important features in prediction as identified by the model. 

\section{Methods}

\subsection{Data}

This study again used data from SANCOD. In addition to standardized VA questions and narratives, the trained physicians who reviewed and coded COD from VA were also asked to provide their subjective evaluation of the sufficiency of VA data - referred to as the sufficiency score. It reflected physicians' perception of "how sufficient the VA information was to certify the cause of death", ranging from 1 (very poor) to 5 (excellent) sufficiency.\cite{awotiwon2022anaconda} We grouped the scores into three levels for analytical purposes: scores 1-2 as low, score 3 as medium, and score 4-5 as high. A total of 4719 adult VAs with valid sufficiency score were used for analysis. 

\subsection{Sufficiency and classification performance}

To understand the association between human performance and information sufficiency, illustrated how model performance varied with information sufficiency, using results from previous chapters with PCVA as the reference. 

To evaluate the performance of physician coding using VA alone, we used a subset of 1924 adult VAs where medical/forensic record certified causes (PCMR) were available. For these cases, the medical/forensic records were also presented at the time of VA and later reviewed by trained physicians to additionally assign PCMR informed by both sources. We, treating PCMR as the true "gold standard", evaluated the accuracy of physician coding using VA alone (PCVA with level 2 COD), as a proxy to understand the validity of physician-perceived sufficiency. 

%\subsection{COD classification with high-sufficiency VA only}

\subsection{Prediction of sufficiency}
We used a feature-fusion approach to predict the levels of sufficiency from VA data. On one hand, the feature-level fusion could better allow for interaction between modalities. On the other hand, the narrative-based features extracted using the traditional NLP had better explanability and were relatively more human-readable, comparing to deep contextual embeddings from transformer-based models. Details of the feature-level fusion strategy can be found in the previous chapters (Section \ref{info:}).

We followed the conventional procedure for preprocessing the narrative data in traditional NLP, including lower casing, punctuation and stop words removal, and lemmatization etc.. (Appendix~\ref{app1:narrative_preprocess}) Then we extracted text features using term frequency–inverse document frequency (TF-iDF) with n-grams, applied singular value decomposition (SVD) for dimensionality reduction. The top 450 components were retained and concatenated with raw tabular features from the questions. The fused features were used as input to AutoGluon tabular model,\cite{erickson2020autogluon, shi2021benchmarking} to predict the sufficiency level. AutoGluon is from the same family of models as AutoMM used in the previous chapter, which automatically optimizes model configurations, trains and selects multiple ML models and ensemble models using structured data. 

Similar to previous analyses, we applied an 80-20 stratified split for training and testing, respectively. Prediction accuracy was used as the metric for hyperparameter optimization, model selection, and evaluation, since the sufficiency levels were relatively well balanced in our data. We trained models under three scenarios - narrative-only, question-only, and multimodal which used features from both modalities. To further understand the contribution of each modality to information sufficiency, we calculated \textit{the average marginal contribution} of each modality as follows:
\begin{align*}
\text{Total Gain} &= \text{acc}_{n \cup q} - 0.5 \times \left( \text{acc}_{n} + \text{acc}_{q} \right) \\
\text{Contribution}_{n} &= \frac{0.5 \times \left( \text{acc}_{n \cup q} - \text{acc}_{q} \right)}{\text{Total Gain}} \times 100\% \\
\text{Contribution}_{q} &= \frac{0.5 \times \left( \text{acc}_{n \cup q} - \text{acc}_{n} \right)}{\text{Total Gain}} \times 100\%
\end{align*}
Where:
\begin{itemize}
  \item $\text{acc}_{n}$ = accuracy using narrative-based features only
  \item $\text{acc}_{q}$ = accuracy using question-based features only
  \item $\text{acc}_{n \cup q}$ = accuracy using both modalities
\end{itemize}

We also explored the feature importance as measured by SHapley Additive exPlanations (SHAP) value. SHAP measures how individual features influence final model prediction by calculating the marginal contribution of each feature across different model subsets in a unified framework. A higher SHAP value indicates that the feature has a stronger influence on the final model prediction.

The analysis was performed using Python 3.12 \cite{python312}, with the Scikit-learn v1.4.2 \cite{pedregosa2011scikit} and the AutoGluon v1.2 \cite{erickson2020autogluon} packages.

\section{Results}

\subsection{Descriptive analysis of information sufficiency level}

\begin{figure}
\begin{center}
\includegraphics[width=5in]{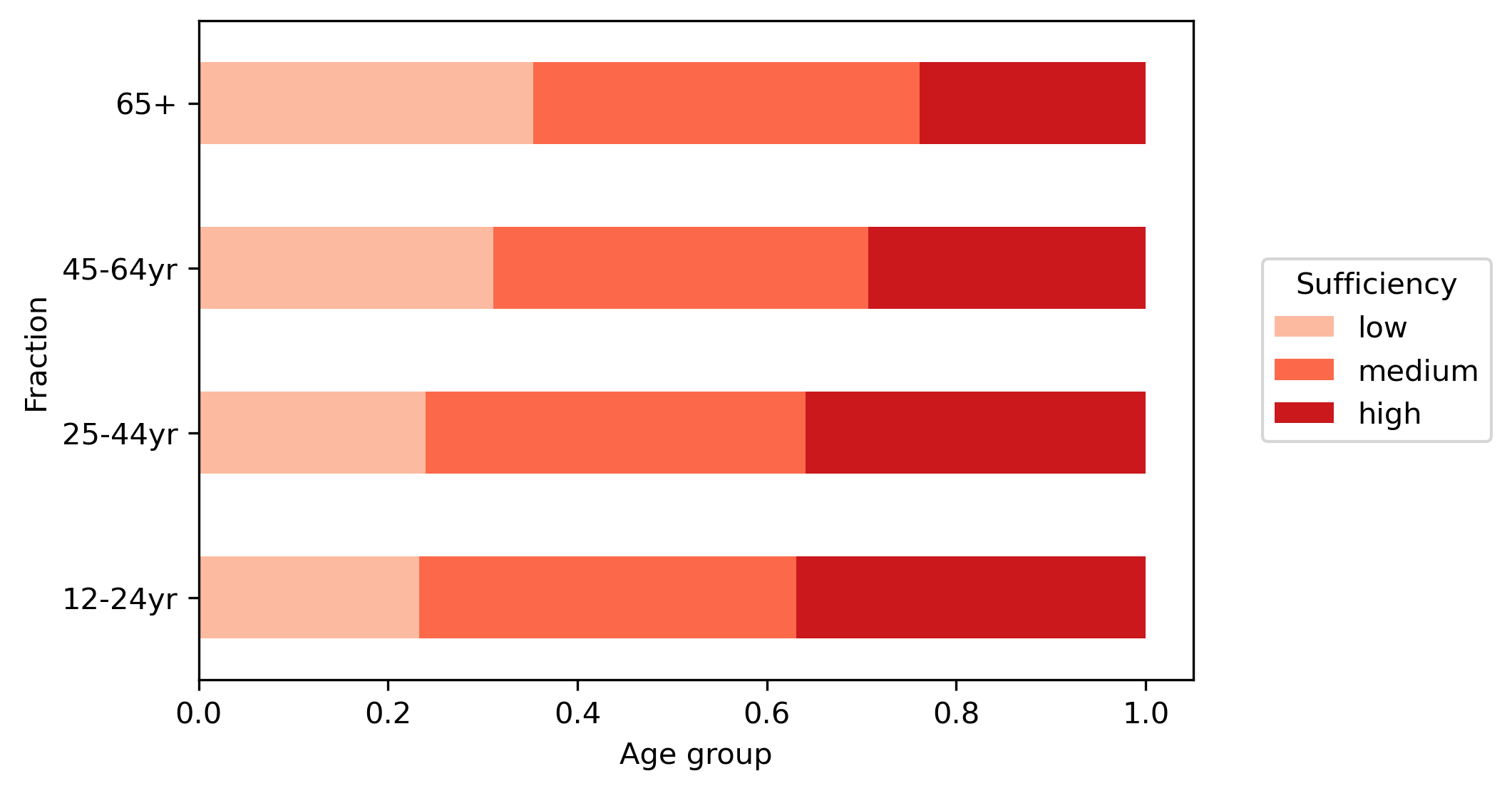}
%\caption{Distribution of sufficiency score by age group}
\caption*{(A) By age group}

\vspace{2em}

\includegraphics[width=6in]{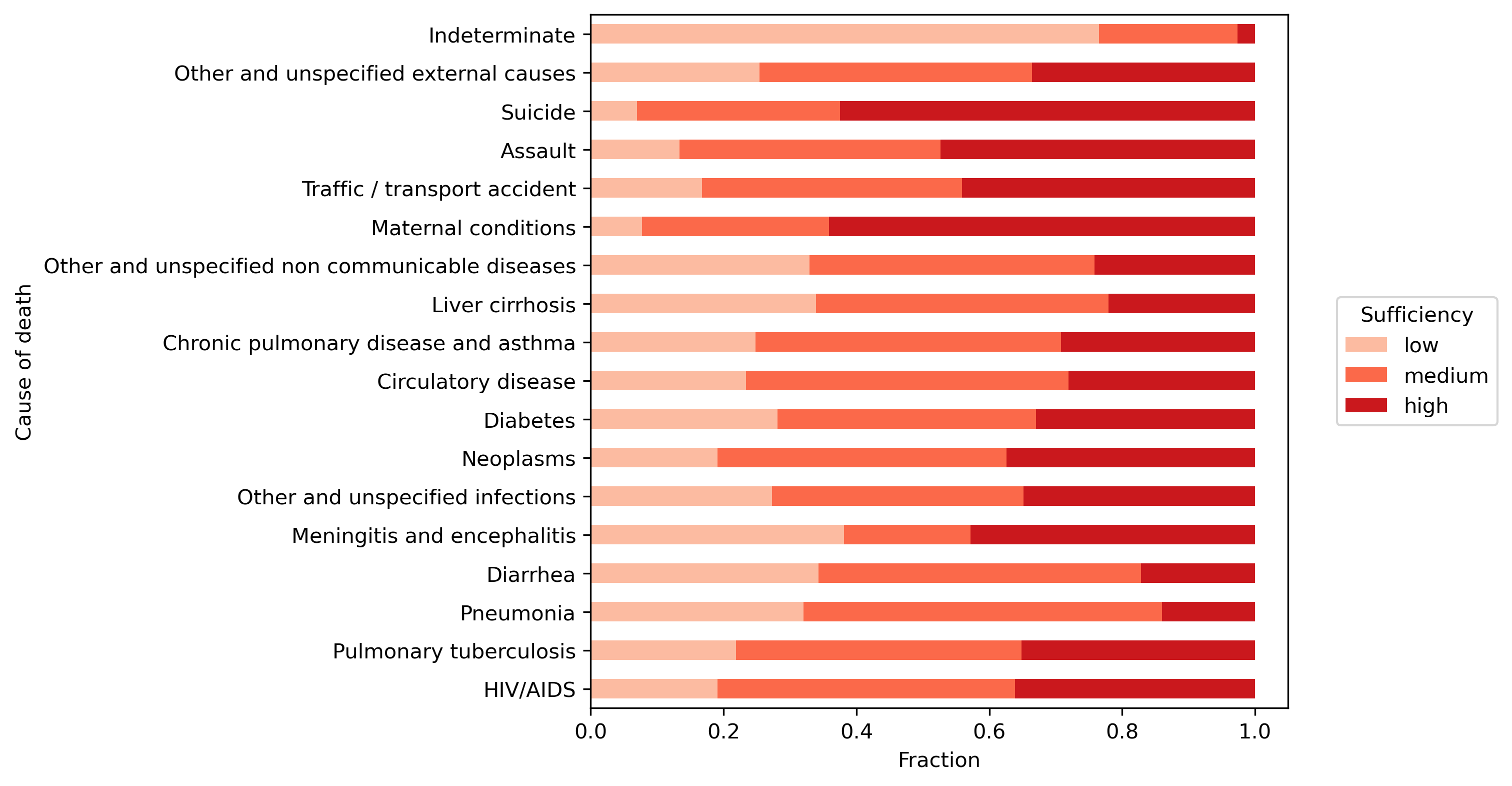}
%\caption{Distribution of sufficiency score by Level-2 COD}
\caption*{(B) By Level-2 COD}
\caption{Distribution of sufficiency score}
\label{fig:suff_dist}
\end{center}
\end{figure}

Overall, the sufficiency level appeared to be nearly evenly distributed, with low and high sufficiency each accounting for roughly 30\% of VAs. However, we noticed salient variation across age and causes. Physicians found it more difficult to get sufficient information for assigning COD for deaths from older adults. External causes and maternal conditions had the highest sufficiency, whereas pneumonia and diarrhea were dominated by low to medium sufficiency. These observations were consistent with the findings reported in previous chapters. Causes with lower sufficiency, such as pneumonia and diarrhea, were also the causes on which most models performed poorly. (Figure~\ref{fig:suff_dist})  

\par\noindent

\begin{figure}[ht]
\begin{center}
\includegraphics[width=6in]{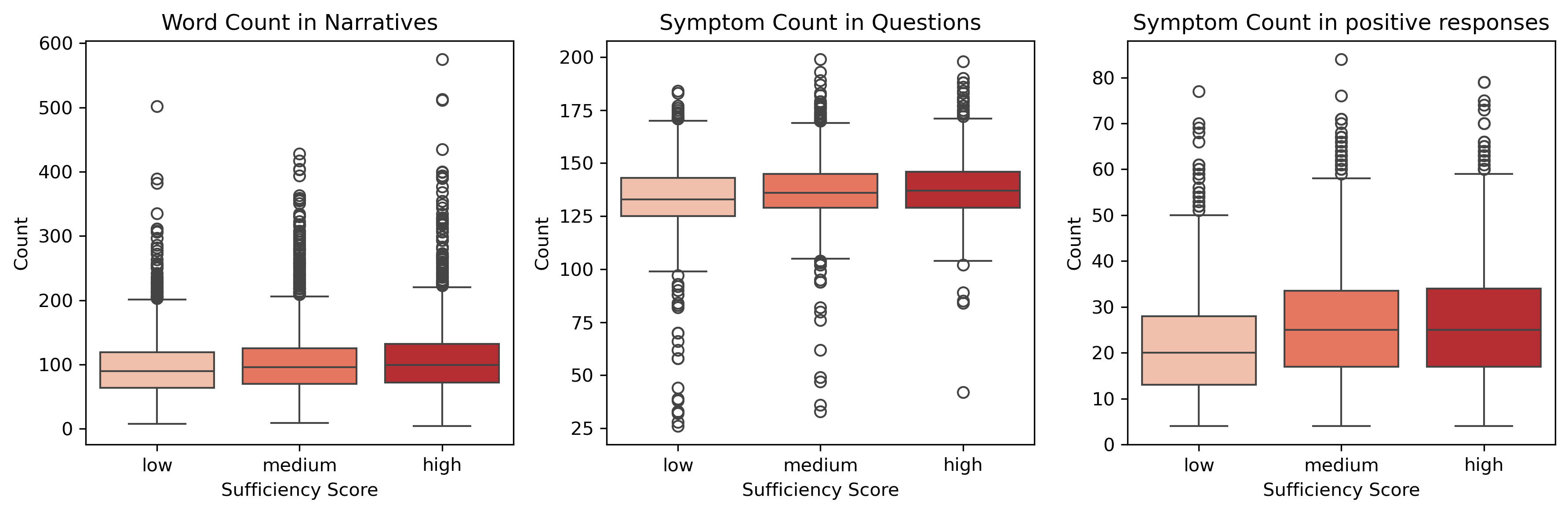}
\caption{Narrative lengths and number of symptoms reported, by sufficiency score}
\label{fig:suff_dist_wordcount}
\end{center}
\end{figure}

We also examine the length of the narrative, the number of symptoms reported, and the number of positive symptoms reported, at different levels of sufficiency. (Figure \ref{fig:suff_dist_wordcount}) The distribution of word counts and symptom counts was largely overlapping between sufficiency levels. This matched our qualitative observations that sufficiency is not necessarily / solely determined by the "quantity" of information provided in VA, but more likely influenced by relevance and specificity of the content to diagnostic decision making.

\subsection{Classification accuracy and information sufficiency}

\subsubsection{Model performance}

\begin{figure}[ht]
\begin{center}
\includegraphics[width=5.5in]{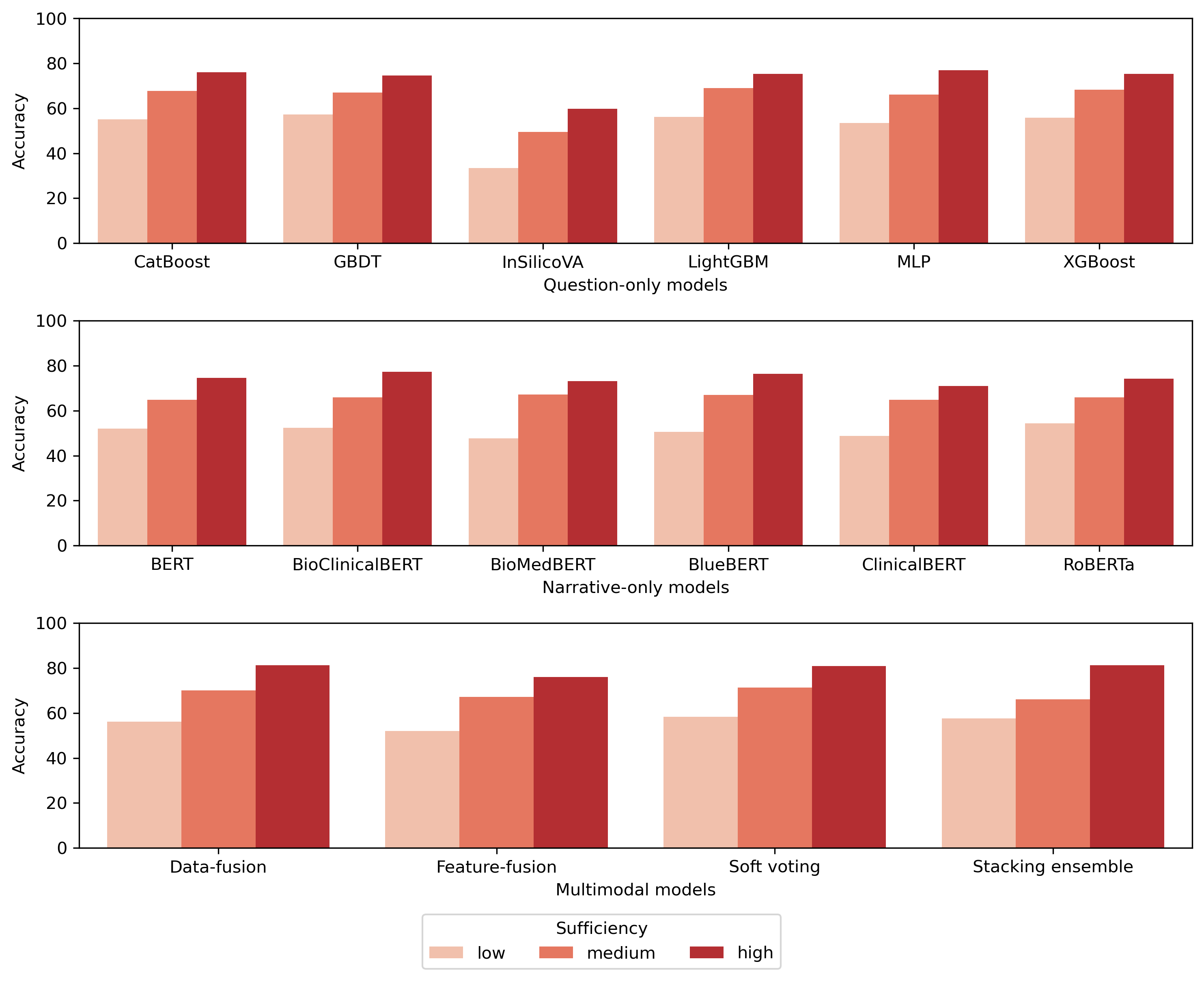}
\caption{Model classification accuracy by sufficiency level}
\label{fig:suff_eval_models}

\vspace{3em}

\includegraphics[width=3.5in]{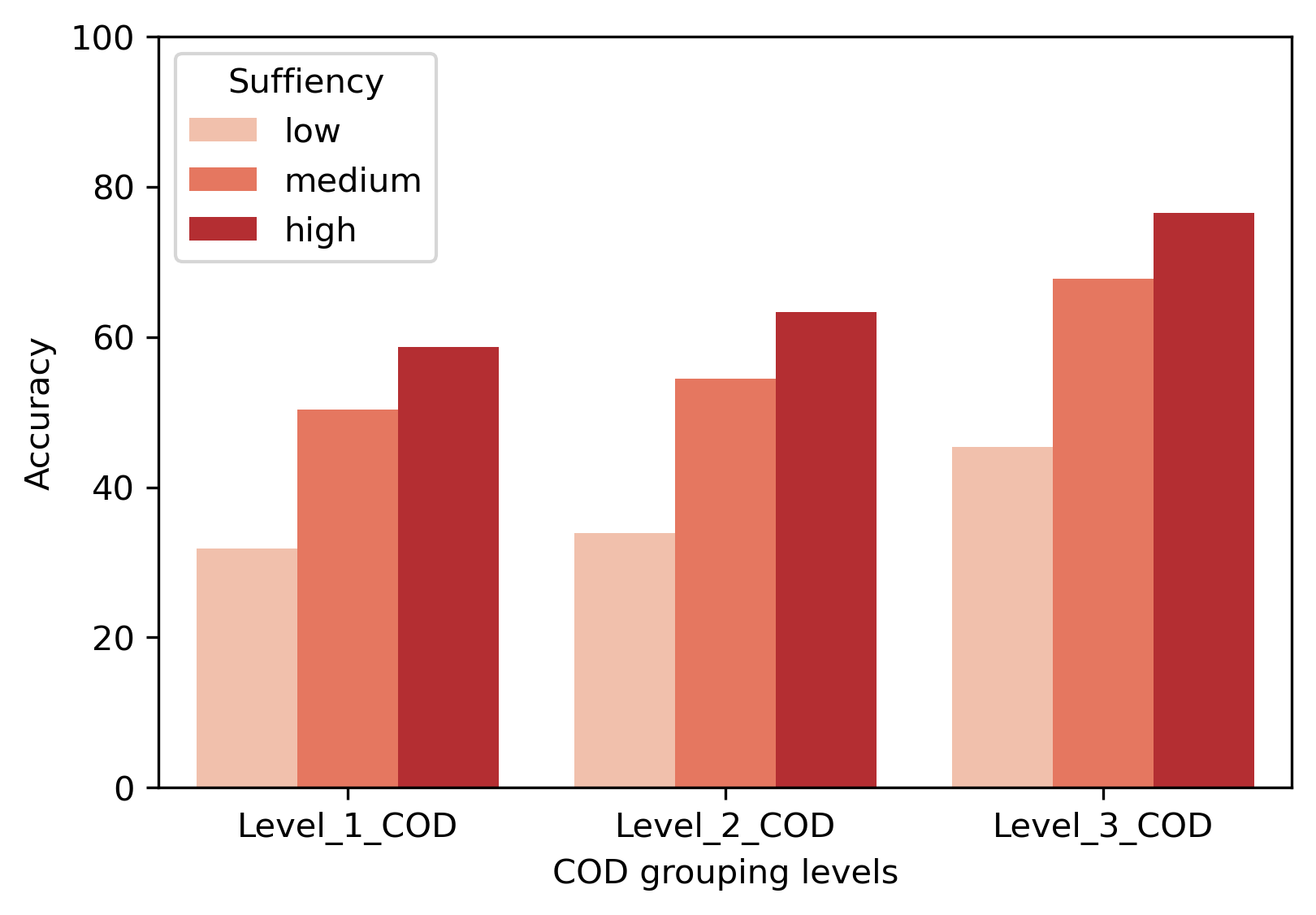}
\caption{Human classification accuracy for sufficiency level, and COD grouping level}
\label{fig:suff_eval_pcva}

\end{center}
\end{figure}

Clear gradients in classification accuracy were observed for all models across the sufficiency level. (Figure \ref{fig:suff_eval_models}) VAs with higher sufficiency as perceived by physicians were also more likely to be correctly classified by all modeling strategies.  

\par\noindent

\subsubsection{Physician performance}

Using a medical/forensic certified cause as the reference standard, we found that physicians' ability to assign accurate COD was strongly associated with their perceived sufficiency of information. (Figure~\ref{fig:suff_eval_pcva}) Notably, the gradient in classification accuracy persisted across all levels of COD grouping, from the most granular causes to the broad COD categories. 

For 83\% of deaths coded as "indeterminate" using information only from VA, physicians were able to assign a determinate COD after reviewing the medical/forensic record. Most of these causes were non-communicable diseases. (More details in confusion matrices comparing PCVA against UCMR by sufficiency level for level 2 COD in Appendix section~\ref{app2:suff_cm_ucmr}) This suggests that details critical for COD assignment might not have been reported in the current VA interviews. Such gaps could be attributed to the lack of information, or reporting bias of the respondents. Alternatively, they might also indicate the need for better probing during the interview, or the addition for other complementary data sources (such as MITS, autopsy, or medical record linkage), to enhance infomration sufficiency for COD determination. 

\subsection{Predicting sufficiency of information}

With multimodal using features from both modalities, we were able to predict sufficiency level at an accuracy of 44.5\%.(Table~\ref{tab:suff_pred}) Both modalities contributed to the prediction of sufficiency, and the majority of the contribution came from questions with an average marginal contribution of 85.5\% for question-based features.  

\begin{table}
\centering
\captionsetup{format=plain,font=small,labelfont=bf, justification=justified, width=\textwidth}
\begin{tabulary}{\linewidth}{cccc}
 \hline
Modality       & Best-performing model   & Accuracy  & Contribution \\
 \hline
Question-only  & LightGBM      & 0.435   & 85.5\%     \\
Narrative-only & LightGBM      & 0.386   & 14.5\%      \\
Multimodal     & Random Forest & 0.445  & -    \\
 \hline
\end{tabulary}
\caption{Accuracy for predicting level of information sufficiency}
\label{tab:suff_pred}
\end{table}

\begin{figure}
\includegraphics[width=6in]{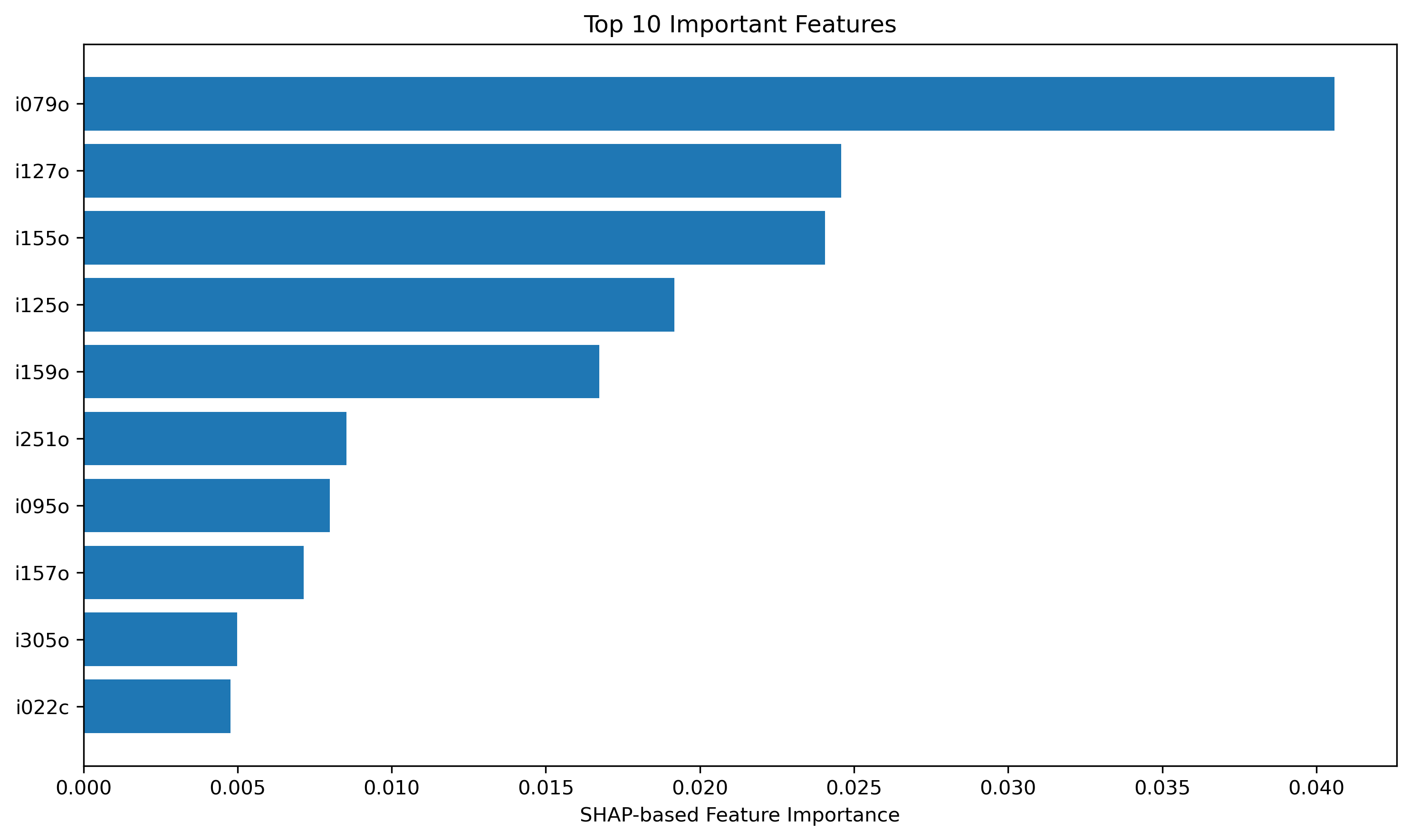}
\caption{Top 10 important features in question-based sufficiency prediction}
\label{fig:suff_pred_fi_tab}
\end{figure}

Figure~\ref{fig:suff_pred_fi_tab} shows the features with the highest SHAP feature importance values in the question-only model, with feature descriptions presented in Table~\ref{tab:suff_pred_fi_tab_desc}. Questions pointing to specific events (e.g., injuries, pregnancy) and signature symptoms (e.g., coughing with sputum, coughing blood), or confirming past diagnosis (e.g., for HIV/AIDS, or tuberculosis), were among the features contributing the most to the sufficiency of information. 

\begin{figure}
\includegraphics[width=6in]{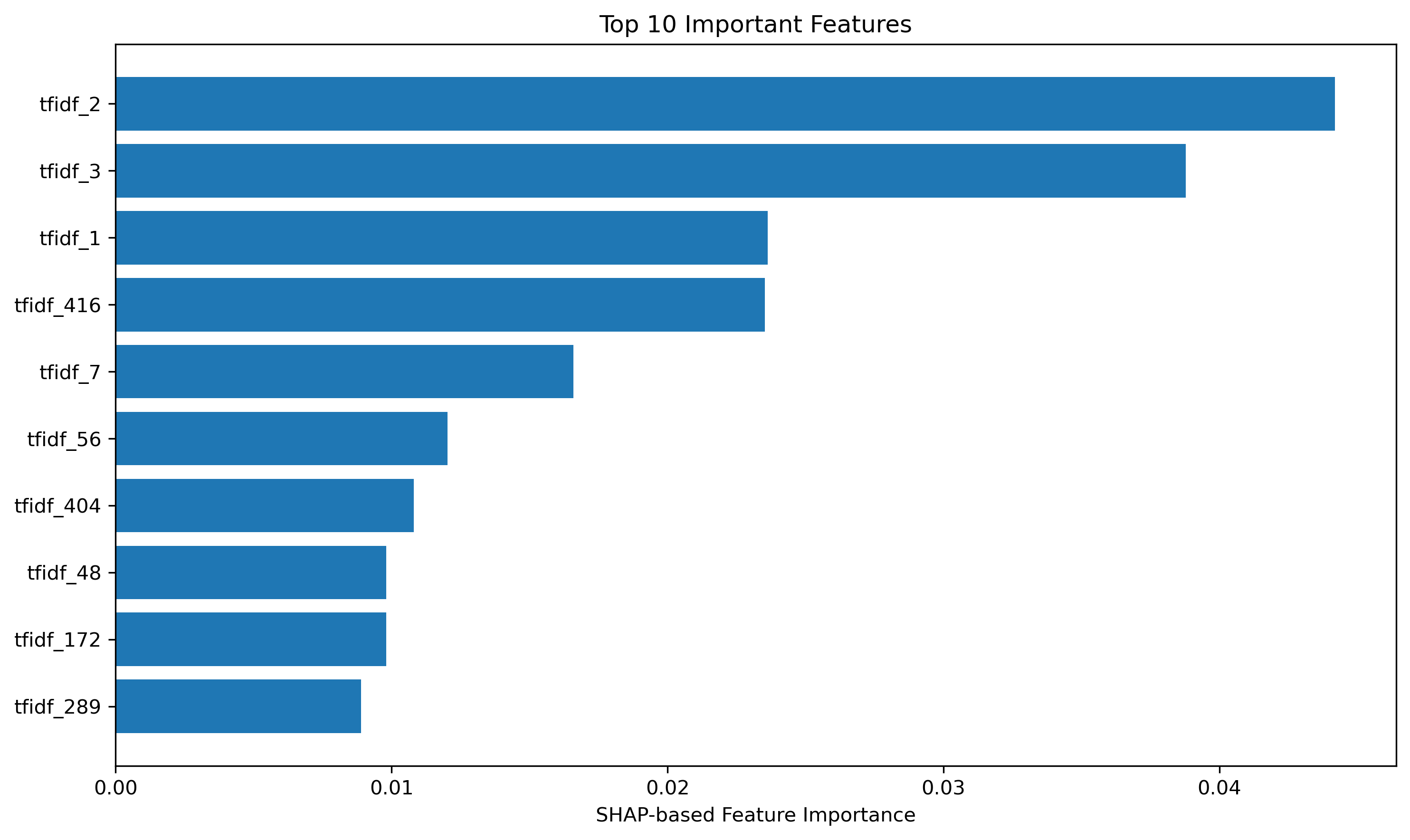}
\caption{Top 10 important features in narrative-based sufficiency prediction}
\label{fig:suff_pred_fi_text}
\end{figure}

Figure~\ref{fig:suff_pred_fi_text} shows the features with top SHAP feature importance values in the narrative-only model, with top contributing N-grams to the SVD component presented in Table~\ref{tab:suff_pred_fi_text_desc}. (More information on narrative features, see figure~\ref{fig:suff_pred_fi_text_ngram}) [High blood pressure, diabetes], and [HIV, TB] were the top two features showing the greatest influence on sufficiency prediction. In multimodal prediction, no single feature stood out in terms of having a significantly higher impact on sufficiency level. (Figure~\ref{fig:suff_pred_fi_mm})

\begin{table}
\centering
\captionsetup{format=plain,font=small,labelfont=bf, justification=justified, width=\textwidth}
\begin{tabulary}{\linewidth}{LL}
 \hline
Feature & Description                                                                    \\
 \hline
i079o   & Was (s)he injured in a road traffic accident?                                   \\
i127o   & Was there any diagnosis by a health professional of HIV/AIDS?                  \\
i155o   & Was the cough productive, with sputum?                                         \\
i125o   & Was there any diagnosis by a health professional of tuberculosis?              \\
i159o   & During the illness that led to death, did (s)he have any difficulty breathing? \\
i251o   & Did (s)he have both feet swollen?                                              \\
i095o   & Was (s)he injured by a force of nature?                                        \\
i157o   & Did (s)he cough up blood?                                                      \\
i305o   & Was she pregnant (incl. in labour) at the time of death?                       \\
i022c   & Was s(he) aged 65 years or more at death?                                     \\
 \hline
\end{tabulary}
\caption{Descriptions of top features in question-based sufficiency prediction}
\label{tab:suff_pred_fi_tab_desc}
\end{table}

\begin{table}
\centering
\captionsetup{format=plain,font=small,labelfont=bf, justification=justified, width=\textwidth}
\begin{tabulary}{\linewidth}{LL}
 \hline
Feature & Top contributing N-grams (summarized)                                                                     \\
 \hline
tfidf\_2   & blood pressure, high, diabetes                    \\
tfidf\_3   & death, hiv, tb, deceased diagnosed               \\
tfidf\_1   & found, went, called, deceased, dead, car, police \\
tfidf\_416 & face, arm, hospitalized, week died               \\
tfidf\_7   & death, cause death, pain                         \\
tfidf\_56  & chest pain, legs, mental confusion               \\
tfidf\_404 & tired, improvement, issues, belly                \\
tfidf\_48  & leg, lost, back                                  \\
tfidf\_172 & eat, refused, collapsed, final, headache         \\
tfidf\_289 & liver, signficant, developed, mouth, swollen legs  \\
 \hline
\end{tabulary}
\caption{Descriptions of top features in narrative-based sufficiency prediction}
\label{tab:suff_pred_fi_text_desc}
\end{table}

\section{Conclusion}

Information in VA is often insufficient. Using empirical VA data from SANCOD, we reported that about one-third of adult VAs were lacking sufficient information to assign an underlying COD, as rated by physicians. The level of sufficiency varies by both the age of the deceased and the COD. And the sufficiency of information is strongly related to the accuracy of COD classification for both the physicians and the models. 

These findings aligned well with the patterns observed in the field as well as prior model performance evaluations. Causes with higher information insufficiency were also often where the models consistently performed poorly. This reinforces our argument that VA should not be used as a standalone instrument. Instead, it should be integrated into an adaptive, live diagnostic system. Cases due to certain causes or lacking sufficiency should be promptly flagged and referred for further investigation with downstream services such as MITS, pathological exams, or full autopsy, to improve the precision of COD classification and CSMF estimation. 

The information in VA appeared to be less sufficient for deaths occurring at older ages. This may be due to comorbidities and complications among older adults. In such cases, detailed information on specific symptoms leading to the death, the timeline, and the progression of events becomes especially important to understand the chain of events, for distinguishing between underlying and contributing COD. This information needs to be either added to structured questions, or better probed or recorded in the narratives. 

The proposed models achieved relatively low accuracy in predicting sufficiency levels. This could be attributed to several factors. First, the input features might not be adequately informative, particularly the ones dervied from the narratives. The top SVD components of narrative-based features explained only a modest proportion of the variance and appeared noisy, which indicated that the information in the text might not have been well condensed and represented with traditional NLP in this context. This also underscored the classic trade-off between feature interpretability and representativeness: features such as TF-iDF N-grams are human-interpretable, but might not capture the rich contextual knowledge that deep word embeddings can provide. Secondly, sufficiency is a highly subjective measure with very fuzzy boundaries between levels, making it difficult for the models to fully learn and clearly define. Such ambiguity makes it particularly challenging to achieve high predictive performance. Future research could incorporate formal qualitative research — such as in-depth interviews and focus groups with physicians — to better understand how sufficiency is assessed by human in practice. These insights could help inform better modeling strategies for sufficiency prediction.

We should also note that the experiments in this study are based on VA data from a single setting. In reality, the quality and sufficiency of VA, especially the interviewer-recorded narratives, may vary considerably between interviewers and across settings. Narratives are particularly vulnerable to cultural differences, as well as interviewer bias in interpretation, translation, and recording. Further research is needed to better understand the scope and patterns of information insufficiency in VA across diverse contexts. 
       % sufficiency
\chapter{Conclusion}
\label{conclusion.ch}

\section{Discussion and future directions}

With three studies, this thesis comprehensively demonstrates how narratives can contribute to automated COD classification from VA, leveraging recent advances in language models and machine learning techniques. We found that with narratives alone, fine-tuned PLMs can outperform existing question-only algorithms at both the individual and population levels. And performance can be further boosted with multimodal fusion strategies that integrate knowledge from both narratives and questions. 

Our study reveals that in a multimodal data setting, \textbf{each modality offers unique contributions}. Unstructured narratives can provide valuable information that might not be captured through structured questions, echoing the anecdotal reports from fieldworkers and physicians. Therefore, narratives should be incorporated into the automated COD classification pipeline in future research and practice.  

We also find that \textbf{better VA data is as critical as a better modeling strategy} in real-world applications. The sufficiency of information in VA is strongly associated with the classification accuracy for both physicians and models, as seen in our findings. It is essential to collect and synthesize more high-quality VA with reference deaths from various settings, to support model development and validation for both PLMs and traditional statistical models. In addition, the analyses in this thesis have primarily focused on adult deaths from a single setting with a specific COD profile. Future research should extend these investigations to younger ages and diverse geographic settings, as more data become available, to test the robustness and generalizability of the proposed methods. 

Furthermore, our findings also highlight the need to \textbf{rethink and redesign the VA instrument} to allow for more efficient data collection and improve data quality for more accurate COD estimation. In current practice, the narrative primarily serves as a warm-up for rapport building, and a means of quality control during the interview. Structured questions are asked with fixed order and length, regardless of the information already provided in the narrative. However, as demonstrated in earlier chapters, narrative by itself might already contain adequate information to achieve strong performance for certain causes, with limited marginal gain from adding the full set of structured responses.

This suggests that the narrative should not be treated as redundant information in the survey, but instead used as a starting point for adaptive and targeted questioning. A dynamic VA data collection system, empowered by PLM and ML, can evaluate the sufficiency of already-collected information in real time and actively assist and guide the interview. For example, the interview can wrap up early once information saturation is reached - where additional questions are unlikely to yield new diagnostic value. This approach has the potential to reduce interview length, lower the burden on both respondents and interviewers, and significantly improve cost-effectiveness, particularly in low-resource settings.

Last but not least, we would like to highlight that powerful off-the-shelf PLM/ML tools have been developed in industry and other scientific disciplines. They are widely accessible, user-friendly, with strong supporting communities, can reach decent performance out-of-the-box and can be further enhanced with proper configuration. However, they have not yet been widely introduced, studied, or adopted in social science education and research. We hope our exploration demonstrating their potential could inspire further investigation and integration of these methods into social science, epidemiology, and population health research. 

\section{Conclusion}

Language models and machine learning have been a hot and rapidly evolving area of research. These emerging techniques open up new opportunities to address challenging problems that we have not been able to tackle with traditional methods. With a real-world application with narrative analysis and multimodal fusion in the context of VA, we demonstrate that recent advances in PLM/ML hold significant promise in augmenting research and improving practice in the field of epidemiology, population health, and social science.

       % conclusion 

% Bibliography 
\newpage
\addcontentsline{toc}{front}{Bibliography} %add toc
%\bibliographystyle{unsrt}
%\bibliography{VAbib}

% Appendices 

%\newpage
\appendix
%%%%%%%%%%%
\chapter{Additional methodological details}
\label{app:method}
%%%%%%%%%%%

%\section{Existing studies on verbal autopsy narrative analysis}
% Summary table for existing studies on verbal autopsy anrratives analysis. Table~\ref{tab:nlplitreview}). 

\section{VA narratives data cleaning}
\label{app1:narrative_clean}

\subsection{Invalid data removal}

If the narratives only contain the following information, it's deemed invalid and thus dropped from the analysis: “nothing”, “done”, “folder empty”, “photo cannot be read”, “unclear photo”, “no preview available”, “va number does not match".

\subsection{Correct typo and errors}

We used GPT-3.5-turbo to correct typos, spelling and grammatical errors, using the following prompt:

\begin{Verbatim}[breaklines=true]
"""
Below is a clinical narrative.

Your task is to correct any spelling errors, typos or grammar errors in the narrative. 
Preserve valid medical terms, abbreviations, and phrasing as-is unless there is a clear spelling or typographical error.
If a term or abbreviation is malformed, unclear, or not commonly recognized in clinical usage, correct it to the standard medical form while preserving its intended meaning.
Do not infer missing content or add new information.
Make only minimal and necessary changes to ensure grammatical correctness, clinical clarity, and conformity with standard medical language.

Respond with only the corrected narrative. Respond with original narrative if there is no correction.

""" 
\end{Verbatim}

\subsection{Preprocessing for feature extraction using N-gram TF-iDF}
\label{app1:narrative_preprocess}

Usually no additional data preprocessing is needed for BERT-based models, but for traditional NLP methods, some basic preprocessing could help reduce noise for extracting most useful feature representing the text.

Standard procedure include:

\begin{itemize}
\item Lower casing: change all words to lower cases
\item Remove punctuation: remove punctuation e.g. ",",";"
\item Remove additional white spaces: remove additional white spaces between words
\item Remove stop words: remove common but less distinguishable words, e.g. "the" "a" 
\item Lemmatization: convert words to their base forms, e.g. "bleeding" as "bleed", "coughing" as "cough" 
\end{itemize}

\renewcommand{\arraystretch}{1.2} % Slightly more space between rows
\newcolumntype{L}[1]{>{\raggedright\arraybackslash}p{#1}}

\begin{landscape}
\section{COD grouping and ICD-10 mapping}
\label{tab:icd10}

\begin{longtable}{|L{0.2\textwidth}|L{0.27\textwidth}|L{0.33\textwidth}|L{0.5\textwidth}|}

\caption{Cause of death categories with ICD-10 mapping}  \\
\toprule
\textbf{Level 3 COD} & \textbf{Level 2 COD} & \textbf{Level 1 COD} & \textbf{ICD-10} \\
\midrule
\endfirsthead
\multicolumn{4}{l}{\small\itshape Continued from previous page} \\
\toprule
\textbf{Level 3 COD} & \textbf{Level 2 COD} & \textbf{Level 1 COD} & \textbf{ICD-10} \\
\midrule
\endhead
\midrule
\multicolumn{4}{r}{\small\itshape Continued on next page} \\
\midrule
\endfoot
\bottomrule
\endlastfoot

\multirow{6}{0.2\textwidth}{Communicable diseases}     & HIV/AIDS   & HIV/AIDS  & B20-B24    \\
					\cline{2-4}
                                            & Pulmonary tuberculosis                                            & Pulmonary tuberculosis                              & A15-A16                                                                                                                                                                              \\
                                            \cline{2-4}
                                            & Pneumonia                                                         & Pneumonia                                           & J00-J22                                                                                                                                                                              \\
                                            \cline{2-4}
                                            & Diarrhea                                                          & Diarrhea                                            & A00-A09                                                                                                                                                                              \\
                                            \cline{2-4}
                                            & Malaria                                                           & Malaria                                             & B50-B54                                                                                                                                                                              \\
                                         	\cline{2-4}
                                            & Other and unspecified infections        & Other and unspecified infections   & A40-A41, B05, A39; G00-G05, A33-A35, A37, A90-A99, A17-A19; A20-A38; A42-A89; B00-B19; B25-B49; B55-B99    \\
\hline
\multirow{6}{0.2\textwidth}{Non-communicable diseases}  & \multirow{6}{*}{Neoplasms}                                        & Other and unspecified neoplasms                     & C00-C06                                                                                                                                                                              \\
					  \cline{3-4}																	
                                            &                                                                   & Digestive neoplasms                                 & C15-C26                                                                                                                                                                              \\
                                            \cline{3-4}
                                            &                                                                   & Respiratory neoplasms                               & C30-C39                                                                                                                                                                              \\
                                            \cline{3-4}
                                            &                                                                   & Breast neoplasms                                    & C50                                                                                                                                                                                  \\
                                            \cline{3-4}
                                            &                                                                   & Reproductive neoplasms                         & C51-C58; C60-C63                                                                                                                                                                     \\
                                            \cline{3-4}
                                            &                                                                   & Other and unspecified neoplasms                     & C07-C14; C40-C49; C60-D48                                                                                                                                                            \\
                                            \cline{2-4}
                                            & Diabetes                                                          & Diabetes                                            & E10-E14                                                                                                                                                                              \\
                                            %\cline{2-4}
                                            & \multirow{3}{0.27\textwidth}{Cardiovascular disease}                           & Acute cardiac disease                               & I20-I25                                                                                                                                                                              \\
                                            \cline{3-4}
                                            &                                                                   & Stroke                                              & D57                                                                                                                                                                                  \\
                                            \cline{3-4}
                                            &                                                                   & Other and unspecified cardiac diseases                   & I00-I15; I26-I52; I70-I99                                                                                                                                                   \\
                                            \cline{2-4}
                                            & \multirow{2}{0.25\textwidth}{Chronic pulmonary disease and asthma}            & Chronic obstructive pulmonary diseases                   & J40-J44                                                                                                                                                                              \\
                                            \cline{3-4}
                                            &                                                                   & Asthma                                              & J45-J46                                                                                                                                                                              \\
                                            \cline{2-4}
                                            & Liver cirrhosis                                                   & Liver cirrhosis                                     & K70-K76                                                                                                                                                                              \\
                                            \cline{2-4}
                                            & \multirow{5}{0.25\textwidth}{Other and unspecified non-communicable diseases} & Acute abdomen                                       & R10                                                                                                                                                                                  \\
                                            \cline{3-4}
                                            &                                                                   & Renal failure                                       & N17-N19                                                                                                                                                                              \\
                                            \cline{3-4}
                                            &                                                                   & Epilepsy                                            & G40-G41                                                                                                                                                                              \\
                                            \cline{3-4}
                                            &                                                                   & Other and unspecified non-communicable diseases                           & D50-D89; E00-E07; E15-E35; E40-E46; E50-E90; F00-F99; G06-G37; G50-G99; H00-H95; I60-I69, J30-J39; J47-J99; K00-K31; K35-K38; K40-K93; L00-L99; M00-M99; N00-N16; N20-N99; R00-R09; R11-R94 \\
                                            \cline{3-4}
                                            &                                                                   & Neonatal causes                    & P05-P07, P20-P22, P23-P25, P36, Q00-Q99, P00-P04; P08-P15; P26-P35; P37-P94; P96, P95     \\
\hline
\multirow[t]{6}{*}{Injuries}                  & Traffic / transport accident                    & Traffic / transport accident  & V01-V99  \\
                                            %\cline{2-4}
                                            & Assault                                                           & Assault                                             & X85-Y09                                                                                                                                                                              \\
                                            \cline{2-4}
                                            & Suicide                                                           & Suicide                                             & X60-X84                                                                                                                                                                              \\
                                            \cline{2-4}
                                            & \multirow{3}{0.25\textwidth}{Other and unspecified external causes}            & Accidental fall                                          & W00-W19                                                                                                                                                                              \\
                                            \cline{3-4}
                                            &                                                                   & Accidental drowning and submersion                       & W65-W74                                                                                                                                                                              \\
                                            \cline{3-4}
                                            &                                                                   & Other and unspecified external causes  & X00-X19, X20-X29, X40-X49, X30-X39, S00-T99; W20-W64; W75-W99; X50-X59; Y10-Y98              \\
\hline
Maternal conditions        & Maternal conditions                             & Maternal conditions               & O00, O03-O08, O10-O16, O46; O67; O72, O63-O66, O85; O75.3, O99.0, O71, O01-O02; O20-O45; O47-O62; O68-O70; O73-O84; O86-O99           \\
\hline
Indeterminate                               & Indeterminate                                                     & Indeterminate                                       & R95-R99        

\end{longtable}
\end{landscape}

%More details on strategies converting each question to text can be found \href{https://www.dropbox.com/scl/fi/e53hd5l6c57ql734l4jg9/qdesc.xls?rlkey=wr0st8xlq8wbfxvrxv7o858bl&dl=0}{here}. 

\begin{landscape}

\section{Additional details for multimodal fusion strategies}

\subsection{Descriptions for structured VA questions}
\label{app1.qdesc}

\captionsetup{format=plain,font=small,labelfont=bf, justification=justified, width=\textwidth}

\begin{longtable}{@{}c cc L{1.0\textwidth}@{}}
\caption{Descriptions for structured VA questions} \\
\toprule
\textbf{Indicator} & \textbf{If "yes"} & \textbf{If "no"} & \textbf{Question description} \\
\midrule
\endfirsthead

\multicolumn{4}{@{}l}{\small\itshape Continued from previous page} \\
\toprule
\textbf{Indicator} & \textbf{If "yes"} & \textbf{If "no"} & \textbf{Question description} \\
\midrule
\endhead

\midrule
\multicolumn{4}{r@{}}{\small\itshape Continued on next page} \\
\midrule
\endfoot

\bottomrule
\endlastfoot

i004a     & did      & did not     & die during wet season                                                                                                                                 \\
i004b     & did      & did not     & die during dry season                                                                                                                                 \\
i019a     & was      & -           & male                                                                                                                                                  \\
i019b     & was      & -           & female                                                                                                                                                \\
i022a     & was      & -           & above 65 years old                                                                                                                                    \\
i022b     & was      & -           & 50 to 64 years old                                                                                                                                    \\
i022c     & was      & -           & 15 to 49 years old                                                                                                                                    \\
i022d     & was      & -           & 5 to 14 years old                                                                                                                                     \\
i022e     & was      & -           & 1 to 4 years old                                                                                                                                      \\
i022f     & was      & -           & postneonate                                                                                                                                           \\
i022g     & was      & -           & neonate                                                                                                                                               \\
i022h     & was      & -           & a baby died within 24 hours of birth                                                                                                                  \\
i022i     & was      & -           & a baby died on the second day after birth                                                                                                             \\
i022j     & was      & -           & a baby died within a week                                                                                                                             \\
i022k     & was      & -           & a baby died within a month                                                                                                                            \\
i022l     & was      & -           & a woman aged 12 to 19 years at death                                                                                                                  \\
i022m     & was      & -           & a woman aged 20 to 34 years at death                                                                                                                  \\
i022n     & was      & -           & a woman aged 35 to 49 years at death                                                                                                                  \\
i059o     & was      & was not     & married at the time of death                                                                                                                          \\
i077o     & had      & had no      & injury that let to the death                                                                                                                          \\
i079o     & was      & was not     & injured in a road traffic accident                                                                                                                    \\
i082o     & was      & was not     & injured in a non-road transport accident                                                                                                              \\
i083o     & was      & was not     & injured in a fall                                                                                                                                     \\
i084o     & was      & was not     & poisoned                                                                                                                                              \\
i085o     & was      & was not     & died of drowning                                                                                                                                      \\
i086o     & was      & was not     & injured by venomous animal                                                                                                                            \\
i087o     & was      & was not     & injured by non-venomous animal                                                                                                                        \\
i089o     & was      & was not     & injured by burns or fire                                                                                                                              \\
i090o     & was      & was not     & injured by violence                                                                                                                                   \\
i091o     & was      & was not     & injured by firearm                                                                                                                                    \\
i092o     & was      & was not     & stabbed, cut or pierced                                                                                                                               \\
i093o     & was      & was not     & strangled                                                                                                                                             \\
i094o     & was      & was not     & injured by blunt force                                                                                                                                \\
i095o     & was      & was not     & injured by force of nature                                                                                                                            \\
i096o     & was      & was not     & electrocuted                                                                                                                                          \\
i098o     & was      & was not     & injured accidentally                                                                                                                                  \\
i099o     & was      & was not     & injured by suicide                                                                                                                                    \\
i100o     & was      & was not     & injured by homicide                                                                                                                                   \\
i104o     & had ever & never       & cried after birth                                                                                                                                     \\
i105o     & had      & never       & cried immediately after birth                                                                                                                         \\
i106a     & had      & never       & first cried more than 5 minutes after birth                                                                                                           \\
i107o     & had      & never       & been unable to cry                                                                                                                                    \\
i108a     & had      & never       & stopped crying more than a day before death                                                                                                           \\
i109o     & had ever & never       & moved after birth                                                                                                                                     \\
i110o     & had ever & never       & breathed after birth                                                                                                                                  \\
i111o     & had      & never       & breathed immediately after birth                                                                                                                      \\
i112o     & had      & didn't have & breathing problem                                                                                                                                     \\
i113o     & had      & didn't have & received assistance to breathe at birth                                                                                                               \\
i114o     & had      & didn't have & shown no sign of life                                                                                                                                 \\
i115o     & had      & didn't have & bruises or signs of injury on child's body after the birth                                                                                            \\
i116o     & had      & didn't have & soft, pulpy and discoloured body with the skin peeling away                                                                                           \\
i120a     & had      & didn't have & final illness last less than 3 weeks                                                                                                                  \\
i120b     & had      & didn't have & final illness last at least 3 weeks                                                                                                                   \\
i123o     & did      & did not     & die suddenly                                                                                                                                          \\
i125o     & was      & was not     & diagnosed with tubeculosis                                                                                                                                 \\
i127o     & was      & was not     & diagnosed with HIV                                                                                                                                         \\
i128o     & was      & was not     & diagnosed with positive malaria                                                                                                                            \\
i129o     & was      & was not     & diagnosed with negative malaria                                                                                                                            \\
i130o     & was      & was not     & diagnosed with dengue fever                                                                                                                                \\
i131o     & was      & was not     & diagnosed with measles                                                                                                                                     \\
i132o     & was      & was not     & diagnosed with high blood pressure                                                                                                                         \\
i133o     & was      & was not     & diagnosed with heart disease                                                                                                                               \\
i134o     & was      & was not     & diagnosed with diabetes                                                                                                                                    \\
i135o     & was      & was not     & diagnosed with asthma                                                                                                                                      \\
i136o     & was      & was not     & diagnosed with epilepsy                                                                                                                                    \\
i137o     & was      & was not     & diagnosed with cancer                                                                                                                                      \\
i138o     & was      & was not     & diagnosed with copd                                                                                                                                        \\
i139o     & was      & was not     & diagnosed with dementia                                                                                                                                    \\
i140o     & was      & was not     & diagnosed with depression                                                                                                                                  \\
i141o     & was      & was not     & diagnosed with stroke                                                                                                                                      \\
i142o     & was      & was not     & diagnosed with sickle cell disease                                                                                                                         \\
i143o     & was      & was not     & diagnosed with kidney disease                                                                                                                              \\
i144o     & was      & was not     & diagnosed with liver disease                                                                                                                               \\
i147o     & had      & didn't have & fever                                                                                                                                                 \\
i148a     & had      & didn't have & fever less than one week                                                                                                                              \\
i148b     & had      & didn't have & fever for one to two weeks                                                                                                                                \\
i148c     & had      & didn't have & fever for at least two weeks                                                                                                                              \\
i149o     & had      & didn't have & fever continued until death                                                                                                                            \\
i150a     & had      & didn't have & severe fever                                                                                                                                          \\
i151a     & had      & didn't have & continuous fever                                                                                                                                      \\
i152o     & had      & didn't have & fever with night sweats                                                                                                                               \\
i153o     & had      & didn't have & cough                                                                                                                                                 \\
i154a     & had      & didn't have & cough less than 3 weeks                                                                                                                               \\
i154b     & had      & didn't have & cough at least 3 weeks                                                                                                                                \\
i155o     & had      & didn't have & productive cough                                                                                                                                      \\
i156o     & had      & didn't have & severe cough                                                                                                                                          \\
i157o     & had      & didn't have & blood cough                                                                                                                                           \\
i158o     & had      & didn't have & whooping cough                                                                                                                                        \\
i159o     & had      & didn't have & difficult breathing                                                                                                                                   \\
i161a     & had      & didn't have & difficult breathing at least 3 days                                                                                                                   \\
i165a     & had      & didn't have & continous difficult breathing                                                                                                                         \\
i166o     & had      & didn't have & fast breathing                                                                                                                                        \\
i167a     & had      & didn't have & fast breathing less than 2 weeks                                                                                                                      \\
i167b     & had      & didn't have & fast breathing at least 2 weeks                                                                                                                       \\
i168o     & had      & didn't have & breathlessness                                                                                                                                        \\
i169a     & had      & didn't have & breathlessness less than 2 weeks                                                                                                                      \\
i169b     & had      & didn't have & breathlessness at least 2 weeks                                                                                                                       \\
i170o     & had      & didn't have & limitted functionality due to breathlessness                                                                                                          \\
i171o     & had      & didn't have & breathless while lyting flat                                                                                                                          \\
i172o     & had      & didn't have & lower chest indraw                                                                                                                                    \\
i173a     & had      & didn't have & wheezing grunting                                                                                                                                     \\
i174o     & had      & didn't have & chest pain                                                                                                                                            \\
i175o     & had      & didn't have & severe chest pain                                                                                                                                     \\
i176a     & had      & didn't have & chest pain at least 3 days                                                                                                                            \\
i178a     & had      & didn't have & chest pain at least 30 minutes                                                                                                                        \\
i181o     & had      & didn't have & diarrhoea                                                                                                                                             \\
i182a     & had      & didn't have & diarrhoea less than 2 weeks                                                                                                                           \\
i182b     & had      & didn't have & diarrhoea at least 2 weeks                                                                                                                            \\
i182c     & had      & didn't have & diarrhoea less than 4 weeks                                                                                                                           \\
i183a     & had      & didn't have & at least 4 loose liquid stools                                                                                                                        \\
i184a     & had      & didn't have & loose liquid stools at least 3 days                                                                                                                   \\
i185o     & had      & didn't have & lower chest indraw                                                                                                                                    \\
i186o     & had      & didn't have & blood in the stools                                                                                                                                   \\
i187o     & had      & didn't have & blood in the stools until death                                                                                                                       \\
i188o     & had      & didn't have & vomit                                                                                                                                                 \\
i189o     & had      & didn't have & vomit in the past week                                                                                                                                \\
i190o     & had      & didn't have & vomit at least 3 days                                                                                                                                 \\
i191o     & had      & didn't have & blood in the vomit                                                                                                                                    \\
i192o     & had      & didn't have & black vomit                                                                                                                                           \\
i193o     & had      & didn't have & abdominal problem                                                                                                                                     \\
i194o     & had      & didn't have & abdominal pain                                                                                                                                        \\
i195o     & had      & didn't have & severe abdominal pain                                                                                                                                 \\
i197a     & had      & didn't have & severe abdominal pain less than 2 weeks                                                                                                               \\
i197b     & had      & didn't have & severe abdominal pain at least 2 weeks                                                                                                                \\
i199a     & had      & didn't have & upper abdomen pain                                                                                                                                    \\
i199b     & had      & didn't have & lower abdomen pain                                                                                                                                    \\
i200o     & had      & didn't have & protruding abdomen                                                                                                                                    \\
i201a     & had      & didn't have & protruding abdomen less than 2 weeks                                                                                                                  \\
i201b     & had      & didn't have & protruding abdomen at least 2 weeks                                                                                                                   \\
i203a     & had      & didn't have & rapid onset protruding abdomen                                                                                                                        \\
i204o     & had      & didn't have & mass in the abdomen                                                                                                                                   \\
i205a     & had      & didn't have & mass in the abdomen less than 2 weeks                                                                                                                 \\
i205b     & had      & didn't have & mass in the abdomen at least 2 weeks                                                                                                                  \\
i207o     & had      & didn't have & severe headache                                                                                                                                       \\
i208o     & had      & didn't have & stiff neck                                                                                                                                            \\
i209a     & had      & didn't have & stiff neck less than one week                                                                                                                         \\
i209b     & had      & didn't have & stiff neck at least one week                                                                                                                          \\
i210o     & had      & didn't have & painful neck                                                                                                                                          \\
i211a     & had      & didn't have & painful neck at least one week                                                                                                                        \\
i212o     & had      & didn't have & mental confusion                                                                                                                                      \\
i213o     & had      & didn't have & mental confusion at least 3 months                                                                                                                    \\
i214o     & had      & didn't have & unconscious during the illness                                                                                                                        \\
i215o     & had      & didn't have & unconscious for at least 24 hours before death                                                                                                        \\
i216a     & had      & didn't have & unconscious for at least 6 hours before death                                                                                                         \\
i217o     & had      & didn't have & sudden unconsciousness                                                                                                                                \\
i218o     & had      & didn't have & unconsciousness continued unti death                                                                                                                  \\
i219o     & had      & didn't have & convulsion                                                                                                                                            \\
i220o     & had      & didn't have & genealized convulsion                                                                                                                                 \\
i221a     & had      & didn't have & convulsion for less than 10 minutes                                                                                                                   \\
i221b     & had      & didn't have & convulsion for at least 10 minutes                                                                                                                    \\
i222o     & had      & didn't have & unconscious immediately after the convulsion                                                                                                          \\
i223o     & had      & didn't have & urine problem                                                                                                                                         \\
i224o     & had      & didn't have & stopped urinating                                                                                                                                     \\
i225o     & had      & didn't have & frequent urine                                                                                                                                        \\
i226o     & had      & didn't have & blood urine                                                                                                                                           \\
i227o     & had      & didn't have & sores or ulcers                                                                                                                                       \\
i228o     & had      & didn't have & sores                                                                                                                                                 \\
i229o     & had      & didn't have & sores with clear fluid and/or pus                                                                                                                     \\
i230o     & had      & didn't have & foot ulcer                                                                                                                                            \\
i231o     & had      & didn't have & ulcer ooze pus                                                                                                                                        \\
i232a     & had      & didn't have & ulcer ooze pus for at least 2 weeks                                                                                                                   \\
i233o     & had      & didn't have & skin rash                                                                                                                                             \\
i234a     & had      & didn't have & skin rash for less than one week                                                                                                                      \\
i234b     & had      & didn't have & skin rash for at least one week                                                                                                                       \\
i235a     & had      & didn't have & rash on face                                                                                                                                          \\
i235b     & had      & didn't have & rach on trunk or abdomen                                                                                                                              \\
i235c     & had      & didn't have & rash on the extremities                                                                                                                               \\
i235d     & had      & didn't have & rash everywhere                                                                                                                                       \\
i236o     & had      & didn't have & measles rash                                                                                                                                          \\
i237o     & had      & didn't have & shingles or herpes zoster                                                                                                                             \\
i238o     & had      & didn't have & skin flake off in patches                                                                                                                             \\
i239o     & had      & didn't have & skin turned black                                                                                                                                     \\
i240o     & had      & didn't have & skin turned red                                                                                                                                       \\
i241o     & had      & didn't have & bleed                                                                                                                                                 \\
i242o     & had      & didn't have & bleed from the nose, mouth or anus                                                                                                                    \\
i243o     & had      & didn't have & noticeable weight loss                                                                                                                                \\
i244o     & had      & didn't have & severely thin or wasted                                                                                                                               \\
i245o     & had      & didn't have & whitish rash inside the mouth or on the tongue                                                                                                        \\
i246o     & had      & didn't have & stiffness of the whole body or was unable to open the mouth                                                                                           \\
i247o     & had      & didn't have & puffiness of the face                                                                                                                                 \\
i248a     & had      & didn't have & puffiness of the face for at least one week before death                                                                                              \\
i249o     & had      & didn't have & swollen legs or feet                                                                                                                                  \\
i250a     & had      & didn't have & swelling last for at least 3 days before death                                                                                                        \\
i251o     & had      & didn't have & both feet swollen                                                                                                                                     \\
i252o     & had      & didn't have & general puffiness all over the body                                                                                                                   \\
i253o     & had      & didn't have & lumps                                                                                                                                                 \\
i254o     & had      & didn't have & lumps in the mouth                                                                                                                                    \\
i255o     & had      & didn't have & lumps in the neck                                                                                                                                     \\
i256o     & had      & didn't have & lumps in the armpit                                                                                                                                   \\
i257o     & had      & didn't have & lumps in the groin                                                                                                                                    \\
i258o     & had      & didn't have & paralysed                                                                                                                                             \\
i259o     & had      & didn't have & paralysis of only one side of the body                                                                                                                \\
i260a     & had      & didn't have & right side of the body paralysed                                                                                                                      \\
i260b     & had      & didn't have & left side of the body paralysed                                                                                                                       \\
i260c     & had      & didn't have & lower side of the body paralysed                                                                                                                      \\
i260d     & had      & didn't have & upper side of the body paralysed                                                                                                                      \\
i260e     & had      & didn't have & one leg paralysed                                                                                                                                     \\
i260f     & had      & didn't have & one arm paralysed                                                                                                                                     \\
i260g     & had      & didn't have & entire body paralysed                                                                                                                                 \\
i261o     & had      & didn't have & difficulty swallowing                                                                                                                                 \\
i262a     & had      & didn't have & difficulty swallowing for at least one week before death                                                                                              \\
i263a     & had      & didn't have & difficulty swallowing solids                                                                                                                          \\
i263b     & had      & didn't have & difficulty swallowing liquids                                                                                                                         \\
i264o     & had      & didn't have & pain upon swallowing                                                                                                                                  \\
i265o     & had      & didn't have & yellow discolouration of the eyes                                                                                                                     \\
i266a     & had      & didn't have & yellow discolouration for at least 3 weeks before death                                                                                               \\
i267o     & had      & didn't have & hair change to a reddish or yellowish colour                                                                                                          \\
i268o     & had      & didn't have & have pale palms, eyes or nail beds                                                                                                                    \\
i269o     & had      & didn't have & sunken eyes                                                                                                                                           \\
i270o     & had      & didn't have & drunk a lot more water than usual                                                                                                                     \\
i271o     & had      & didn't have & been able to suckle or bottle-feed within the first 24 hours after birth                                                                              \\
i272o     & had      & didn't have & suckled in a normal way                                                                                                                               \\
i273o     & had      & didn't have & stopped suckling                                                                                                                                      \\
i274a     & had      & didn't have & stopped suckling on the 2nd day of life or later                                                                                                      \\
i275o     & had      & didn't have & convulsions starting within the first 24 hours of life                                                                                                \\
i276o     & had      & didn't have & convulsions starting more than 24 hours after birth                                                                                                   \\
i277o     & had      & didn't have & body become stiff with the back arched backwards                                                                                                      \\
i278o     & had      & didn't have & a bulging or raised fontanelle                                                                                                                        \\
i279o     & had      & didn't have & a sunken fontanelle                                                                                                                                   \\
i281o     & had      & didn't have & become unresponsive or unconscious                                                                                                                    \\
i282o     & had      & didn't have & become unresponsive or unconscious soon after birth within less than 24 hours                                                                         \\
i283o     & had      & didn't have & become unresponsive or unconscious more than 24 hours after birth                                                                                     \\
i284o     & had      & didn't have & become cold to touch                                                                                                                                  \\
i285a     & had      & didn't have & been more than 3 days old when it started feeling cold to touch                                                                                       \\
i286o     & had      & didn't have & become lethargic after a period of normal activity                                                                                                    \\
i287o     & had      & didn't have & redness or discharge from the umbilical cord stump                                                                                                    \\
i288o     & had      & didn't have & skin ulcers or pits                                                                                                                                   \\
i289o     & had      & didn't have & yellow skin, palms (hand) or soles (foot)                                                                                                             \\
i290o     & had      & didn't have & appearred to be healthy and then just die suddenly                                                                                                    \\
i294o     & had      & didn't have & swelling or lump in the breast                                                                                                                        \\
i295o     & had      & didn't have & ulcers (pits) in the breast                                                                                                                           \\
i296o     & had      & didn't have & a period or menstruate                                                                                                                                \\
i297o     & had      & didn't have & excessive vaginal bleeding in between menstrual periods                                                                                               \\
i298o     & had      & didn't have & excessive bleeding                                                                                                                                    \\
i299o     & had      & didn't have & menstrual period stopped naturally because of menopause                                                                                               \\
i300o     & had      & didn't have & vaginal bleeding after cessation of menstruation                                                                                                      \\
i301o     & had      & didn't have & excessive vaginal bleeding in the week prior to death                                                                                                 \\
i302o     & had      & didn't have & her period overdue at the time of death                                                                                                               \\
i303a     & had      & didn't have & period overdue for at least 4 weeks                                                                                                                   \\
i304o     & had      & didn't have & a sharp pain in her abdomen shortly before death                                                                                                      \\
i305o     & had      & didn't have & been pregnant at the time of death                                                                                                                    \\
i306o     & had      & didn't have & died within 6 weeks of delivery, abortion or miscarriage                                                                                              \\
i309o     & had      & didn't have & been pregnant for less than 6 months when she died                                                                                                    \\
i310o     & had      & didn't have & been NEITHER pregnant NOR had recently been pregnant NOR had recently delivered when she died                                                         \\
i312o     & had      & didn't have & died during labour or delivery                                                                                                                        \\
i313o     & had      & didn't have & died after delivering a baby                                                                                                                          \\
i314o     & had      & didn't have & died within 24 hours after delivery                                                                                                                   \\
i315o     & had      & didn't have & died within 6 weeks of childbirth                                                                                                                     \\
i316o     & had      & didn't have & given birth to a live baby within 6 weeks of her death                                                                                                \\
i317o     & had      & didn't have & died during or after a multiple pregnancy                                                                                                             \\
i318o     & had      & didn't have & been breastfeeding the child in the days before death                                                                                                 \\
i319a     & had      & didn't have & died during or after her first pregnancy                                                                                                              \\
i319b     & had      & didn't have & four or more pregnancies before this one                                                                                                              \\
i320o     & had      & didn't have & previous Caesarean section                                                                                                                            \\
i321o     & had      & didn't have & high blood pressure during pregnancy                                                                                                                  \\
i322o     & had      & didn't have & foul smelling vaginal discharge during pregnancy or after delivery                                                                                    \\
i323o     & had      & didn't have & convulsions during the last 3 months of pregnancy                                                                                                     \\
i324o     & had      & didn't have & blurred vision during the last 3 months of pregnancy                                                                                                  \\
i325o     & had      & didn't have & excessive bleeding during pregnancy or shortly after delivery                                                                                         \\
i326o     & had      & didn't have & vaginal bleeding during the first 6 months of pregnancy                                                                                               \\
i327o     & had      & didn't have & vaginal bleeding during the last 3 months of pregnancy but before labour started                                                                      \\
i328o     & had      & didn't have & excessive bleeding during labour, before delivery                                                                                                     \\
i329o     & had      & didn't have & excessive bleeding after delivery or abortion                                                                                                         \\
i330o     & had      & didn't have & delivered the placenta completely                                                                                                                     \\
i331o     & had      & didn't have & delivered an abnormally positioned baby                                                                                                               \\
i332a     & had      & didn't have & her labour lasting longer than 24 hours                                                                                                               \\
i333o     & had      & didn't have & attemptted to terminate the pregnancy                                                                                                                 \\
i334o     & had      & didn't have & recently had a pregnancy that ended in an abortion (spontaneous or induced)                                                                           \\
i335o     & had      & didn't have & died during an abortion                                                                                                                               \\
i336o     & had      & didn't have & died within 6 weeks of having an abortion                                                                                                             \\
i337a     & had      & didn't have & delivered at a health facility or clinic                                                                                                              \\
i337b     & had      & didn't have & delivered at home                                                                                                                                     \\
i337c     & had      & didn't have & delivered elsewhere, not at a health facility nor at home                                                                                             \\
i338o     & had      & didn't have & received professional assistance during the delivery                                                                                                  \\
i340o     & had      & didn't have & an operation to remove her uterus shortly before death                                                                                                \\
i342o     & had      & didn't have & delivery via normal vaginal, without forceps or vacuum                                                                                                \\
i343o     & had      & didn't have & vaginal delivery with forceps or vacuum                                                                                                               \\
i344o     & had      & didn't have & Caesarean section                                                                                                                                     \\
i347o     & had      & didn't have & her baby born more than one month early                                                                                                               \\
i354o     & had      & didn't have & been part of multiple birth                                                                                                                           \\
i355a     & had      & didn't have & been born first in a multiple birth                                                                                                                   \\
i356o     & had      & didn't have & a mother still alive                                                                                                                                  \\
i357o     & had      & didn't have & mother died during or shortly after the delivery                                                                                                      \\
i358a     & had      & didn't have & mother die in the baby's first year of life                                                                                                           \\
i360a     & had      & didn't have & the baby born in a health facility or clinic                                                                                                          \\
i360b     & had      & didn't have & been born at home                                                                                                                                     \\
i360c     & had      & didn't have & been born somewhere else (e.g. on the way to a clinic)                                                                                                \\
i361o     & had      & didn't have & received professional assistance during the delivery                                                                                                  \\
i362o     & had      & didn't have & been of usual size at birth                                                                                                                           \\
i363o     & had      & didn't have & been smaller than normal (weighing under 2.5 kg) at birth                                                                                             \\
i364o     & had      & didn't have & been very much smaller than usual, (weighing under 1 kg) at birth                                                                                     \\
i365o     & had      & didn't have & been larger than normal (weighing over 4.5 kg) at birth                                                                                               \\
i367a     & had      & didn't have & been born during the ninth month (at least 37 weeks) of pregnancy                                                                                     \\
i367b     & had      & didn't have & been born during the eighth month (34 to 37 weeks) of pregnancy                                                                                       \\
i367c     & had      & didn't have & been born before the eighth month (less than 34 weeks) of pregnancy                                                                                   \\
i368o     & had      & didn't have & complications in the late part of the pregnancy (defined as the last 3 months), but before labour                                                     \\
i369o     & had      & didn't have & complications during labour or delivery                                                                                                               \\
i370o     & had      & didn't have & been physically abnormal at time of delivery, for example body part too large or too small, additional growth on body                                 \\
i371o     & had      & didn't have & a swelling or defect on the back                                                                                                                      \\
i372o     & had      & didn't have & a very large head                                                                                                                                     \\
i373o     & had      & didn't have & a very small head                                                                                                                                     \\
i376o     & had      & didn't have & been moving in the last few days before the birth                                                                                                     \\
i377o     & had      & didn't have & stopped moving in the womb before labour started                                                                                                      \\
i382a     & had      & didn't have & been in labor for more than 24 hours                                                                                                                  \\
i383o     & had      & didn't have & a baby born 24 hours or more after the waters broke                                                                                                   \\
i384o     & had      & didn't have & foul smelling liquor when the waters broke                                                                                                            \\
i385a     & had      & didn't have & a green or brown colour liquor when the waters broke                                                                                                  \\
i387o     & had      & didn't have & normal vaginal delivery without forceps or vacuum                                                                                                     \\
i388o     & had      & didn't have & vaginal delivery, with forceps or vacuum                                                                                                              \\
i389o     & had      & didn't have & a Caesarean section                                                                                                                                   \\
i391o     & had      & didn't have & mother received vaccinations since reaching adulthood including during this pregnancy                                                                 \\
i393o     & had      & didn't have & mother receiving tetanus toxoid (TT) vaccine                                                                                                          \\
i394a     & had      & didn't have & been born from the mother's first pregnancy                                                                                                           \\
i394b     & had      & didn't have & a mother having four or more births before this one                                                                                                   \\
i395o     & had      & didn't have & a mother sufferring from fever during labour                                                                                                          \\
i396o     & had      & didn't have & a mother suffer from high blood pressure during the last 3 months of pregnancy, labour or delivery                                                    \\
i397o     & had      & didn't have & a mother with diabetes mellitus                                                                                                                       \\
i398o     & had      & didn't have & a mother having foul smelling vaginal discharge during pregnancy or after delivery                                                                    \\
i399o     & had      & didn't have & a mother suffer from convulsions during the last 3 months of pregnancy, labour or delivery                                                            \\
i400o     & had      & didn't have & a mother suffer from blurred vision during the last 3 months of pregnancy                                                                             \\
i401o     & had      & didn't have & a mother with severe anaemia                                                                                                                          \\
i402o     & had      & didn't have & a mother with vaginal bleeding during the last 3 months of pregnancy but before labour started                                                        \\
i403o     & had      & didn't have & bottom, feet, arm or hand coming out of the vagina before its head                                                                                    \\
i404o     & had      & didn't have & umbilical cord wrapped more than once around the neck at birth                                                                                        \\
i405o     & had      & didn't have & umbilical cord delivered first                                                                                                                        \\
i406o     & had      & didn't have & been blue in colour at birth                                                                                                                          \\
i408o     & had      & didn't have & been growing normally before the illness that led to death                                                                                            \\
i411o     & had      & didn't have & drunk alcohol                                                                                                                                         \\
i412o     & had      & didn't have & used tobacco                                                                                                                                          \\
i413o     & had      & didn't have & smoked tobacco (cigarette, cigar, pipe, etc.)                                                                                                         \\
i414a     & had      & didn't have & used non-smoking tobacco                                                                                                                              \\
i415a     & had      & didn't have & been smoking at least 10 cigarettes daily                                                                                                             \\
i418o     & had      & didn't have & received treatment for the illness that led to death                                                                                                  \\
i419o     & had      & didn't have & received oral rehydration salts                                                                                                                       \\
i420o     & had      & didn't have & received or needed intravenous fluids (drip) treatment                                                                                                \\
i421o     & had      & didn't have & received a blood transfusion                                                                                                                          \\
i422o     & had      & didn't have & treatment/food through a tube passed through the nose                                                                                                 \\
i423o     & had      & didn't have & received injectable antibiotics                                                                                                                       \\
i424o     & had      & didn't have & received antiretroviral therapy (ART)                                                                                                                 \\
i425o     & had      & didn't have & received or needed an operation for the illness                                                                                                       \\
i426o     & had      & didn't have & the operation within 1 month before death                                                                                                             \\
i427o     & had      & didn't have & been discharged from hospital very ill                                                                                                                \\
i428o     & had      & didn't have & received appropriate immunizations                                                                                                                    \\
i450o     & had      & didn't have & travelled to a hospital or health facility in the final days before death                                                                             \\
i451o     & had      & didn't have & used motorised transport to get to the hospital or health facility                                                                                    \\
i452o     & had      & didn't have & problems during admission to the hospital or health facility                                                                                          \\
i453o     & had      & didn't have & problems with the way (s)he was treated (medical treatment, procedures, interpersonal attitudes, respect, dignity) in the hospital or health facility \\
i454o     & had      & didn't have & problems getting medications, or diagnostic tests in the hospital or health facility                                                                  \\
i455o     & had      & didn't have & taken more than 2 hours to get to the nearest hospital or health facility from the deceased's household                                               \\
i456o     & had      & didn't have & doubts about whether medical care was needed in the final days before death                                                                           \\
i457o     & had      & didn't have & used traditional medicine in the final days before death                                                                                              \\
i458o     & had      & didn't have & used a telephone or cell phone to call for help in the final days before death                                                                        \\
i459o     & had      & didn't have & total costs of care and treatment prohibit other household payments over the course of illness                                     \\

\end{longtable}
\end{landscape}

\begin{landscape}

\subsection{Additional results on performance of models}
\label{app1:mm_cod_all}

\captionsetup{format=plain,font=small,labelfont=bf, justification=justified, width=\textwidth}

\begin{longtable}{@{}l L{0.48\textwidth} cccccc@{}}
\caption{Leaderboard: COD classification with unimodal and multimodal approaches}\label{tab:mm_cod_all} \\
\toprule
\textbf{Modality} & \textbf{Model} & \textbf{Accuracy} & \textbf{\thead{\makecell{Weighted \\ F1}}} & \textbf{\thead{\makecell{Weighted \\ Precision}}} & \textbf{\thead{\makecell{Weighted \\ Recall}}} & \textbf{\thead{\makecell{CSMF \\ Accuracy}}} & \textbf{\thead{\makecell{CCCSMF \\ Accuracy}}} \\
\midrule
\endfirsthead

\multicolumn{8}{@{}l}{\small\itshape Continued from previous page} \\
\toprule
\textbf{Modality} & \textbf{Model} & \textbf{Accuracy} & \textbf{\thead{\makecell{Weighted \\ F1}}} & \textbf{\thead{\makecell{Weighted \\ Precision}}} & \textbf{\thead{\makecell{Weighted \\ Recall}}} & \textbf{\thead{\makecell{CSMF \\ Accuracy}}} & \textbf{\thead{\makecell{CCCSMF \\ Accuracy}}} \\
\midrule
\endhead

\midrule
\multicolumn{8}{r@{}}{\small\itshape Continued on next page} \\
\midrule
\endfoot

\bottomrule
\endlastfoot

Multimodal & C - Decision-level fusion (Soft voting)                      & \textbf{0.695}    & \textbf{0.681}       & \textbf{0.680}              & \textbf{0.695}           & 0.923         & 0.790           \\
Multimodal & A - Data-level fusion (Longformer)                           & 0.683    & 0.669       & 0.666              & 0.683           & 0.902         & 0.732           \\
Multimodal & A - Data-level fusion (BigBird)                              & 0.675    & 0.664       & 0.663              & 0.675           & 0.918         & 0.776           \\
Multimodal & C - Decision-level fusion (Stacking ensemble with KNN)       & 0.674    & 0.666       & 0.665              & 0.674           & 0.950         & 0.863           \\
Multimodal & A - Data-level fusion (Clinical Longformer)                  & 0.674    & 0.664       & 0.664              & 0.674           & 0.928         & 0.803           \\
Question   & LightGBM                                                     & 0.665    & 0.654       & 0.665              & 0.665           & 0.899         & 0.727           \\
Question   & XGBoost                                                      & 0.662    & 0.651       & 0.659              & 0.662           & 0.898         & 0.724           \\
Question   & CatBoost                                                     & 0.662    & 0.652       & 0.662              & 0.662           & 0.897         & 0.721           \\
Question   & GBDT                                                         & 0.657    & 0.645       & 0.651              & 0.657           & 0.903         & 0.735           \\
Multimodal & C - Decision-level fusion (Stacking ensemble with SVM)       & 0.650    & 0.646       & 0.646              & 0.650           & 0.954         & 0.874           \\
Question   & MLP                                                          & 0.648    & 0.636       & 0.644              & 0.648           & 0.907         & 0.746           \\
Narrative  & BioClinicalBERT                                              & 0.646    & 0.638       & 0.646              & 0.646           & 0.937         & 0.828           \\
Multimodal & B - Feature-level fusion (AutoMM with RoBERTa-PM backbone)   & 0.646    & 0.640       & 0.647              & 0.646           & 0.933         & 0.817           \\
Narrative  & BlueBERT                                                     & 0.642    & 0.635       & 0.646              & 0.642           & 0.942         & 0.842           \\
Narrative  & RoBERTa-PM                                                   & 0.641    & 0.636       & 0.643              & 0.641           & 0.955         & 0.877           \\
Narrative  & BERT                                                         & 0.631    & 0.623       & 0.618              & 0.631           & \textbf{0.958}         & \textbf{0.885}           \\
Multimodal & C - Decision-level fusion (Stacking ensemble with AutoGluon) & 0.631    & 0.628       & 0.633              & 0.631           & 0.948         & 0.858           \\
Narrative  & BioMedBERT                                                   & 0.624    & 0.617       & 0.621              & 0.624           & 0.955         & 0.877           \\
Narrative  & ClinicalBERT                                                 & 0.612    & 0.601       & 0.604              & 0.612           & 0.946         & 0.853           \\
Question   & InSilicoVA                                                   & 0.470    & 0.465       & 0.497              & 0.470           & 0.819         & 0.508          \\

\hline
\multicolumn{8}{l}{\footnotesize{Note: KNN: K-Nearest Neighbor; GBDT: Gradient-Boosted Decision Trees; SVM: Support Vector Machine; MLP: Multilayer Perceptron.}} 
\end{longtable}
\end{landscape}

%\section{Additional results from sensitivity analysis}

%This section includes additional results from sensitivity analysis that are not included in the main text. 

\chapter{Additional figures}
\label{app:figs}

This appendix includes additional figures of the results.

\section{Additional figures for Chapter~\ref{plmva.ch}}

\subsection{Confusion matrices for narrative-based predictions and PCVA}
\label{fig:plm_confusion}

Below are confusion matrices highlighting the misclassification by models with level 2 COD categories. 

%models = ['BioClinicalBERT', 'BERT', 'BioMedBERT', 'BlueBERT', 'ClinicalBERT', 'RoBERTa']

\begin{figure}
\includegraphics[width=6in]{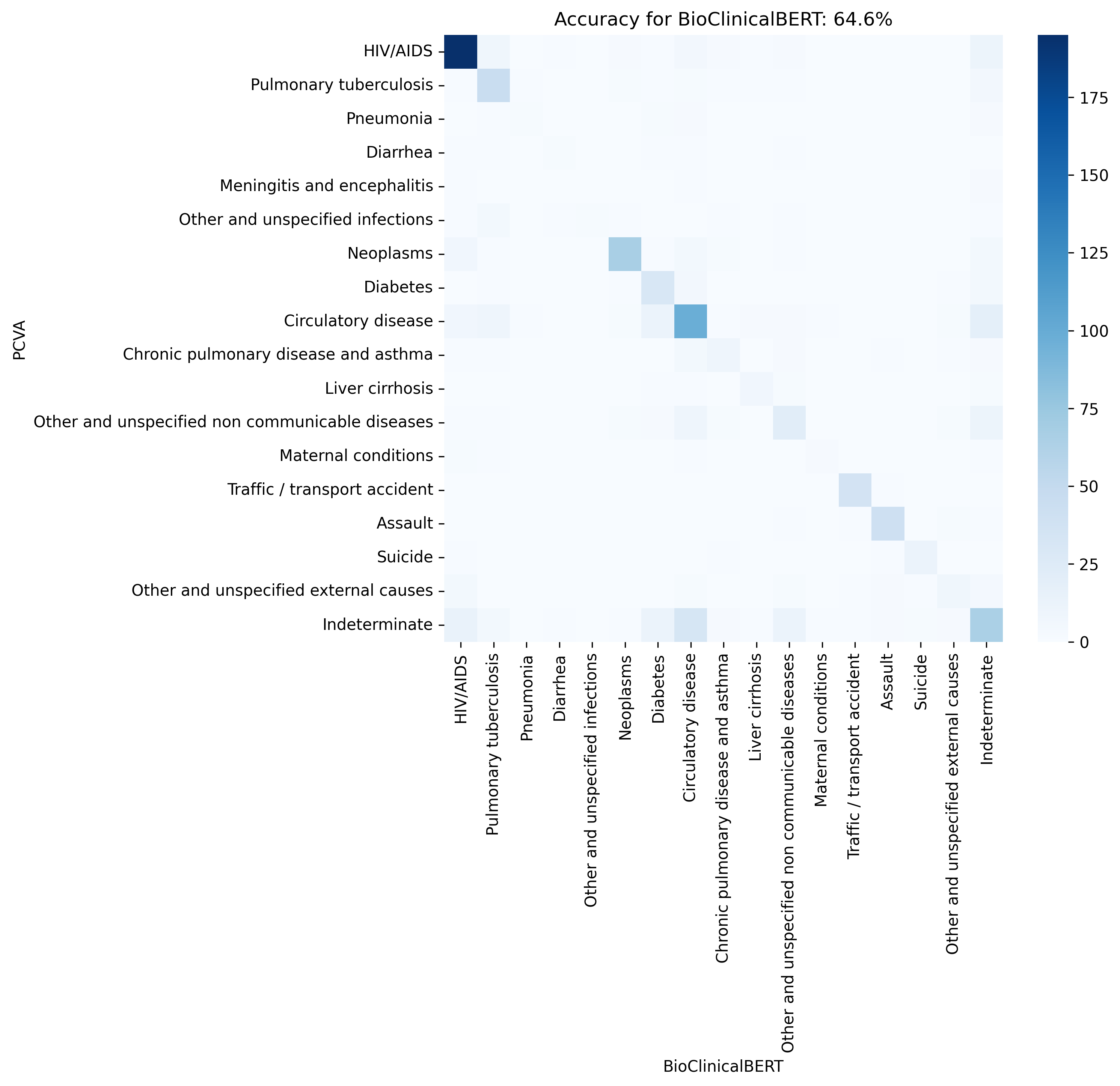}
\caption{Confusion matrices for BioClinicalBERT}
%\label{fig:}
\end{figure}

\begin{figure}
\begin{center}
\includegraphics[width=6in]{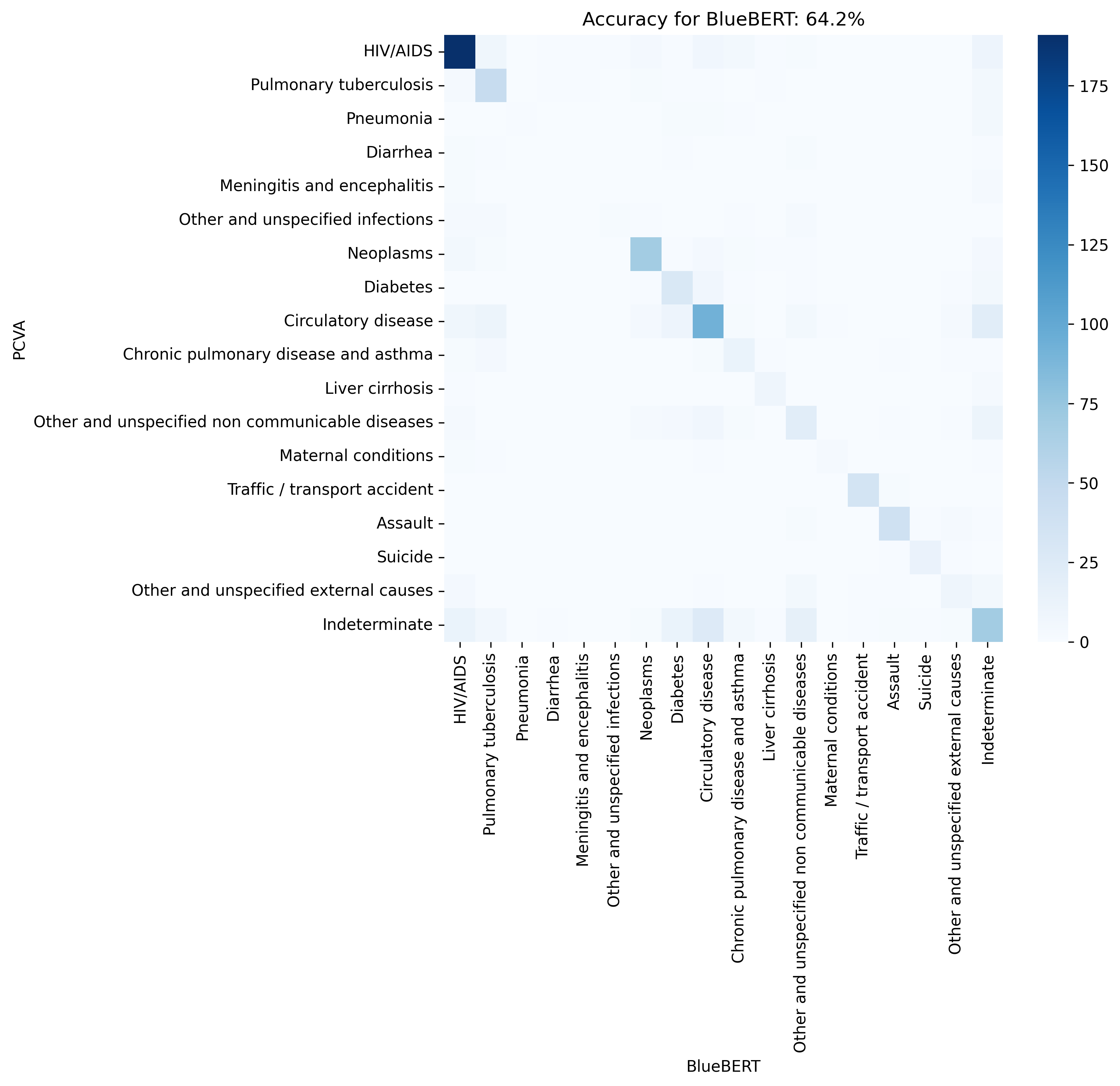}
\caption{Confusion matrices for BlueBERT}
%\label{fig:}
\end{center}
\end{figure}

\begin{figure}
\begin{center}
\includegraphics[width=6in]{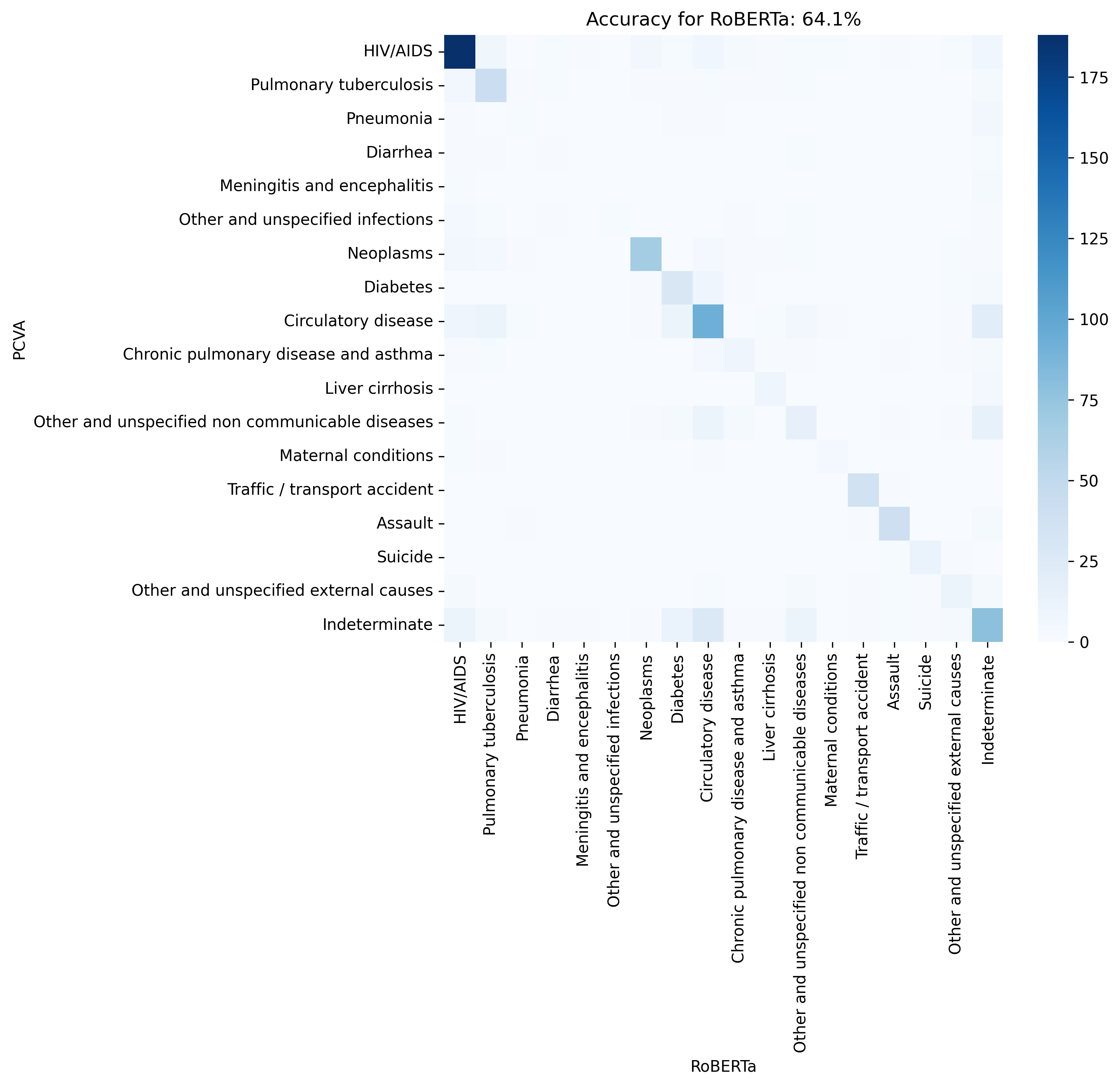}
\caption{Confusion matrices for RoBERTa-PM}
%\label{fig:}
\end{center}
\end{figure}

\begin{figure}
\begin{center}
\includegraphics[width=6in]{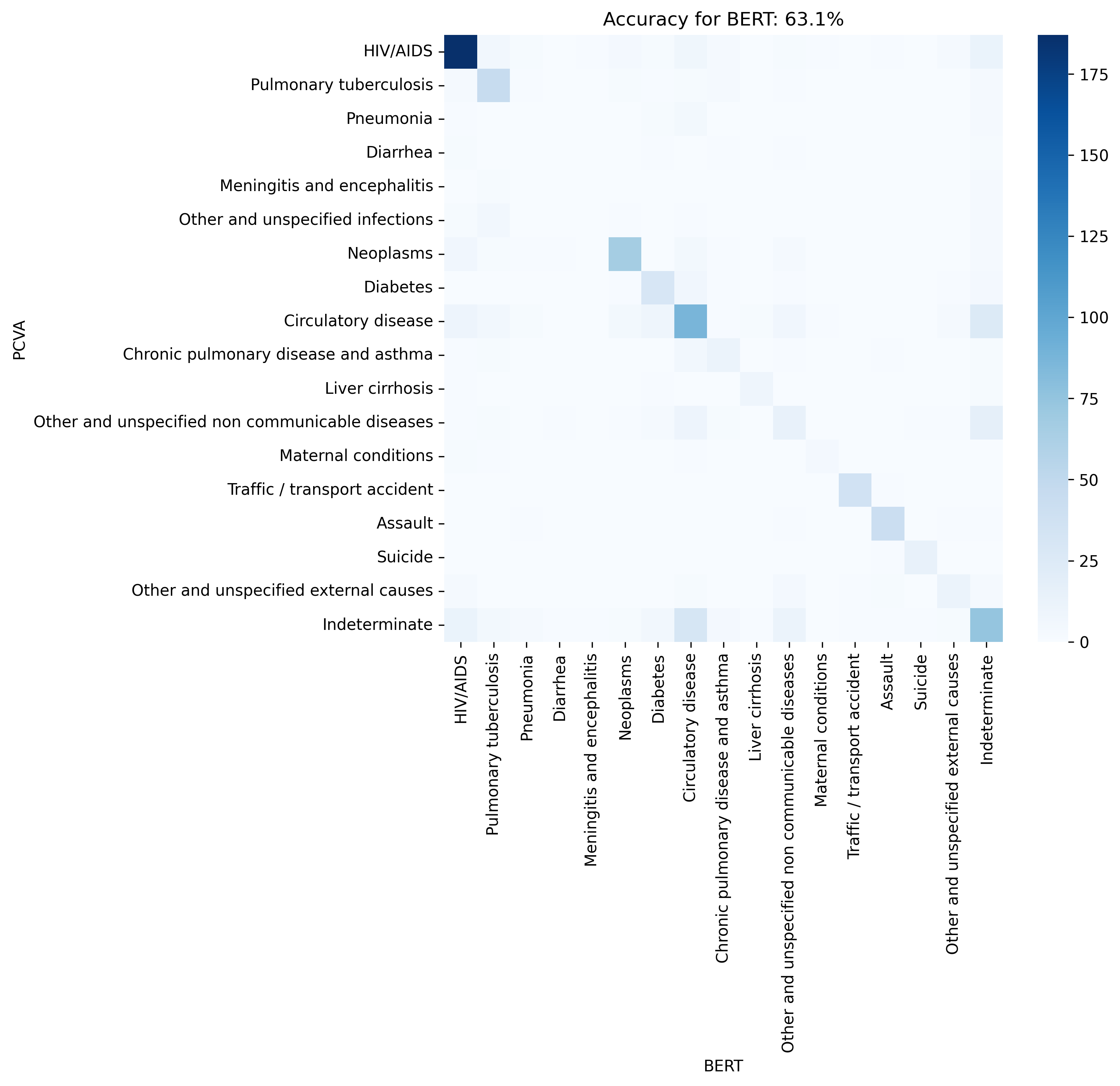}
\caption{Confusion matrices for BERT}
%\label{fig:}
\end{center}
\end{figure}

\begin{figure}
\begin{center}
\includegraphics[width=6in]{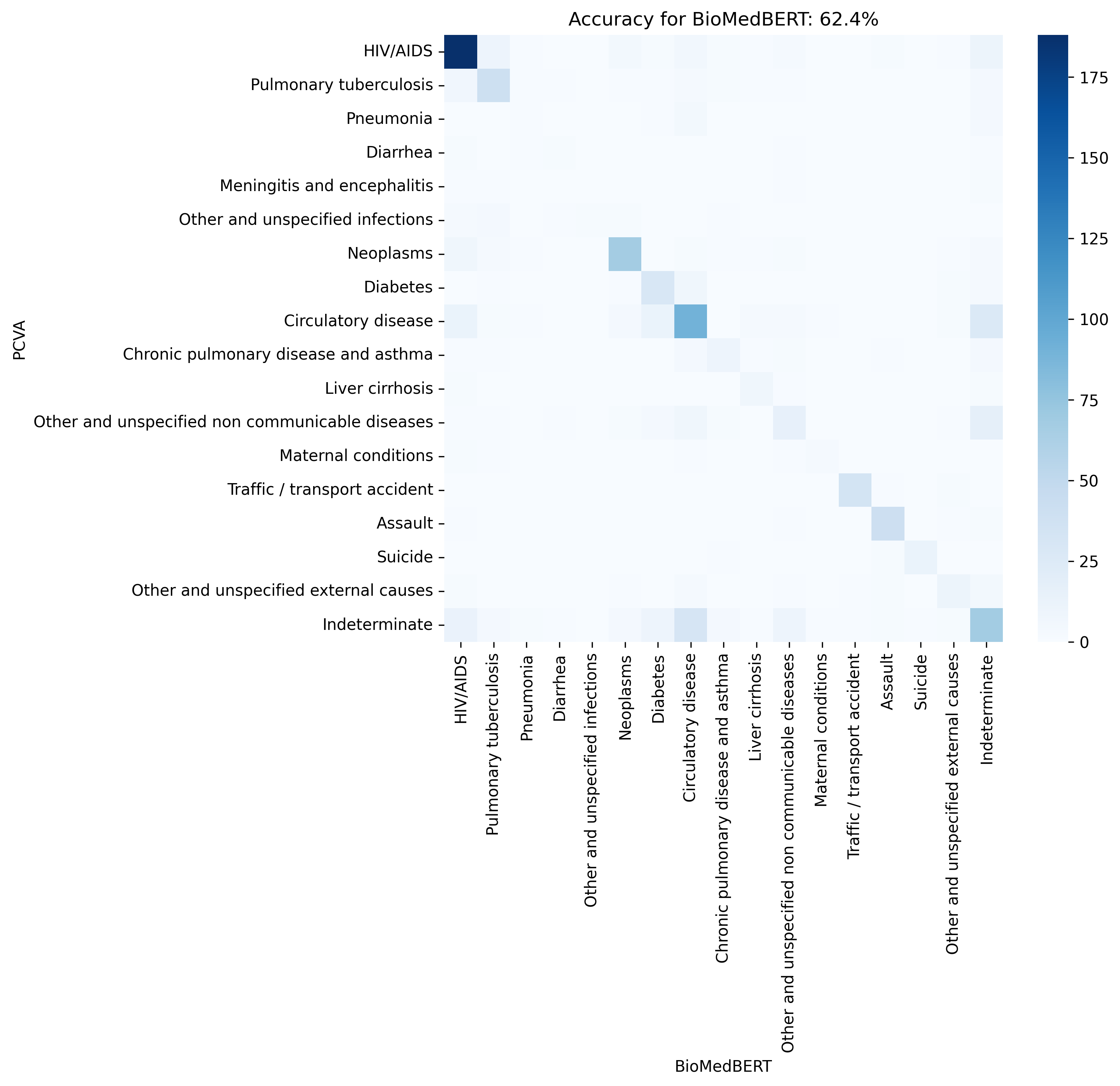}
\caption{Confusion matrices for BioMedBERT}
%\label{fig:}
\end{center}
\end{figure}

\begin{figure}
\begin{center}
\includegraphics[width=6in]{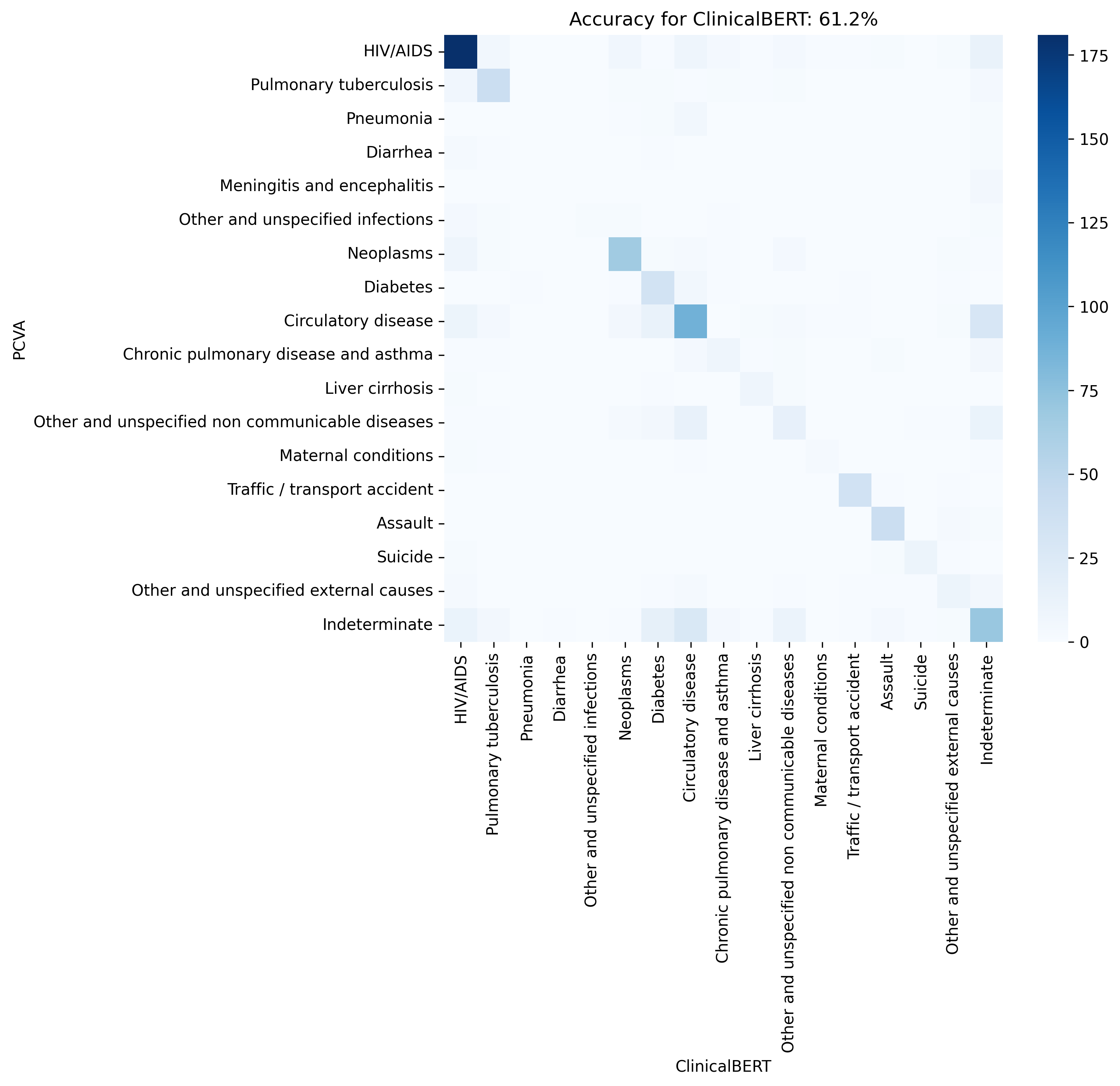}
\caption{Confusion matrices for ClinicalBERT}
%\label{fig:}
\end{center}
\end{figure}

\newpage

\subsection{Confusion matrices by COD grouping level: BioClinicalBERT vs PCVA}
\label{fig:plm_confusion_cod}

Below are confusion matrices highlighting the misclassification between BioClinicalBERT and PCVA, with different COD grouping level. 

\begin{figure}
\includegraphics[width=6in]{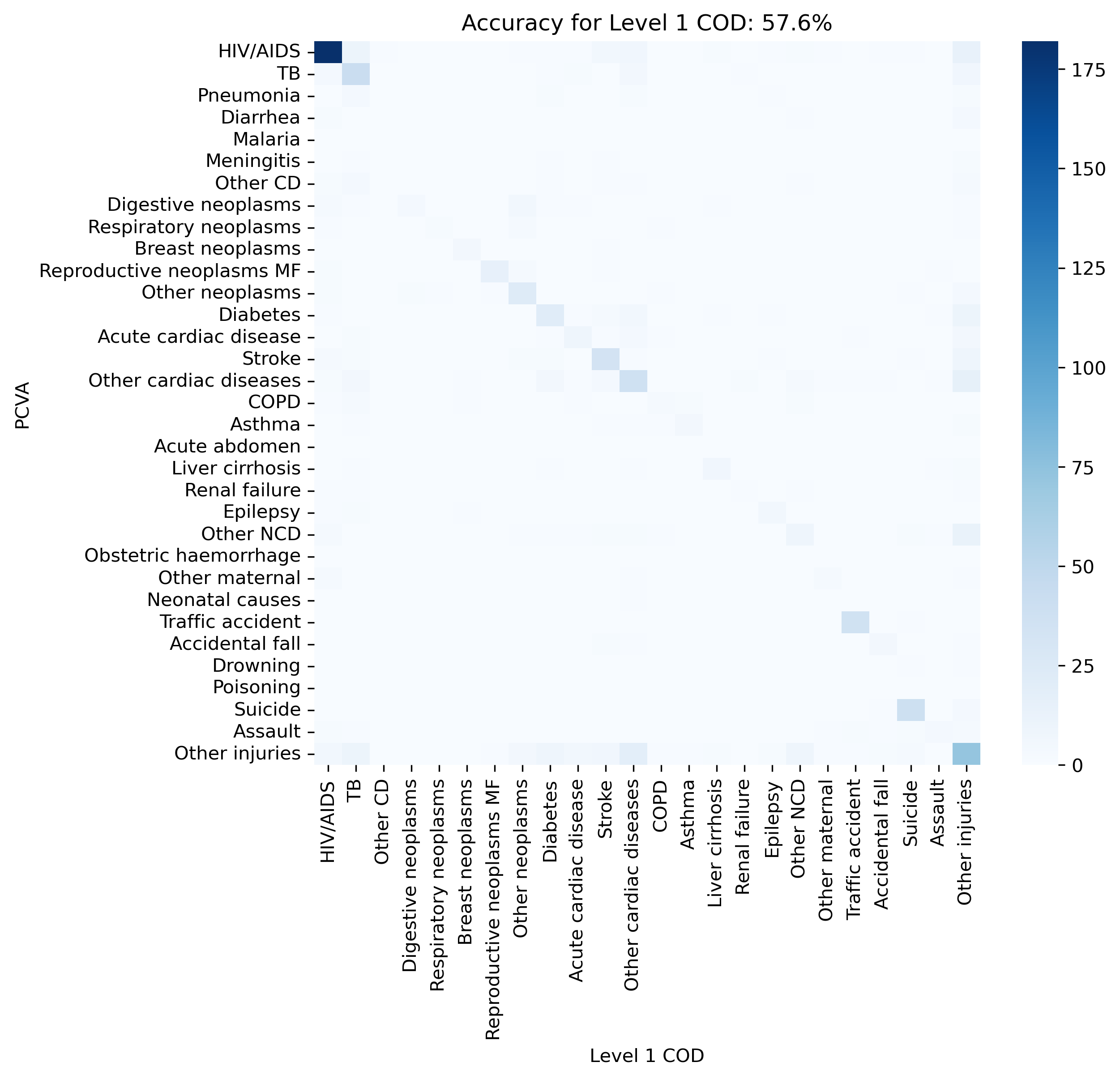}
\caption{Confusion matrices for Level 1 COD - BioClinicalBERT vs PCVA}
\end{figure}

\begin{figure}
\includegraphics[width=6in]{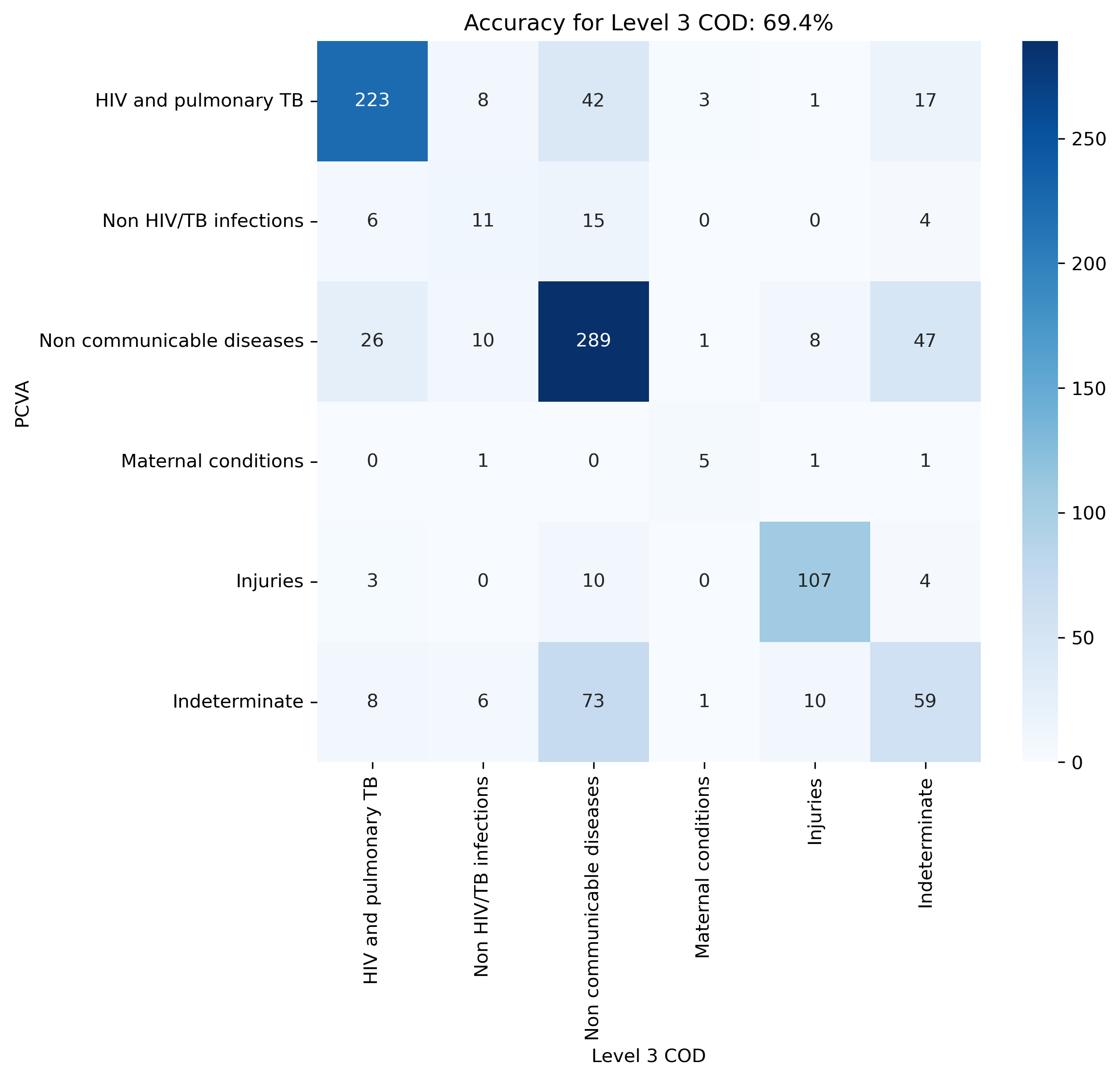}
\caption{Confusion matrices for Level 3 COD - BioClinicalBERT vs PCVA}
\end{figure}

\newpage

\subsection[Classification accuracy by training data size]{Classification accuracy by training data size - BioClinicalBERT as example}
\label{fig:plm_sa_samplesize_heatmap}

\begin{figure}[H]
\includegraphics[width=6in]{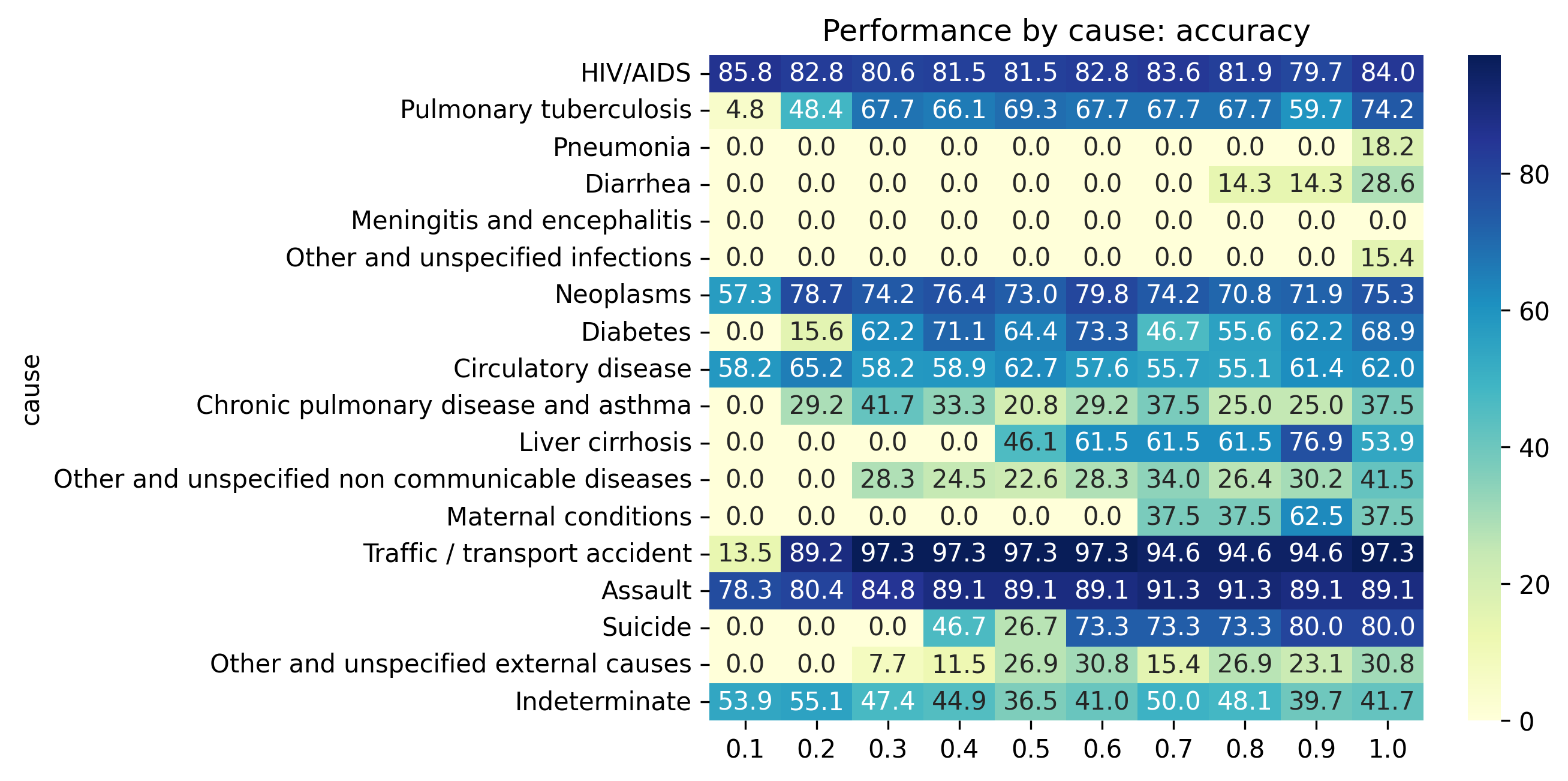}
\caption{Classification accuracy by training data size}
\caption*{\textit{Note:} Training data size as ratio of sampled size out of full training data}
\end{figure}

\newpage

\section{Additional figures for Chapter~\ref{multimodal.ch}}

\subsection{Confusion matrices for multimodal predictions and PCVA}
\label{fig:plm_confusion}

Below are confusion matrices highlighting the misclassification by models with level 2 COD categories. 

\begin{figure}
\includegraphics[width=6in]{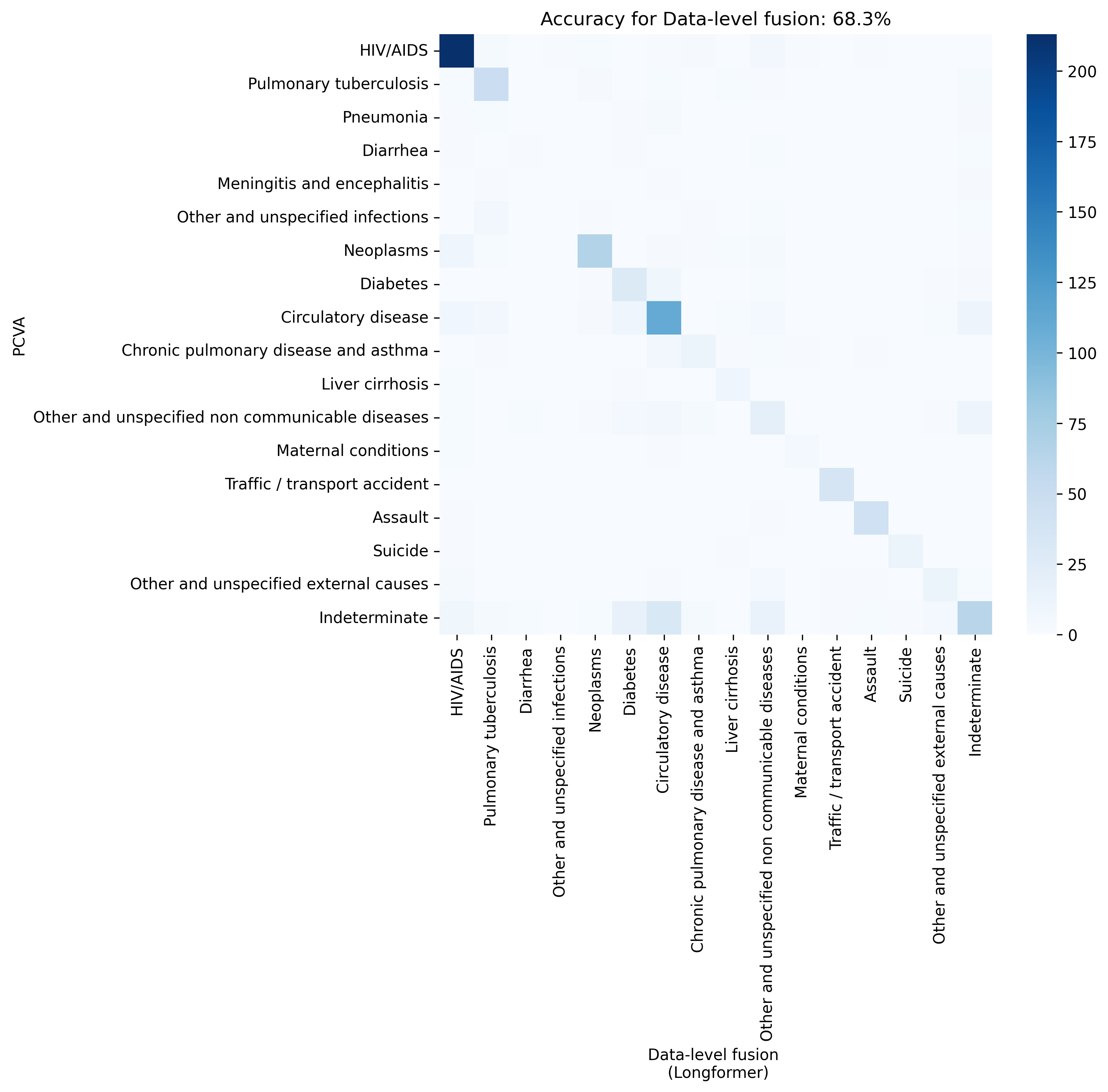}
\caption{Confusion matrices for data-level fusion}
\caption*{\textit{Note:} Data-level fusion with LongFormer.}
\end{figure}

\begin{figure}
\includegraphics[width=6in]{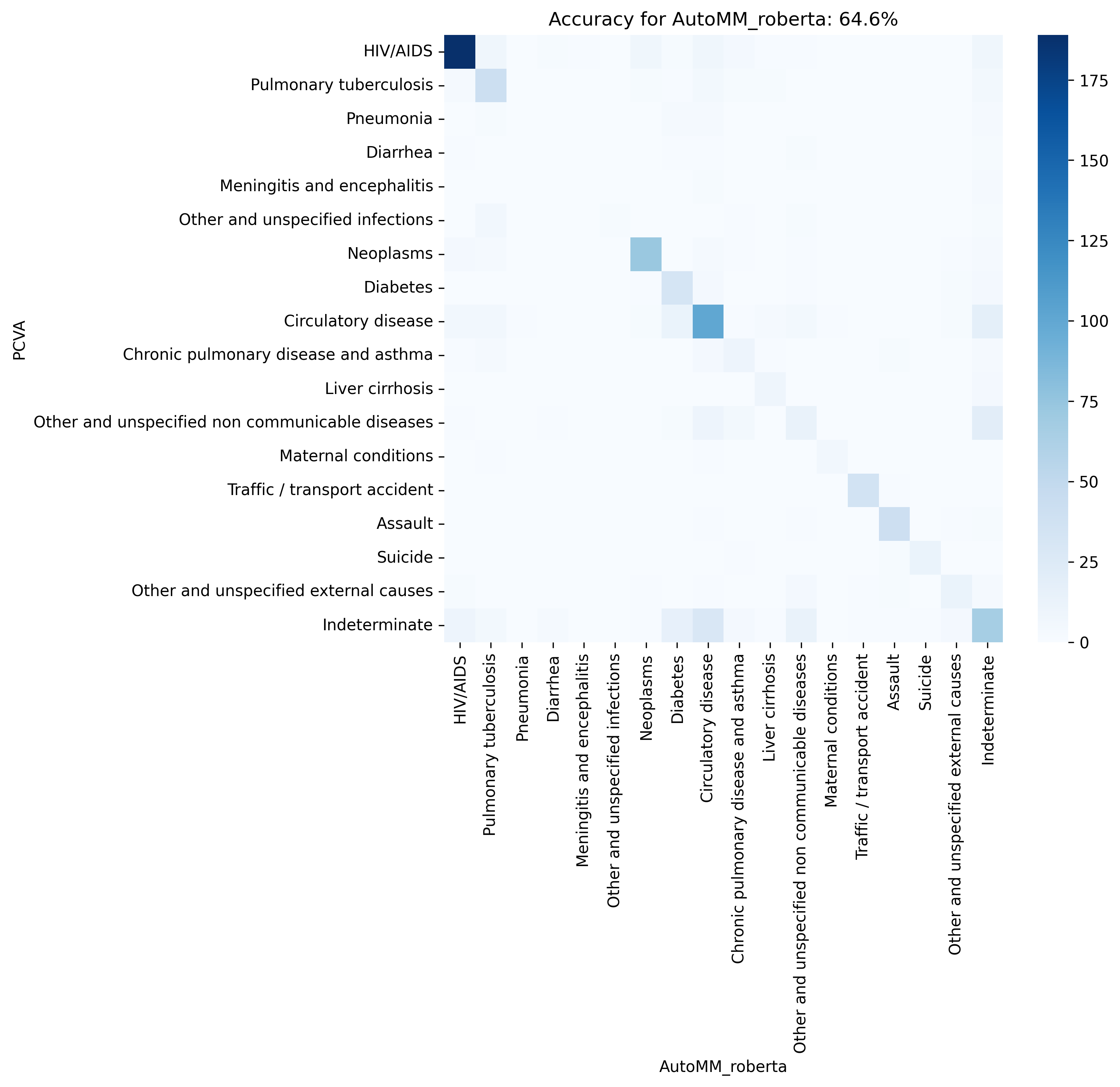}
\caption{Confusion matrices for feature-level fusion}
\caption*{\textit{Note:} Feature-level fusion with AutoMM with RoBERTaPM backbone.}
\end{figure}

\begin{figure}
\includegraphics[width=6in]{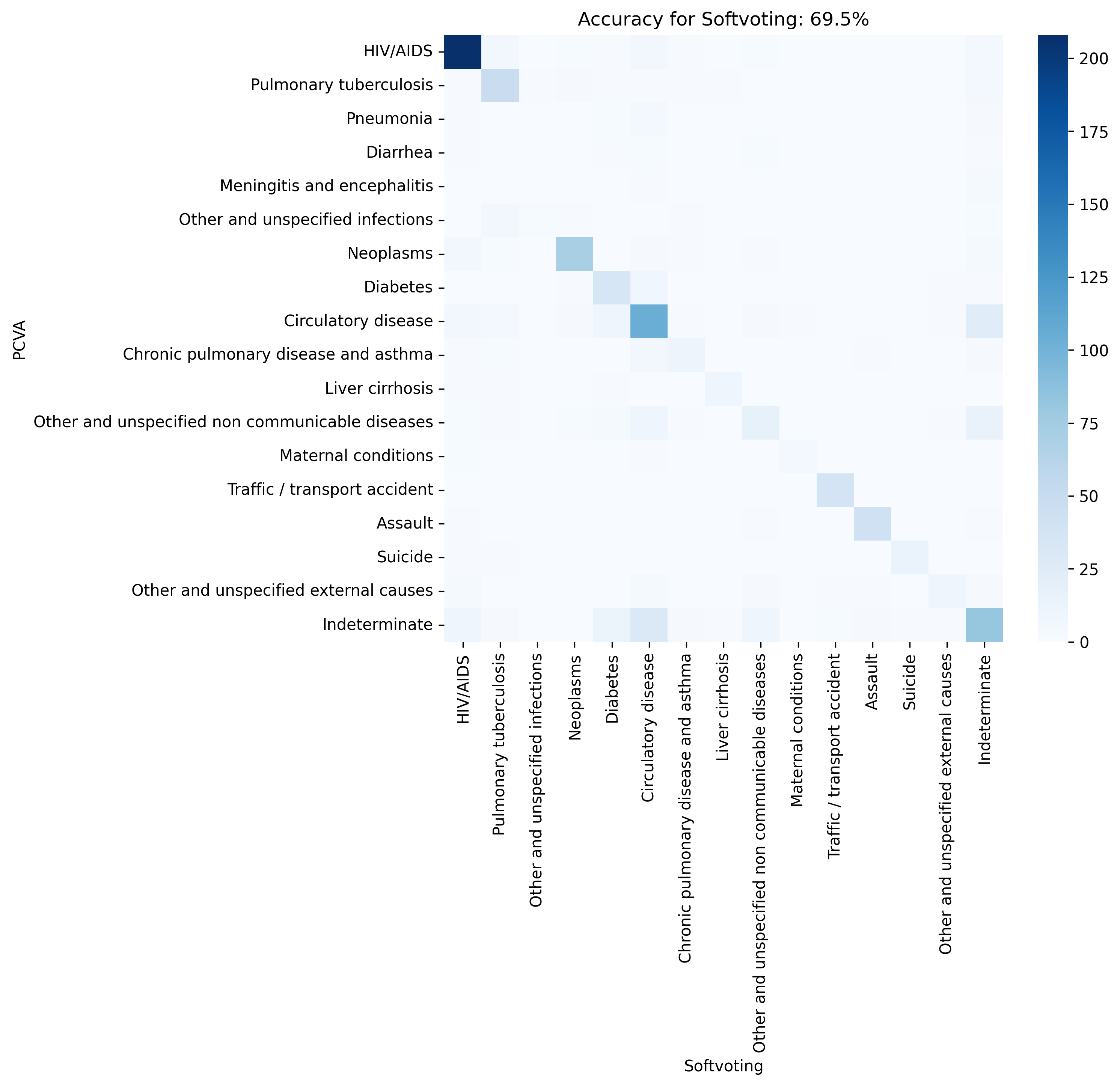}
\caption{Confusion matrices for decision-level fusion: soft voting ensemble}
\caption*{\textit{Note:} Decision-level fusion with soft voting ensemble.}
\end{figure}

\begin{figure}
\includegraphics[width=6in]{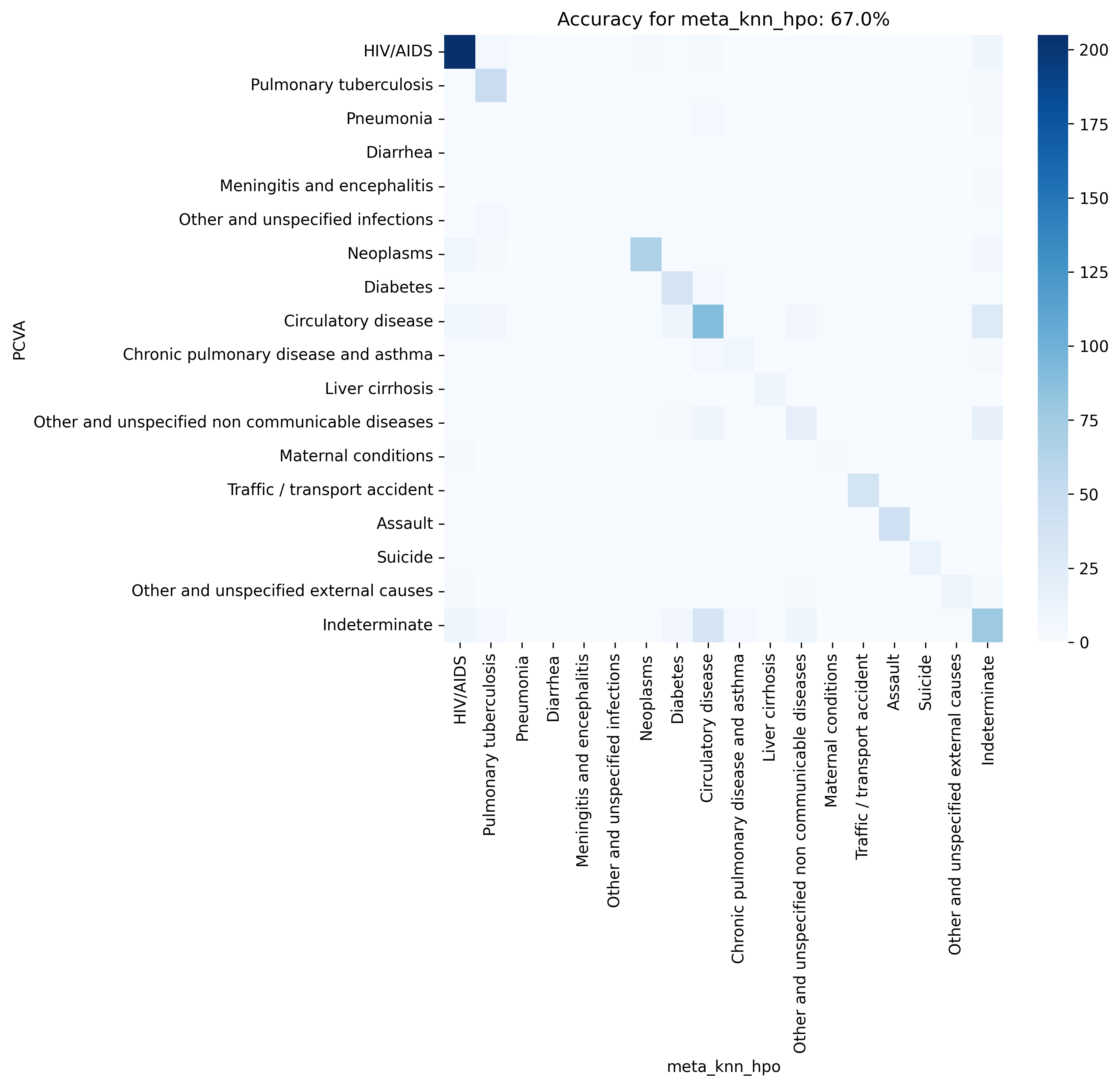}
\caption{Confusion matrices for decision-level fusion: stacking ensemble}
\caption*{\textit{Note:} Decision-level fusion with stacking ensemble with KNN as meta-learner.}
\end{figure}

\newpage 

\subsection[Classification accuracy by training data size]{Classification accuracy by training data size - data-level fusion with Longformer as example}
\label{fig:ensem_sa_samplesize_heatmap}

\begin{figure}[H]
\includegraphics[width=6in]{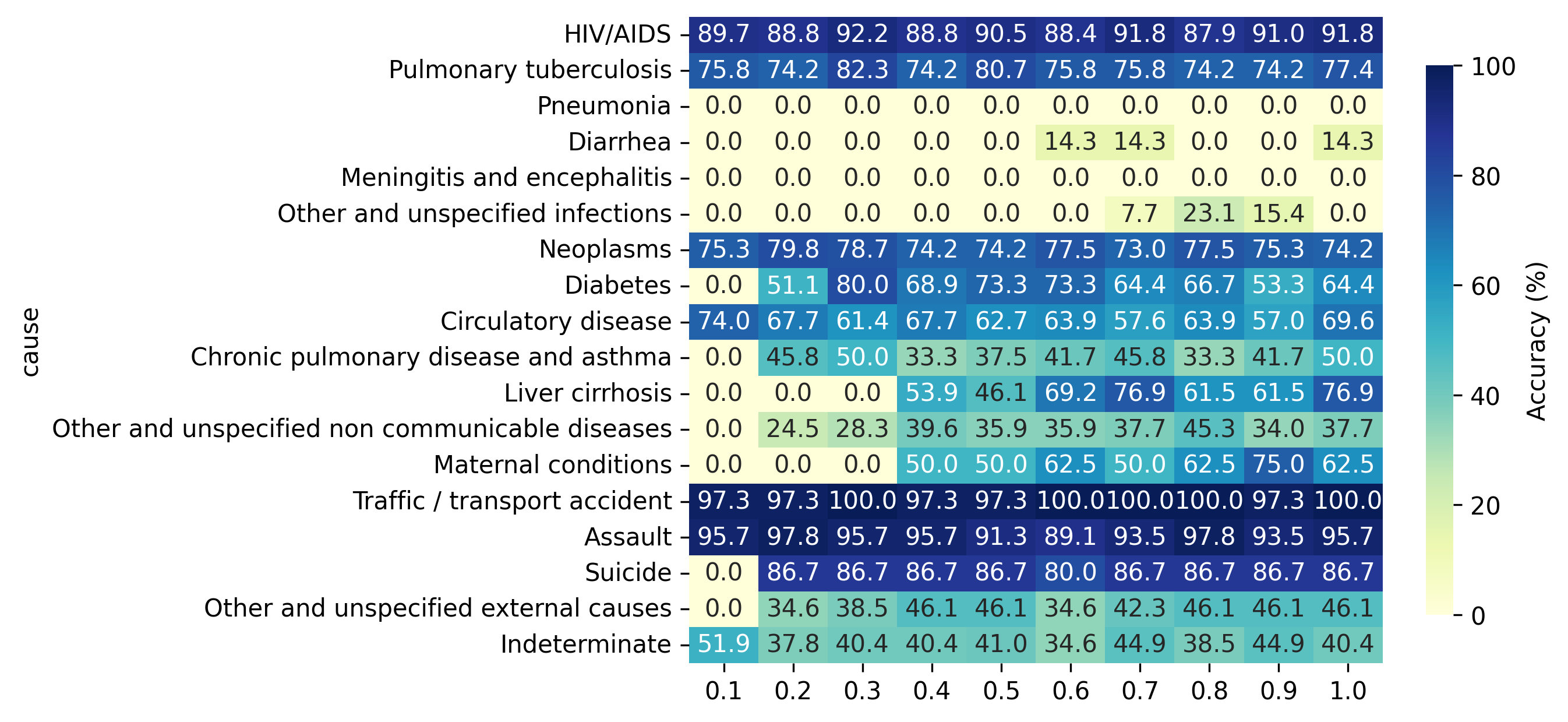}
\caption{Classification accuracy by training data size}
\caption*{\textit{Note:} Training data size as ratio of sampled size out of full training data}
\end{figure}

\newpage

\section{Additional figures for Chapter~\ref{sufficiency.ch}}

\subsection{Confusion matrices comparing PCVA against UCMR, by sufficiency level}
\label{app2:suff_cm_ucmr}

\begin{figure}
\begin{center}
\includegraphics[width=6in]{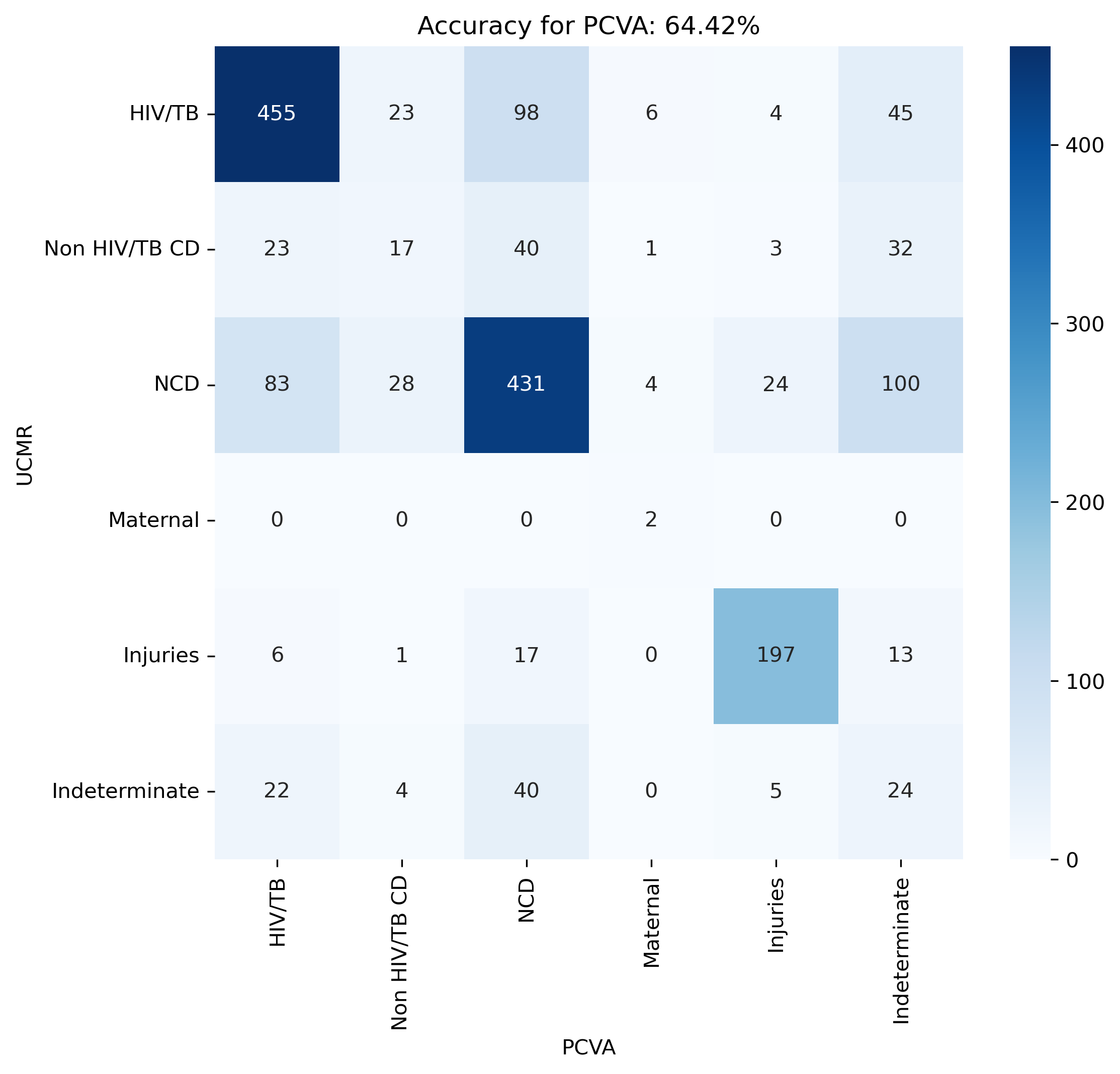}
\caption{Confusion matrices for Level 3 COD - PCVA vs UCMR}
%\label{fig:}
\end{center}
\end{figure}

\begin{figure}
\begin{center}
\includegraphics[width=6in]{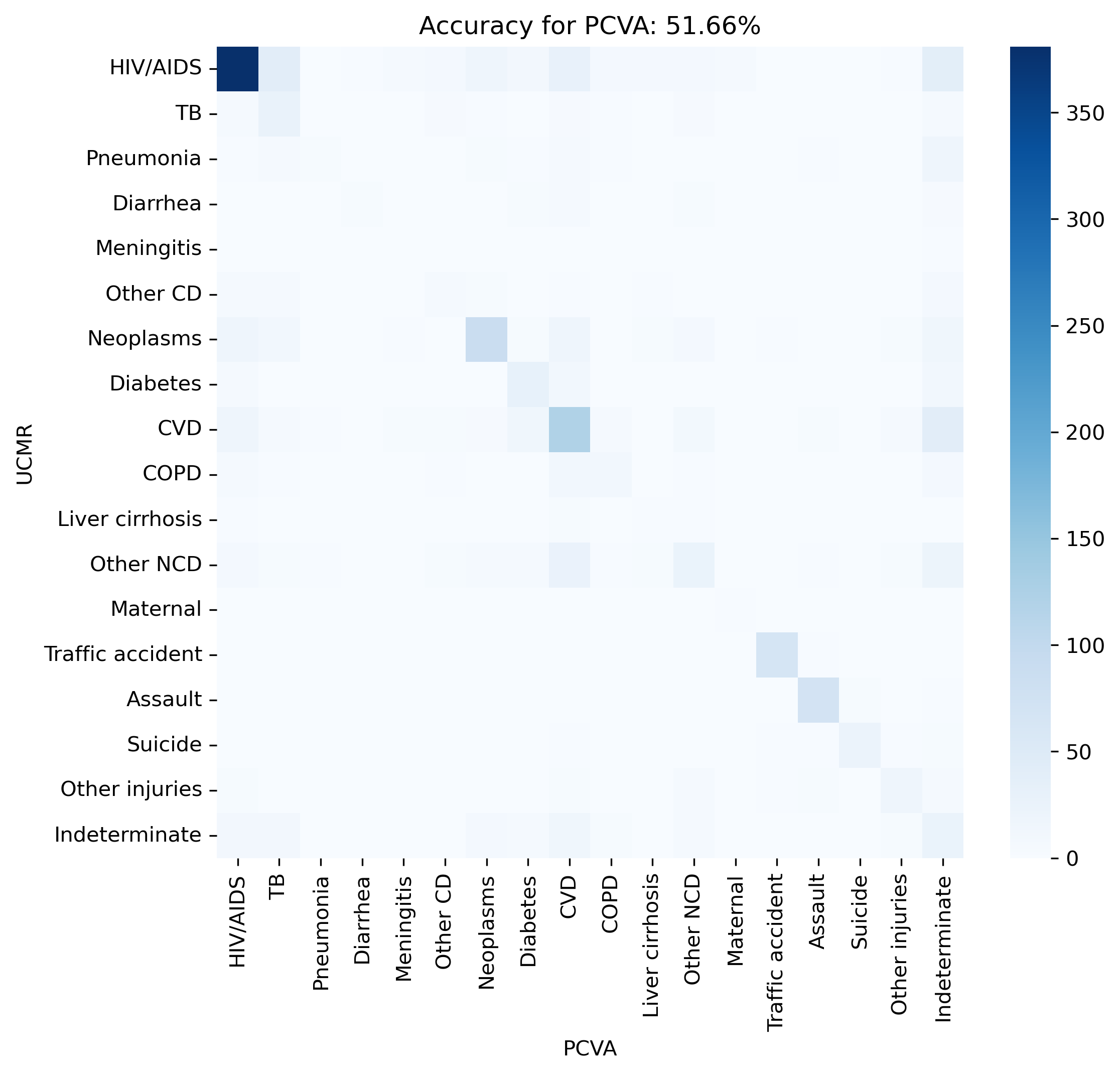}
\caption{Confusion matrices for Level 2 COD - PCVA vs UCMR}
%\label{fig:}
\end{center}
\end{figure}

\begin{figure}
\begin{center}
\includegraphics[width=6in]{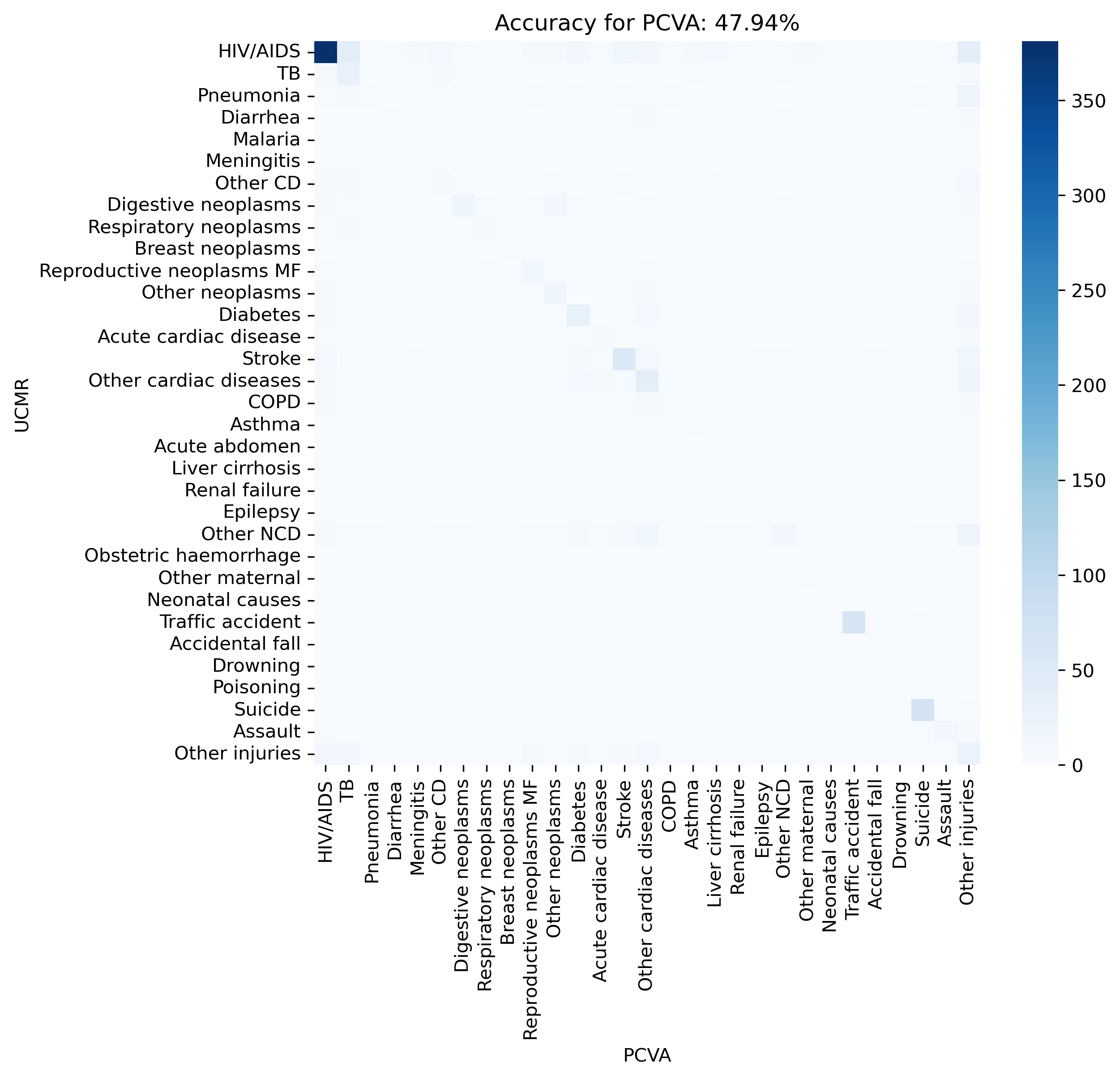}
\caption{Confusion matrices for Level 1 COD - PCVA vs UCMR}
%\label{fig:}
\end{center}
\end{figure}

\newpage

\subsection{Top 10 important features in sufficiency prediction models}
\label{app2:suff_pred_fi}

\begin{figure}[H]
\begin{center}
\includegraphics[width=6in]{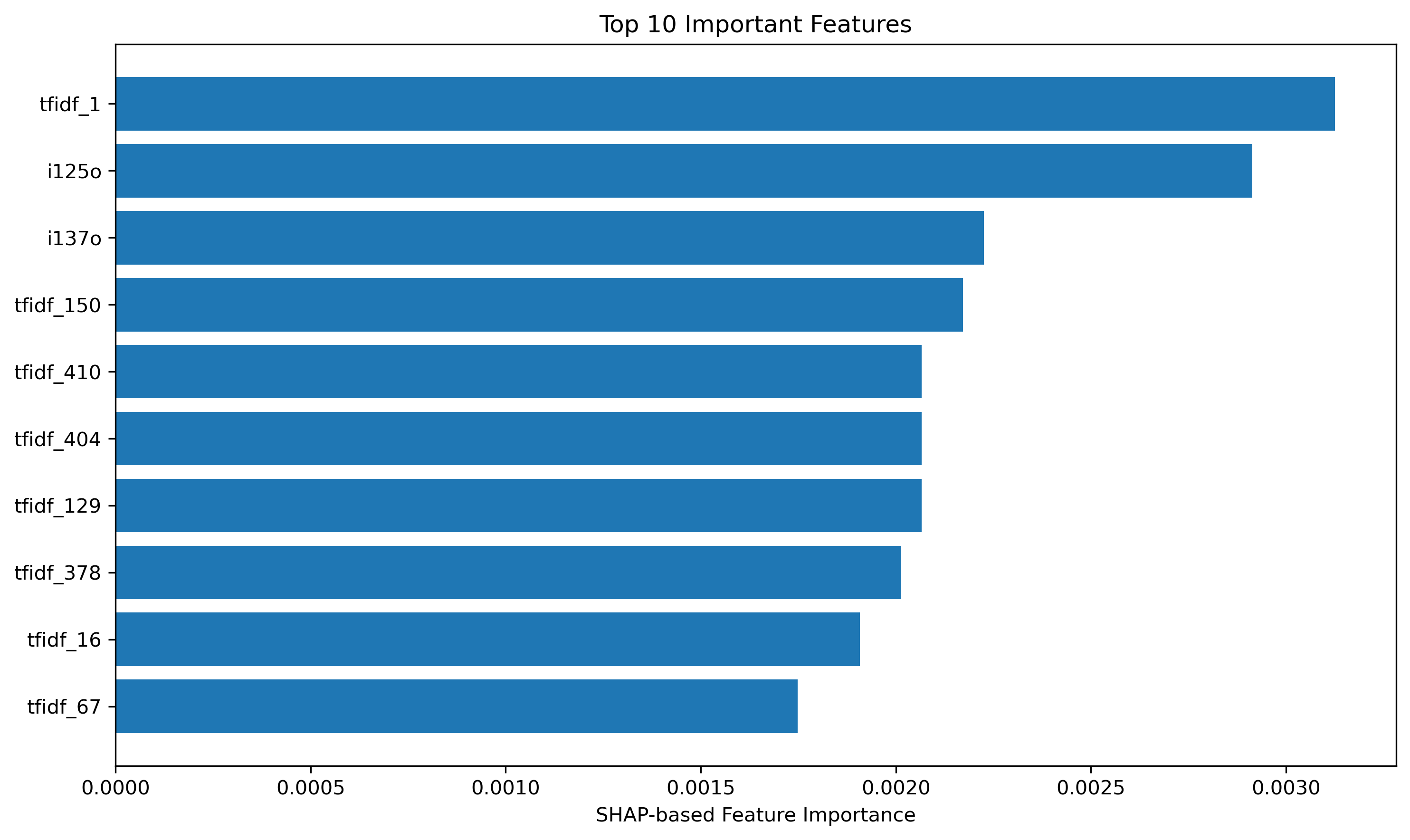}
\caption{Top 10 important features in multimodal sufficiency prediction}
\label{fig:suff_pred_fi_mm}
\end{center}
\end{figure}

\begin{landscape}

\begin{figure}
\begin{center}
\includegraphics[width=8in]{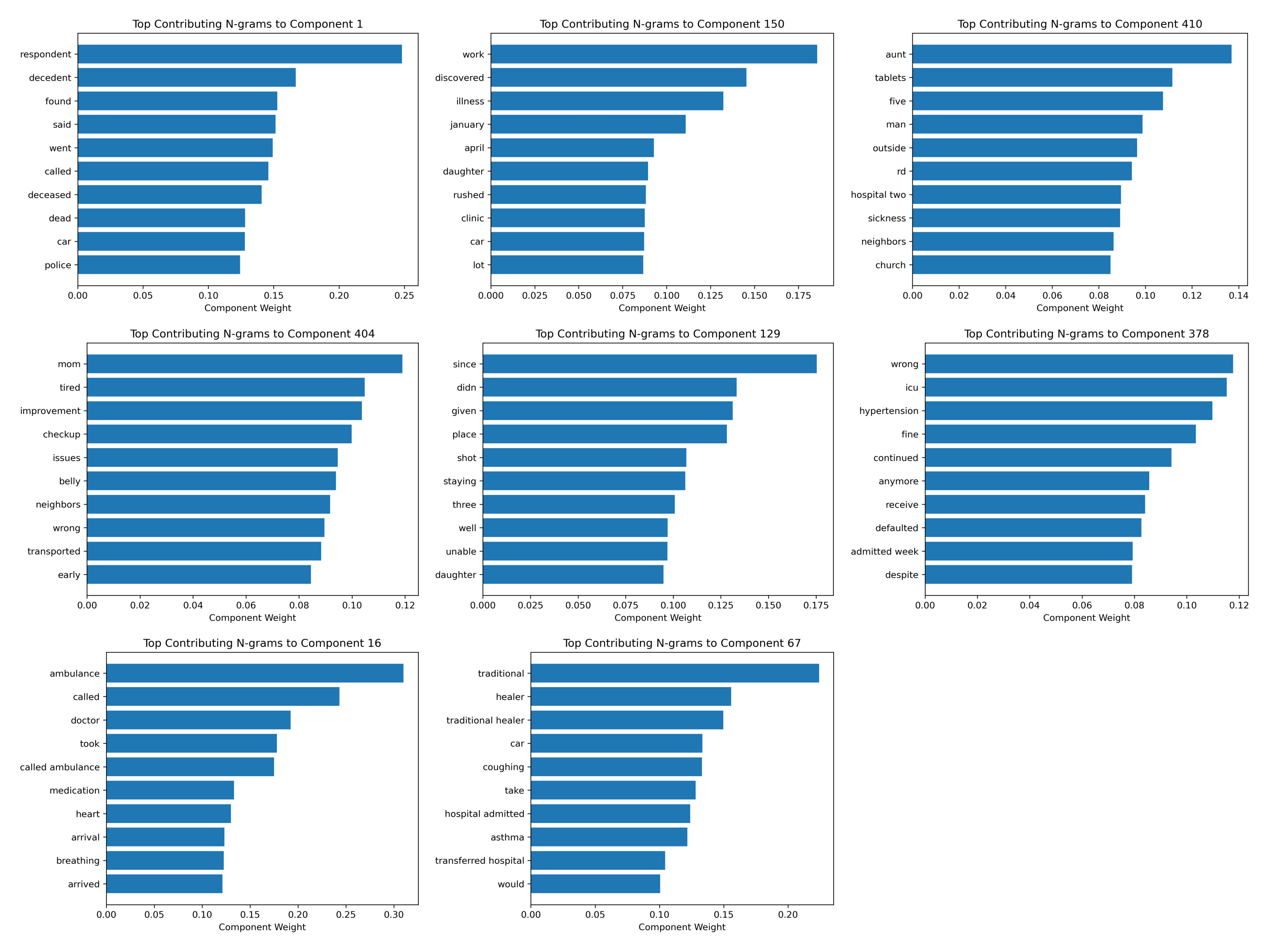}
\caption{Top contributing N-grams in top features in multimodal model}
\label{fig:suff_pred_fi_mm}
\end{center}
\end{figure}

\end{landscape}

\begin{figure}
\begin{center}
\includegraphics[width=4.3in]{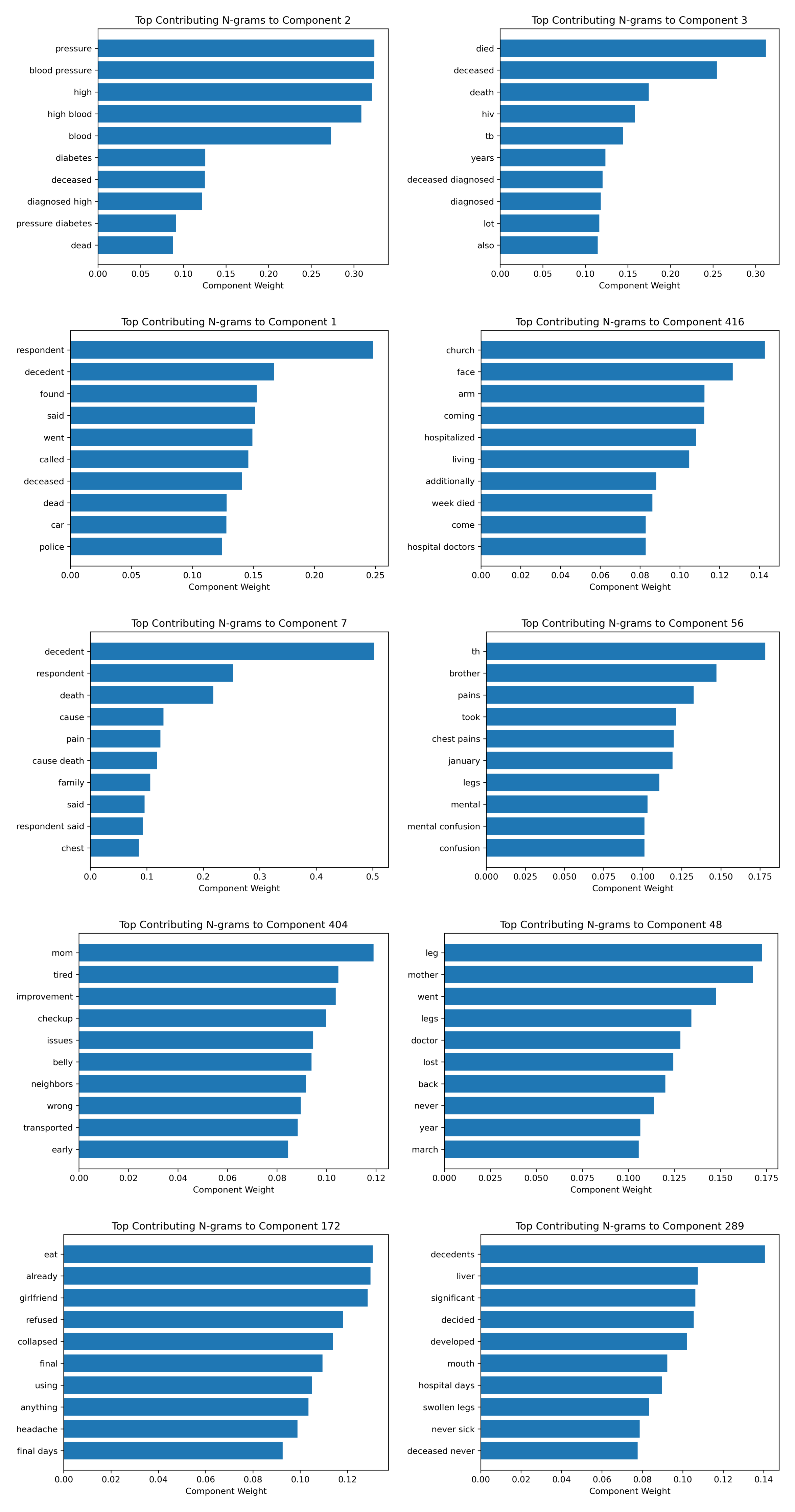}
\caption{Top contributing N-grams in top features in narrative-only model}
\label{fig:suff_pred_fi_text_ngram}
\end{center}
\end{figure}

\end{document}